\documentclass{article}

\usepackage[preprint, nonatbib]{neurips_2019}

\usepackage[utf8]{inputenc} % allow utf-8 input
\usepackage[T1]{fontenc}    % use 8-bit T1 fonts
\usepackage{hyperref}       % hyperlinks
\usepackage{url}            % simple URL typesetting
\usepackage{booktabs}       % professional-quality tables
\usepackage{amsfonts}       % blackboard math symbols
\usepackage{nicefrac}       % compact symbols for 1/2, etc.
\usepackage{microtype}      % microtypography

\usepackage{graphicx}
\usepackage[font=small]{subfig}
\usepackage{multirow}

\title{VarGNet: Variable Group Convolutional Neural Network for Efficient Embedded Computing}

\author{%
Qian Zhang, 
Jianjun Li, Meng Yao, Liangchen Song, Helong Zhou, \\ 
\textbf{Zhichao Li, Wenming Meng, Xuezhi Zhang, Guoli Wang}\\
\textit{Horizon Robotics}
}

\usepackage{xcolor}
\definecolor{bleudefrance}{rgb}{0.19, 0.55, 0.91}

\begin{document}

\maketitle

\begin{abstract}
In this paper, we propose a novel network design mechanism for efficient embedded computing. Inspired by the limited computing patterns, we propose to fix the number of channels in a group convolution, instead of the existing practice that fixing the total group numbers. Our solution based network, named Variable Group Convolutional Network (VarGNet), can be optimized easier on hardware side, due to the more unified computing schemes among the layers. Extensive experiments on various vision tasks, including classification, detection, pixel-wise parsing and face recognition, have demonstrated the practical value of our VarGNet.
\end{abstract}

\section{Introduction}
Empowering embedded systems to run the well-known deep learning architectures, such as convolutional neural networks (CNNs), has been a hot topic in recent years. For smart Internet of Things applications, the challenging part is that the whole system is required to be both energy-constrained and of small size. To meet the challenge, the work of improving the efficiency of the whole computing process can be roughly broken into two directions: The first is to design lightweight networks which has a small MAdds \cite{howard2017mobilenets, sandler2018mobilenetv2, zhang2018shufflenet, ma2018shufflenet}, thus friendly to low power consumption platforms; The second is to optimize hardware-side configurations, such as FPGA based accelerators \cite{FarabetPHL09, ZhangLSGXC15}, or to make the whole computing process more efficient by improving the compiler and generating more smart instructions \cite{abdelfattah2018dla, chen2018tvm, xing2019dnnvm}. 

All of the mentioned works above have demonstrated their great practical value in various applications. However, the real performance may not live up to the designer's expectations, due to the gap between the two different optimization directions. Specifically, for elaborately tuned networks with small MAdds, the overall latency may be high \cite{ma2018shufflenet}, while for carefully designed compilers or accelerators, the real networks may be hard to be processed. 

In this work, we intend to close the exiting gap by systematically analyze the necessary properties of a lightweight network that is friendly to the embedded hardware and the corresponding compilers. More precisely, since the computation patterns of a chip in a embedded system is strictly limited, we propose that a embedded-system-friendly network should fit into the targeted computation patterns and also the ideal data layout. By fitting into the ideal data layout, we can reduce the communication cost between on-chip memory and off-chip memory, thus fully exploit the computation throughput. 

Inspired by the observation that the computation graph of a network is easier to be optimized, if the computational intensity of the operations in a network is more balanced. We propose the variable group convolution, which is based on depthwise separable convolution \cite{krizhevsky2012imagenet, chollet2017xception, xie2017aggregated}.
In variable group convolution, the number of input channels in each group is fixed and can be tuned as a hyperparameter, which is different from the group convolution where the number of groups are fixed. 
The benefits are two folds: Fixing the number of channels is more suitable for optimization from the perspective of compilers, due to the more coherent computation pattern and data layout; Compared with depthwise convolution in \cite{howard2017mobilenets, sandler2018mobilenetv2}, which set the group number to be the channel number, variable group convolution has a larger network capacity \cite{sandler2018mobilenetv2}, thus allowing the smaller channel numbers, which helps relief the time consuming off-chip communication. 
% depthwise convolution, which is widely used in lightweight networks, requires a lot off-chip traffic   

Another key component in our network is to better exploit the on-chip memory based on the inverted residual block \cite{sandler2018mobilenetv2}. 
However, in MobileNetV2 \cite{sandler2018mobilenetv2}, the number of channels are adjusted by pointwise convolutions, which has a different computing pattern with the $3\times 3$ depthwise convolution in between and then is hard to be optimized due to limited computation patterns. Therefore, we propose that the input feature with $C$ channels is first expanded to $2C$ by variable group convolution and returned to $C$ by pointwise convolution. In this manner, the computational costs between the two types of layers are more balanced, thus being more hardware and compiler friendly. To sum up, our contributions can be summed as follows:
\begin{itemize}
\item We systematically analyze how to optimize the computation of CNNs from the perspective of both network architectures and hardware/compilers on embedded systems. We found that there exists a gap between the two optimization directions that some elaborately designed architectures are hard to be optimized due to limited computation patterns in an embedded system.
\item Observing that more unified computation pattern and data layout are more friendly to an embedded system, we propose the variable group convolution and the corresponding improved whole network, named variable group network and VarGNet for short.
\item Experiments on prevalent vision tasks, such as classification, detection, segmentation, face recognition and etc., and corresponding large scale datasets verify the practical value of our proposed VarGNet.
\end{itemize}

\subsection{Related works}

\paragraph{Lightweight CNNs.}
Designing lightweight CNNs has been a hot topic in recent years. Representative manual designed networks include SqueezeNet \cite{2016_SqueezeNet}, Xception \cite{chollet2017xception}, MobileNets \cite{howard2017mobilenets, sandler2018mobilenetv2}, ShuffleNets \cite{zhang2018shufflenet, ma2018shufflenet} and IGC \cite{zhang2017interleaved, xie2018interleaved, sun2018igcv3}.
Besides, neural architecture search (NAS) \cite{zoph2016neural, pham2018efficient, Real2018Regularized, zoph2017learning, liu2018darts} is a promising direction for automatically designing lightweight CNNs. 
The above methods are capable to effectively speed up the recognition process. 
More recently, platforms aware NAS methods are proposed \cite{cai2018proxylessnas, fbnet, dai2018chamnet, stamoulis2019single} to search some specific networks that are efficient on certain hardware platforms. 
Our network, VarGNet, is complementary to the existing platforms aware NAS methods, since the proposed variable group convolution is helpful for setting the search space in NAS methods.

\paragraph{Optimizations on CNN accelerators.}
To accelerate neural networks, FPGAs \cite{FarabetPHL09, ZhangLSGXC15, gupta2015deep, ma2017optimizing} and ASIC designs \cite{chen2014diannao, reagen2016minerva, jouppi2017datacenter, luo2017dadiannao, hegde2018ucnn} have been widely studied. 
Generally speaking, Streaming Architectures (SAs) \cite{venieris2017fpgaconvnet, xiao2017exploring} and Single Computation Engines (SCEs) \cite{guo2016angel, chang2017compiling, abdelfattah2018dla} are two kinds of FPGA based accelerators \cite{venieris2018toolflows}. 
The difference between the two directions is on customization and generality. SAs designs seek customization more than generality, while SCEs emphasize the tradeoff between flexibility and customization. 
In this work, we hope to propose a network that can be optimized by existing accelerators more easily, thus improve the overall performance.

\section{Designing Efficient Networks on Embedded Systems}
For chips used on embedded systems, such as FPGA or ASIC, a low unit price as well as a fast time to market are critical factors in designing the whole system. Such crucial points result in a relative simple chip configuration. In other words, the computation schemes are strictly limited  when compared with general-purpose processing units.
However, operators in a SOTA network are so complex that some layers can be accelerated by hardware design while others not.
Thus, for designing efficient networks on embedded systems, the first intuition here is that the layers in a network should be similar as each other in some sense. 

Another important intuition is based on two properties of convolutions used in CNNs. The first property is the computation pattern. In convolution, several filters (kernels) slide over the whole feature map, indicating that the kernels are repeatedly used while values from the feature map are only used once. The second property is the data size of convolutional kernels and feature maps. Typically, the size of convolutional kernels is much lower than the size of feature maps, such as $k^2C$ for kernels and $2HWC$ for feature maps in 2D convolutions. 
In light of the above two properties, an ingenious solution is to load all the data of kernels first and then perform the convolution with popping and popping out feature data sequentially \cite{xing2019dnnvm} .
% \notice{other citation? correctly presentation?}. 
Such practical solution is the second intuition for our following two guidelines for efficient network design on embedded systems:
\begin{itemize}
\item It will be better if the size of intermediate feature maps between blocks is smaller.
\item The computational intensity of layers in a block should be balanced.
\end{itemize}
Next, we introduce the two guidelines in detail.

\paragraph{Small intermediate feature maps between blocks.}
In SOTA networks, a common practice is to first design a normal block and a down sampling block first, and then stack several blocks together to get a deep network. 
Also, in these blocks, residual connections \cite{he2016deep} are widely adopted. So, in recent compiler-side optimizations \cite{xing2019dnnvm}
% \notice{other citation}
, layers in a block are usually grouped and computed together. 
In such manner, off-chip memory and on-chip memory only communicates when starting or ending computing a block in the network.
Therefore, a smaller intermediate feature map between blocks will certainly help reduce the data transfer time.

% \subsection{Balanced computational intensity}
\paragraph{Balanced computational intensity inside a block.}
As mentioned before, in practice, weights in several layers are loaded before performing convolution. If the loaded layers have a large divergence in terms of the computational intensity, extra on-chip memory is needed to store the intermediate slices of feature maps. In MobileNetV1 \cite{howard2017mobilenets}, a depthwise conv and a pointwise conv are used. Different from previous definitions, in our implementation, weights are already loaded. So, computational intensity is computed as MAdds divide the size of feature maps. Then, if the feature map is of size $28\times28\times256$, the computational intensity of depthwise convolution and pointwise convolution are 9 and 256, respectively. 
As a result, when running the two layers, we have to increase the on-chip buffer to satisfy the pointwise, or not grouping the computation of the two layers together.

% \subsection{Small intermediate feature maps between blocks}

\begin{figure}
  \centering
  \subfloat[Normal block.]{\includegraphics[width=0.45\textwidth]{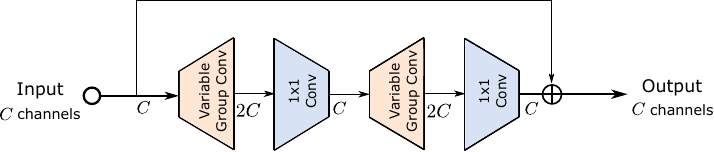}}  \quad
  \subfloat[Down sampling block.]{\includegraphics[width=0.5\textwidth]{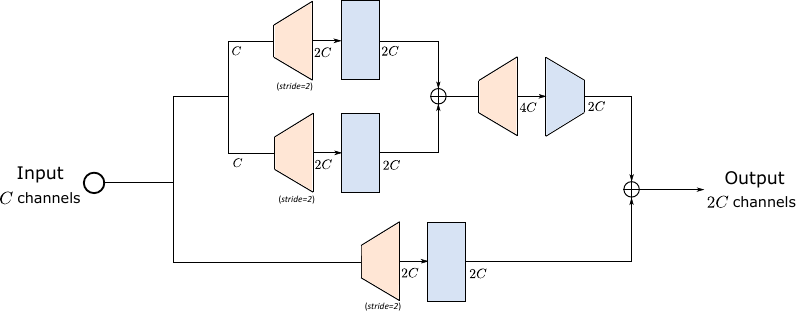}} \\
  \caption{Variable Group Network.} \label{fig:vargnet}
\end{figure}

\section{Variable Group Network}
Based on the previous mentioned two guidelines, we propose a novel network in this section. To balance the computation intensity, we set the channel numbers in a group in a network to be constant, resulting in variable groups in each convolution layers. 
The motivation of fixing the channel numbers is not hard to understand if we look at the MAdds of a convolution,
\[
\frac{k^2\times \mathrm{Height} \times \mathrm{Width} \times \mathrm{Channels}^2}{\mathrm{Groups}}.
\]
Thus, if the size of feature map is a constant, then by fixing $G=\frac{\mathrm{Channels}}{\mathrm{Groups}}$, the computational intensity inside a block is more balanced. Further, the number of channels in a group can be set to satisfy the configurations of the processing elements, in which channels of a certain number will be processed every time.

\begin{figure}
  \centering
  \includegraphics[width=\textwidth]{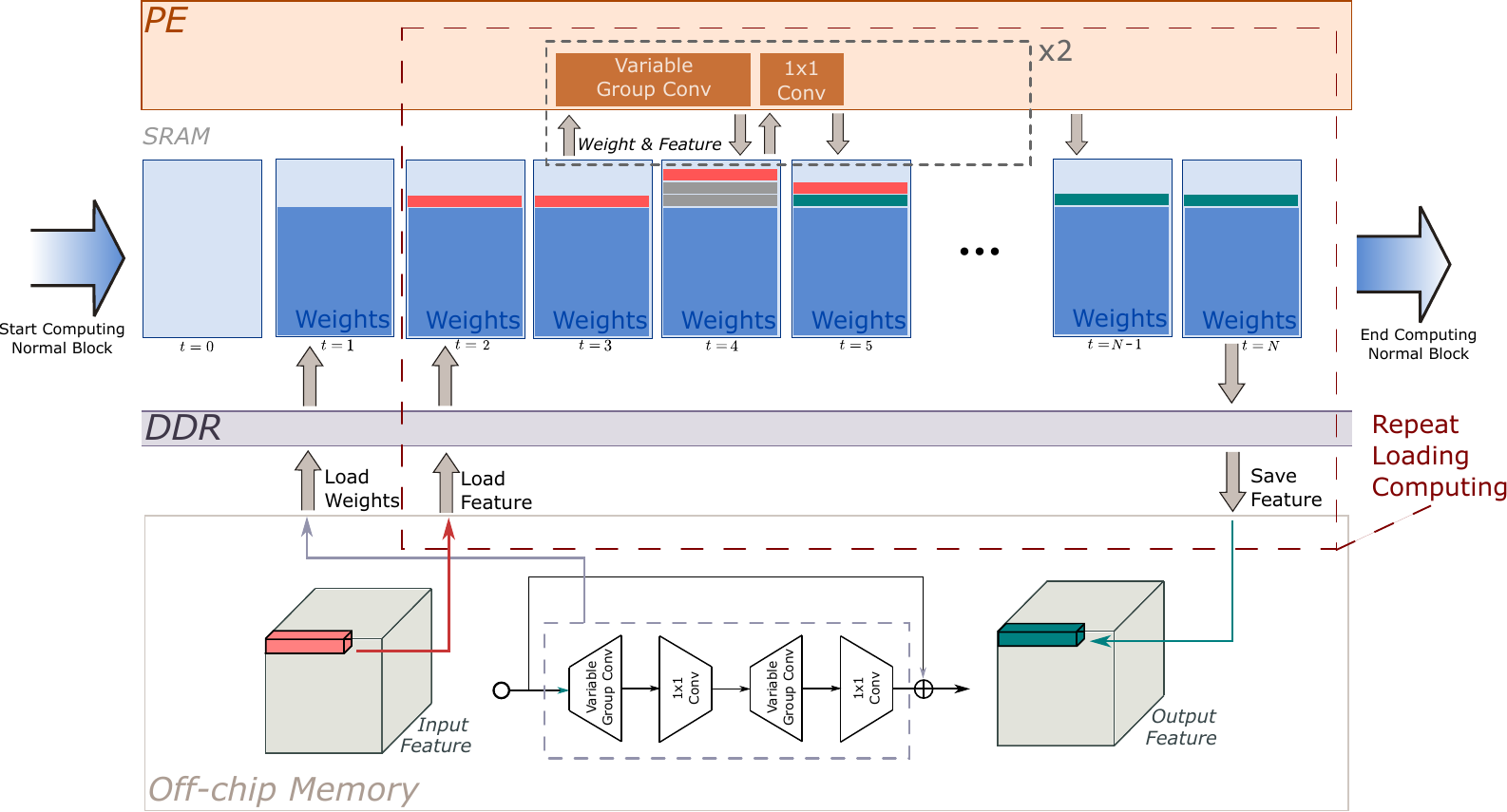}
  \caption{Computing scheme of a normal block in Variable Group Network. The weights of four convolution operations are first loaded onto on-chip memory, and then processing the features.} \label{fig:compute}
\end{figure}

Compared with depthwise convolution, the variable group convolution increases the MAdds as well as the expressiveness \cite{sandler2018mobilenetv2}. Thus, now we are able to reduce the channel number of intermediate feature maps, while keeping the same generalizing ability as previous networks. Specifically, we design novel network blocks as shown in Fig. \ref{fig:vargnet}. For the normal block used in the early stages in the whole network, since the size of weights are relatively small at this time, the weights of the four layers can be all cached into the on-chip memory. When entering the late stages, where channel numbers increase and the size of weights increase as well, the normal block is also able to be optimized by only loading a variable group conv and a pointwise conv. Similarly, the operations in down sampling block are also friendly to the compiler-side and hardware-side optimizations. 
The whole computing process for a normal block is demonstrated in Fig. \ref{fig:compute}.
Then, based on the architecture of MobileNetV1 \cite{howard2017mobilenets}, we substitute their basic blocks to ours and the whole detailed network architecture is shown in Tab. \ref{tab:vargnetv1}. Also, another ShuffleNet v2 based architecture is shown in Tab. \ref{tab:vargnetv2}.

\begin{table}
\centering
% \caption{Overall architecture of MobileNetV1 \cite{howard2017mobilenets} based Variable Group Network (VarGNet v1).}
\caption{Overall architecture of Variable Group Network v1.}
\label{tab:vargnetv1}
\resizebox{\textwidth}{!}{
  % \begin{tabular}{c|c|c|c|c|c|c|c|c|c|c|c|c}
  \begin{tabular}{c|c|c|c|c|c|c|c|c|c|c|c}
  \hline\hline
  \multirow{2}{*}{Layer} & \multirow{2}{*}{Output Size} & \multirow{2}{*}{KSize} & \multirow{2}{*}{Stride} & \multirow{2}{*}{Repeat} & 
  \multicolumn{7}{c}{Output Channels}
   \\  \cline{6-12}
  % \multicolumn{8}{c}{Output Channels}
  %  \\  \cline{6-13}
  %  & & & & & 0.25x & 0.5x & 0.75x & 1x & 1.25x & 1.5x & 1.75x & 2x \\ \hline
  %  Image & 224 x 224 &  &  &  & 3 & 3 & 3 & 3 & 3 & 3 & 3 & 3 \\ \hline
  %  Conv 5 & 112 x 112 & 3 x 3 & 2 & 1 & 8 & 16 & 24 & 32 & 40 & 48 & 56 & 64 \\ \hline
  %  DownSample & 56 x 56 &  & 2 & 3 & 16 & 32 & 48 & 64 & 80 & 96 & 112 & 128 \\ \hline
  %  DownSample & 28 x28 &  & 2 & 1 & 32 & 64 & 96 & 128 & 160 & 192 & 224 & 256 \\ \hline
  %  DownSample & 14 x 14 &  & 2 & 1 & 64 & 128 & 192 & 256 & 320 & 384 & 448 & 512 \\ \hline
  %  Stage Block & 14 x 14 &  & 1 & 2 & 64 & 128 & 192 & 256 & 320 & 384 & 448 & 512 \\ \hline
  %  DownSample & 7 x 7 &  & 2 & 1 & 128 & 256 & 384 & 512 & 640 & 768 & 896 & 1024 \\ \hline
  %  Stage Block & 7 x 7 &  & 1 & 1 & 128 & 256 & 384 & 512 & 640 & 768 & 896 & 1024 \\ \hline
  %  Conv 5 & 7 x 7 & 1 x 1 & 1 & 1 & 1024 & 1024 & 1024 & 1024 & 1280 & 1536 & 1792 & 2048 \\ \hline
  %  Global Pool & 1 x 1 & 7 x 7 &  &  &  &  &  &  &  &  &  &  \\ \hline
  %  FC &  &  &  &  & 1000 & 1000 & 1000 & 1000 & 1000 & 1000 & 1000 & 1000 \\  \hline\hline
   & & & & & 0.25x & 0.5x & 0.75x & 1x & 1.25x & 1.5x & 1.75x\\ \hline
   Image & 224 x 224 &  &  &  & 3 & 3 & 3 & 3 & 3 & 3 & 3\\ \hline
   Conv 1 & 112 x 112 & 3 x 3 & 2 & 1 & 8 & 16 & 24 & 32 & 40 & 48 & 56\\ \hline
   DownSample & 56 x 56 &  & 2 & 3 & 16 & 32 & 48 & 64 & 80 & 96 & 112\\ \hline
   DownSample & 28 x28 &  & 2 & 1 & 32 & 64 & 96 & 128 & 160 & 192 & 224\\ \hline
   DownSample & 14 x 14 &  & 2 & 1 & 64 & 128 & 192 & 256 & 320 & 384 & 448\\ \hline
   Stage Block & 14 x 14 &  & 1 & 2 & 64 & 128 & 192 & 256 & 320 & 384 & 448\\ \hline
   DownSample & 7 x 7 &  & 2 & 1 & 128 & 256 & 384 & 512 & 640 & 768 & 896\\ \hline
   Stage Block & 7 x 7 &  & 1 & 1 & 128 & 256 & 384 & 512 & 640 & 768 & 896\\ \hline
   Conv 5 & 7 x 7 & 1 x 1 & 1 & 1 & 1024 & 1024 & 1024 & 1024 & 1280 & 1536 & 1792\\ \hline
   Global Pool & 1 x 1 & 7 x 7 &  &  &  &  &  &  &  &  &  \\ \hline
   FC &  &  &  &  & 1000 & 1000 & 1000 & 1000 & 1000 & 1000 & 1000\\  \hline\hline
\end{tabular}}
\end{table}

\begin{table}
  \centering
  % \caption{Overall architecture of ShuffleNet \cite{zhang2018shufflenet} based Variable Group Network (VarGNet v2).}\label{tab:vargnetv2}
  \caption{Overall architecture of Variable Group Network v2. Head Block is a modified version of Normal Block, by setting the stride to 2 and keeping the channel numbers unchanged after two variable convolution layers.}\label{tab:vargnetv2}
  \resizebox{\textwidth}{!}{
  \begin{tabular}{c|c|c|c|c|c|c|c|c|c|c|c|c}
  \hline\hline
  \multirow{2}{*}{Layer} & \multirow{2}{*}{Output Size} & \multirow{2}{*}{KSize} & \multirow{2}{*}{Stride} & \multirow{2}{*}{Repeat} & 
  \multicolumn{8}{c}{Output Channels}
   \\  \cline{6-13}
   & & & & & 0.25x & 0.5x & 0.75x & 1x & 1.25x & 1.5x & 1.75x & 2x \\ \hline
   Image& 224 x 224& & & & 3& 3& 3& 3& 3& 3& 3& 3 \\ \hline
   Conv 1& 112 x 112& 3 x 3& 2& 1& 8& 16& 24& 32& 40& 48& 56& 64 \\ \hline
   Head Block& 56 x 56& & 2& 1& 8& 16& 24& 32& 40& 48& 56& 64 \\ \hline
   \multirow{2}{*}{Stage 2}& 28 x28& & 2  & 1 &  \multirow{2}{*}{16}
   & \multirow{2}{*}{32} & \multirow{2}{*}{48} & \multirow{2}{*}{64} & \multirow{2}{*}{80} 
   & \multirow{2}{*}{96} & \multirow{2}{*}{112} & \multirow{2}{*}{128}  \\ \cline{2-5}
   & 28 x 28& & 1& 2& & & & & & & &  \\ \hline
   \multirow{2}{*}{Stage 3}& 14 x 14& & 2 & 1 & \multirow{2}{*}{32}
   & \multirow{2}{*}{64}& \multirow{2}{*}{96}& \multirow{2}{*}{128}& \multirow{2}{*}{160}
   & \multirow{2}{*}{192}& \multirow{2}{*}{224}& \multirow{2}{*}{256} \\ \cline{2-5}
   & 14 x 14& & 1& 6& & & & & & & &  \\ \hline
   \multirow{2}{*}{Stage 4}& 7 x 7& & 2 & 1 & \multirow{2}{*}{64}
   & \multirow{2}{*}{128}& \multirow{2}{*}{192}& \multirow{2}{*}{256}& \multirow{2}{*}{320}
   & \multirow{2}{*}{384}& \multirow{2}{*}{448}& \multirow{2}{*}{512} \\ \cline{2-5}
   & 7 x 7& & 1& 3& & & & & & & &  \\ \hline
   Conv 5& 7 x 7& 1 x 1& 1& 1& 1024& 1024& 1024& 1024& 1280& 1536& 1792& 2048 \\ \hline
   Global Pool& 1 x 1& 7 x 7& & & & & & & & & &  \\ \hline
   FC& & & & & 1000& 1000& 1000& 1000& 1000& 1000& 1000& 1000 \\ \hline\hline
  \end{tabular}}
\end{table}

\begin{table}
  \centering
  \caption{VarGNet v1 performance on ImageNet. ($G$ is the number of channels in a group.)}\label{tab:vargnetv1_comp}
  \begin{tabular}{c}
    (a) $G=4$ \\
    \begin{tabular}{cccccc}
      \toprule
      Model Scale & Acc-top1 (float32) & Acc-top1 (int8) & Model size & MAdds & Max Channels \\ \midrule
      0.25 & 64.57\% & 65.02\% & 1.44M & 55M & 128\\
      0.5 & 69.67\% & 70.33\% & 2.23M & 157M & 256\\
      0.75 & 72.36\% & 72.56\% & 3.43M & 309M & 384\\
      1 & 74.04\% & 74.11\% & 5.02M & 509M & 512\\
      % 1.25 & 74.34\% & 7.42M & 767M & 640\\
      % 1.5 & 74.47\% & 10.28M & 1.05G & 768\\
      % 2 &  & 15M & 1.8G & 1024\\ 
      \bottomrule
    \end{tabular} \\
    (b) $G=8$ \\
    \begin{tabular}{cccccc}
      \toprule
      Model Scale & Acc-top1 (float32) & Acc-top1 (int8) & Model size & MAdds & Max Channels \\ \midrule
      % Model Scale & Acc(top1) & Model size & MAdds & Max Channels \\ \midrule
      0.25 & 65.44\% & 65.61\% & 1.5M & 75M & 128 \\
      0.5 & 70.67\% & 70.84\% & 2.37M & 198M & 256 \\
      0.75 & 73.28\% & 73.35\% & 3.66M & 370M & 384 \\
      1 & 74.87\% & 74.90\% & 5.33M & 590M & 512 \\
      % 1.25 & 74.70\% & 7.8M & 869M & 640 \\
      % 1.5 & 75.00\% & 10.7M & 1.17G & 768 \\
      % 1.75 & 75.30\% & 14.1M & 1.54G & 1024 \\
      % 2 &  & 18.0M & 1.9G & 2048 \\
      \bottomrule
    \end{tabular} \\
    (c) Comparison network: MobileNet v1\\
    \begin{tabular}{ccccc}
      \toprule
      Model Scale & Acc(top1) & Model size & MAdds & Max Channels \\ \midrule
      0.35 & 60.4\% & 0.7 M & 72 M & 358 \\
      0.6 & 68.6\% & 1.7 M & 201 M & 614 \\
      0.85 & 72.0\% & 3.1M & 394 M & 870  \\
      1.0 & 73.3\% & 4.1M & 542 M & 1024  \\
      1.05 & 73.5\% & 4.4 M & 594 M & 1075 \\
      1.3 & 74.7\% & 6.4 M & 903 M & 1331 \\
      1.5 & 75.1\% & 8.3 M & 1.17 G & 1536  \\
      % 1.75 & & 11 M & 1.59 G & 1792 \\
      \bottomrule
    \end{tabular}
  \end{tabular}
\end{table}

\begin{table}
  \centering
  \caption{VarGNet v2 performance on ImageNet. ($G$ is the number of channels in a group.)}\label{tab:vargnetv2_comp}
  \begin{tabular}{c}
    (a) $G=4$ \\
    \begin{tabular}{cccccc}
      \toprule
      Model Scale & Acc-top1 (float32) & Acc-top1 (int8) & Model size & MAdds & Max Channels \\ \midrule
      % Model Scale & Acc(top1) & Model size & MAdds & Max Channels \\ \midrule
      0.25 & 60.53\% & 61.33\% & 1.27M & 35M & 64 \\
      0.5 & 67.52\% & 68.60\% & 1.72M & 92M & 128 \\
      0.75 & 70.97\% & 71.84\% & 2.35M & 173M & 192 \\
      1 & 73.18\% & 73.55\% & 3.19M & 278M & 256 \\
      % 1.25 & 74.08\% & 4.55M & 411M & 320 \\
      % 1.5 & 74.91\% & 6.14M & 569M & 384 \\
      % 2 & 75.44\% & 10.0M & 961M & 512 \\
      \bottomrule
    \end{tabular} \\
    (b) $G=8$ \\
    \begin{tabular}{cccccc}
      \toprule
      Model Scale & Acc-top1 (float32) & Acc-top1 (int8) & Model size & MAdds & Max Channels \\ \midrule
      % Model Scale & Acc(top1) & Model size & MAdds & Max Channels \\ \midrule
      0.25 & 61.74\% & 62.46\% & 1.35M & 51M & 64 \\
      0.5 & 68.40\% & 68.83\% & 1.87M & 124M & 128 \\
      0.75 & 71.48\% & 72.07\% & 2.58M & 222M & 192 \\
      1 & 73.76\% & 73.76\% & 3.49M & 343M & 256 \\
      % 1.25 & 74.34\% & 4.94M & 492M & 320 \\
      % 1.5 & 75.04\% & 6.60M & 666M & 384 \\
      % 1.75 & 75.49\% & 8.50M & 866M & 448 \\
      % 2 & 75.71\% & 10.6M & 1.06G & 512 \\
      \bottomrule
    \end{tabular} \\
    (c) Comparison network: ShuffleNet v2\\
    \begin{tabular}{ccccc}
      \toprule
      Model Scale & Acc(top1) & Model size & MAdds & Max Channels \\ \midrule
      0.25 (60) & 63.85\% & 1.47M & 51M & 240 \\
      0.5 (108) & 68.74\% & 2.1M & 123M & 432 \\
      0.75 (154) & 71.65\% & 2.92M & 223M & 616 \\
      1 (196) & 73.17\% & 3.87M & 342M & 784 \\
      1.25 (228) & 74.15\% & 6.63M & 494M & 912 \\
      1.5 (270) & 74.56\% & 8.06M & 666M & 1080 \\
      1.75 (312) & 75.24\% & 9.68M & 863M & 1248 \\
      \bottomrule
    \end{tabular}
  \end{tabular}
\end{table}

\section{Experiments}
\subsection{ImageNet Classification}
The results of our model on ImageNet are presented in Tab. \ref{tab:vargnetv1_comp} and Tab. \ref{tab:vargnetv2_comp}.
Training hyperparameters are set as: batch size 1024, crop ratio 0.875, learning rate 0.4, cosine learning rate schedule, weight decay 4e-5 and training epochs 240. We can observe that VarGNet v1 performs better than MobileNet v1, as shown in Tab. \ref{tab:vargnetv1_comp}. From (c) in Tab. \ref{tab:vargnetv2_comp}, we can see that when the model scale is small, the performance of VarGNet v2 is worse than ShuffleNet v2, due to less channels used in our VarGNet v2. Then, when the model size is large, our network performs better.

\subsection{Object Detection}

% In Tab. \ref{tab:det1} and Tab. \ref{tab:det2}, 
In Tab. \ref{tab:det2}, we present the performance of our proposed VarGNet as well as comparison methods. 
We evaluate the object detection performance of our proposed networks on COCO datasets \cite{Lin2014MicrosoftCC} and compare them with other state-of-the-art lightweight architectures.  We choose FPN-based Faster R-CNN \cite{Lin2017FeaturePN} as the framework and all the experiments are implemented under the same settings with the input resolution being 800$\times$1333 and the number of epochs being 18. Specially, we find that ShuffleNet v2 achieves better accuracy if trained with more epochs so a model with 27 epochs is trained for ShuffleNet v2. 1000 proposals per image are evaluated in RPN stage at test time.  We use train+val set for training except 8000 minimal images and finally test on minival set. 
% The mAP along with the model complexity are listed in the Table. 

% \begin{table}
%   \centering
%   \caption{Performance on COCO object detection with Faster R-CNN. The input image size is 600 $\times$ 1000.}\label{tab:det1}
%   \begin{tabular}{lcc}
%   \toprule
%   Network & MAdds & mAP \\ \midrule
%   MobileNet v1 1.0 & 18.99 & 26.4 \\  
%   MobileNet v2 1.0 & 12.00 & 24.3 \\ \midrule
%   ShuffleNet v1 1.0 & 10.58 & 3.3? \\  
%   ShuffleNet v2 1.0 (inv\_res=32) & 8.31 & 3.1? \\ \midrule
%   VarGNet v1 1.0 & 19.65 & 27.5 \\  
%   VarGNet v2 1.0 (with head pooling) & 12.51 & 25.9 \\  
%   VarGNet v2 1.0 (wo head pooling) & 12.86 & 26.5 \\  
%   VarGNet v2 0.5 (wo head pooling) & 7.73 & 22.6 \\ \bottomrule
%   \end{tabular}
% \end{table}

\begin{table}
  \centering
  \caption{Performance on COCO object detection with FPN based Faster R-CNN. The input image size is 800 $\times$ 1333.}\label{tab:det2}
  \begin{tabular}{lcc}
  \toprule
  Network & MAdds (G) & mAP \\ \midrule
  MobileNet v1 1.0 & 24.15 & 31.1 \\  
  MobileNet v2 1.0 & 18.71 & 31.0 \\ \midrule
  ShuffleNet v1 1.0 & 15.31 & 27.9 \\  
  ShuffleNet v2 1.0 & 15.55 & 27.5 \\ 
  ShuffleNet v2 1.0 (27 epochs) & 15.55 & 28.9 \\ 
  \midrule
  VarGNet v1 1.0 & 24.91 & 33.7 \\  
  % VarGNet v2 1.0 (with head pooling) & 18.99 & 33.1 \\  
  VarGNet v2 0.5 & 14.98 & 28.6 \\ 
  VarGNet v2 1.0 & 19.61 & 33.3 \\ \bottomrule
  \end{tabular}
\end{table}

\subsection{Pixel Level Parsing}

\subsubsection{Cityscapes}
\begin{figure}
  \centering
  \subfloat[The multi-task network used in Cityscapes experiments.]{\label{fig:netscs} \includegraphics[width=0.45\textwidth]{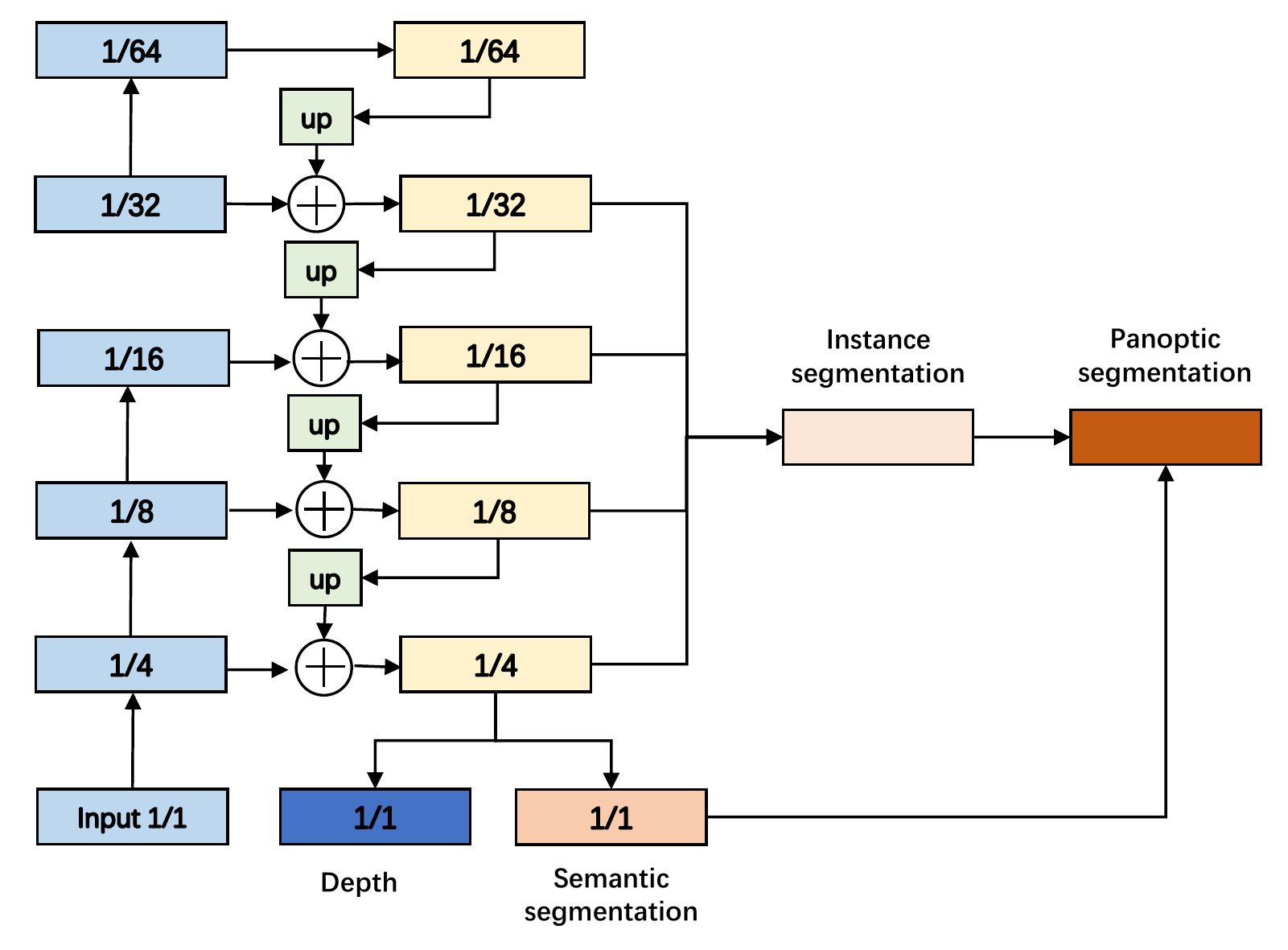}} \hfill
  \subfloat[The U-Net style network used in KITTI experiments.]{\label{fig:netskitti} \includegraphics[width=0.45\textwidth]{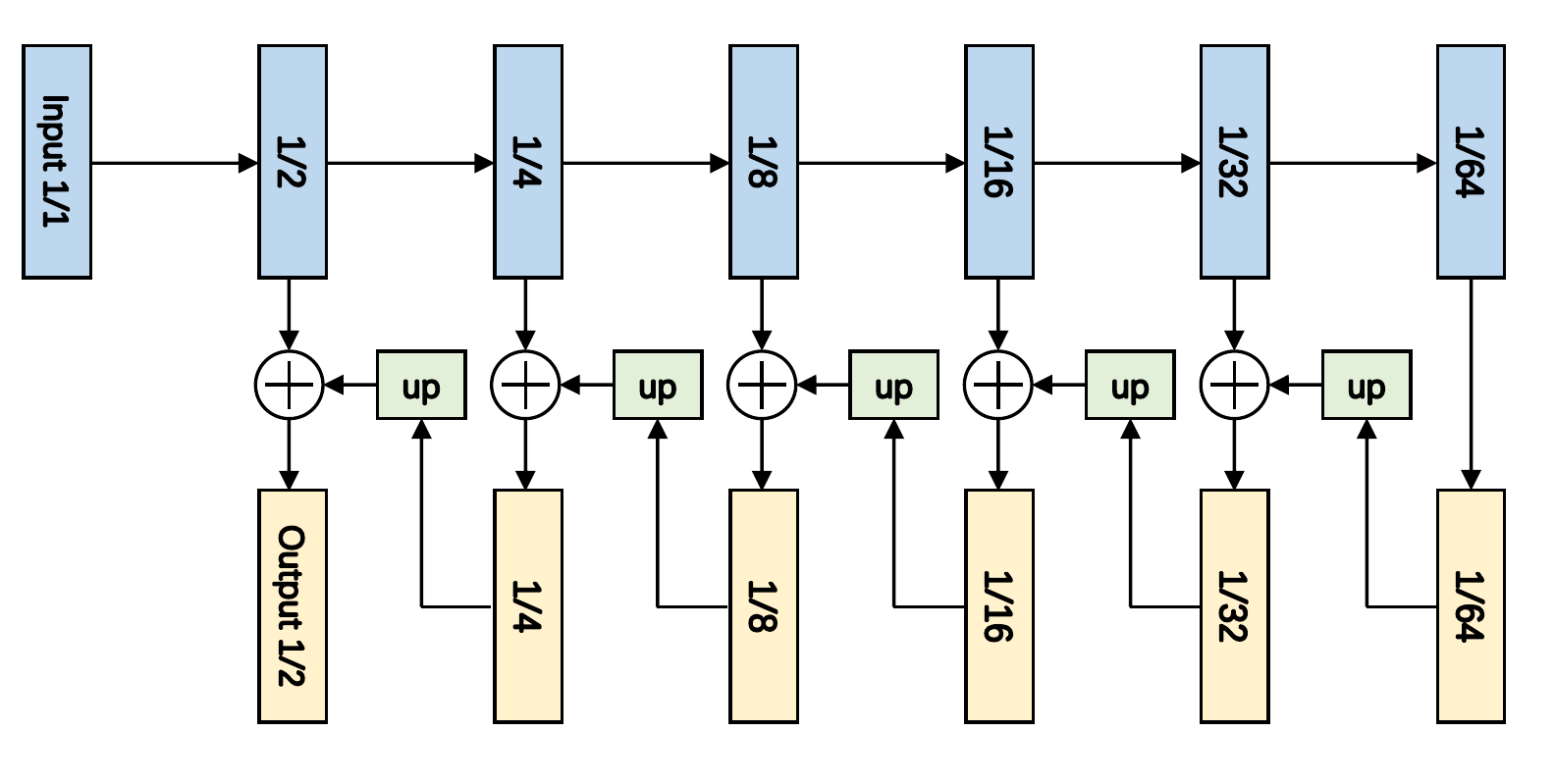}}
  \caption{Network architectures.} \label{fig:nets}
\end{figure}

On Cityscapes dataset \cite{cordts2016cityscapes}, we designed a multi-task structure (Fig. \ref{fig:netscs}) to conduct two important pixel level parsing tasks: single image depth prediction and segmentation. 

\paragraph{Traning setup.} We use the standard Adam Optimizer with weight decay set to 1e-5 and batch size set to 16. The learning rate is initialized as 1e-4 and follows a polynomial decay with power of 0.9. Total training epochs are set as 100. For data augmentation, random horizontal flip is used and images are resized with scale randomly chosen from 0.6-1.2.
For multitask training, we have the loss function defined as
\[
L_{\mathrm{total}}=\lambda_{\mathrm{instance}} L_{\mathrm{instance}}+\lambda_{\mathrm{semantic}} L_{\mathrm{semantic}}+\lambda_{\mathrm{depth}} L_{\mathrm{depth}}.
\]
When the task is panoptic segmentation, we set $\lambda_{\mathrm{instance}}=0.2, \lambda_{\mathrm{semantic}}=1.0$. After adding depth task, we set $\lambda_{\mathrm{depth}}=0.08$.

\begin{figure}
  \centering
  \resizebox{\textwidth}{!}{
  \begin{tabular}{cccc}
    \toprule
    \small{Image} & \small{VarGNet v1} & \small{VarGNet v2} & \small{GT} \\ \midrule
    \includegraphics[width=0.23\textwidth]{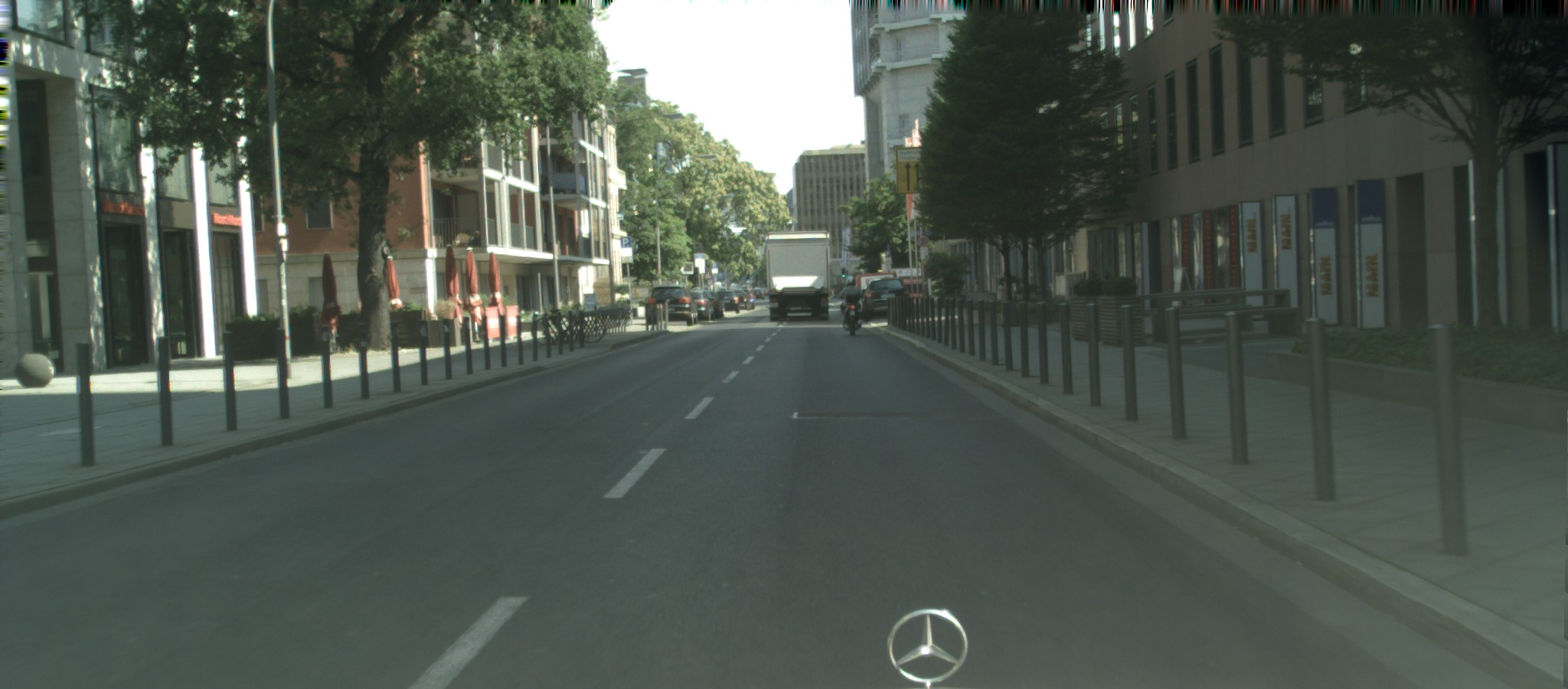} &
    \includegraphics[width=0.23\textwidth]{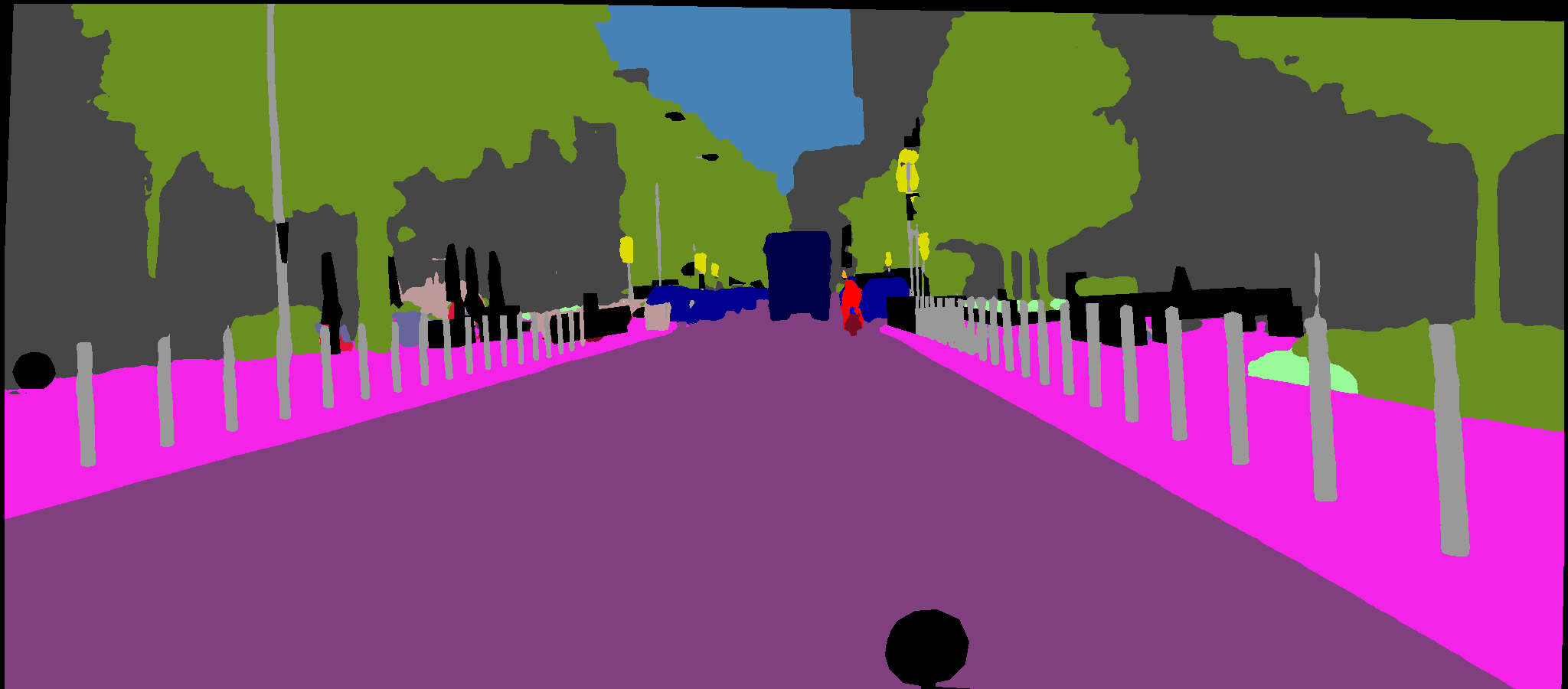} &
    \includegraphics[width=0.23\textwidth]{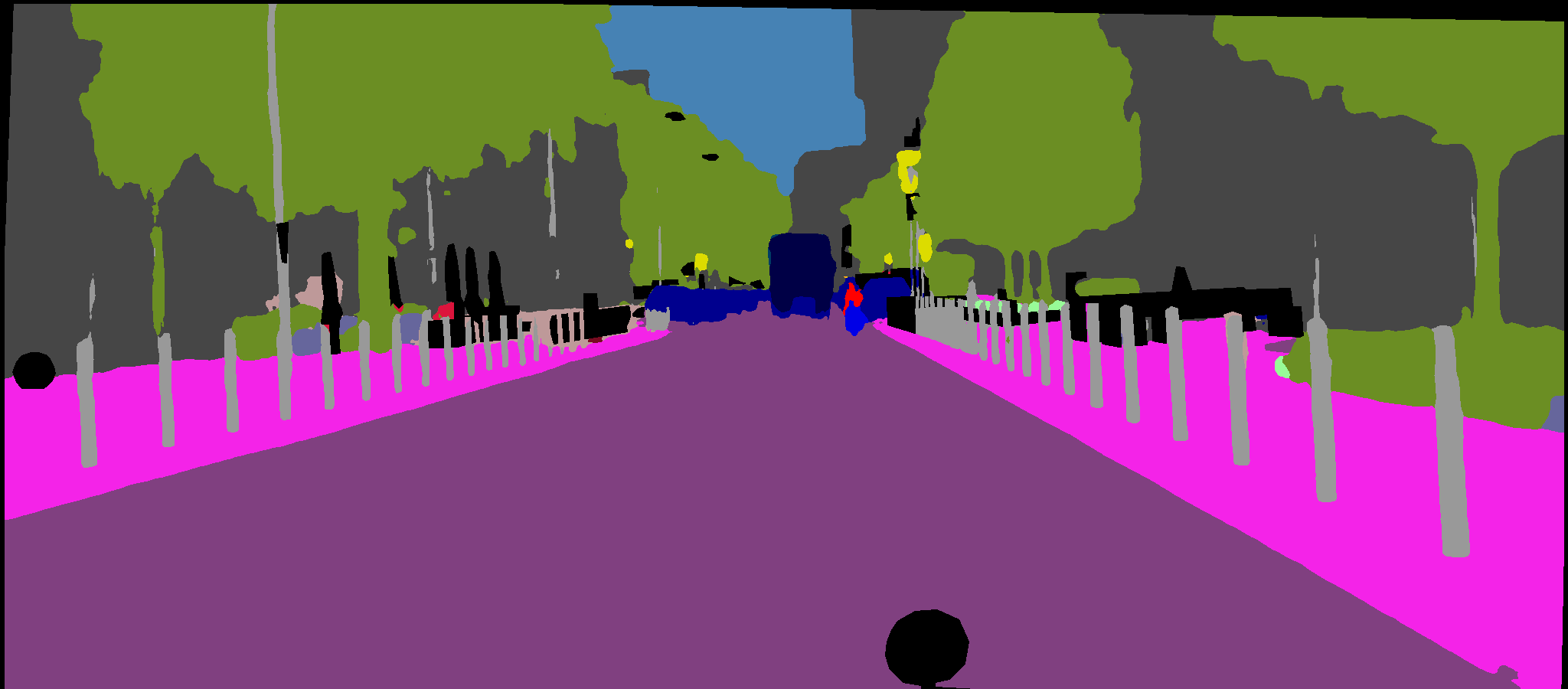} &
    \includegraphics[width=0.23\textwidth]{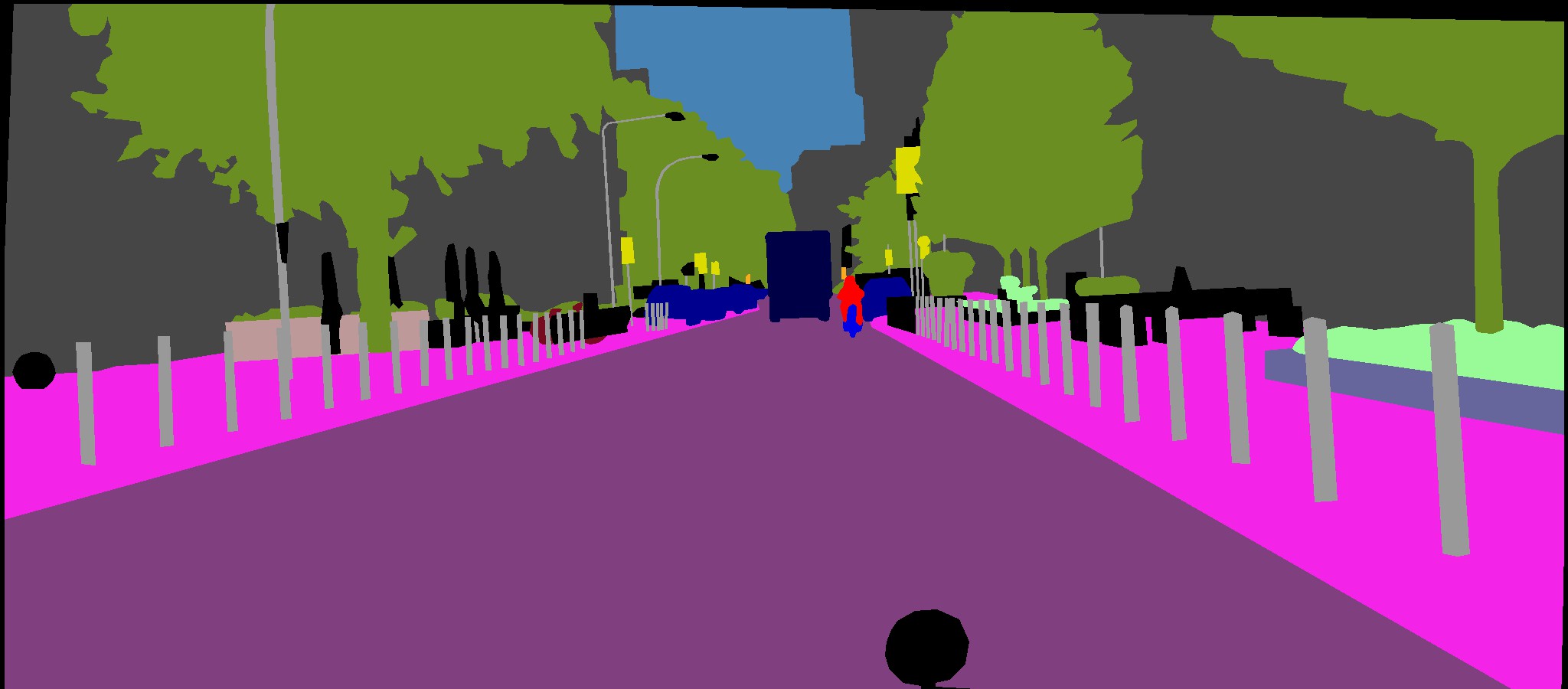} \\
    & \multicolumn{3}{c}{\small{Semantic Segmentation}} \\
    & \includegraphics[width=0.23\textwidth]{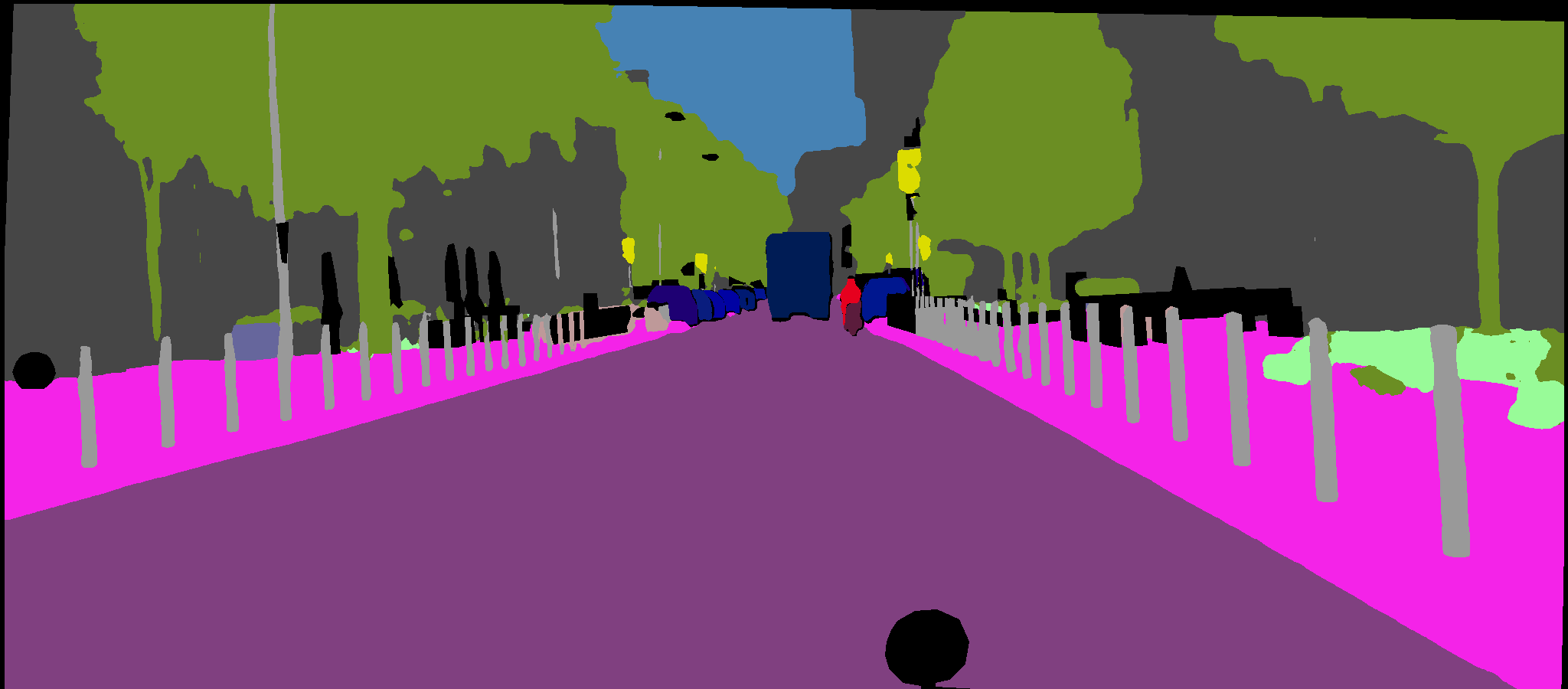} &
    \includegraphics[width=0.23\textwidth]{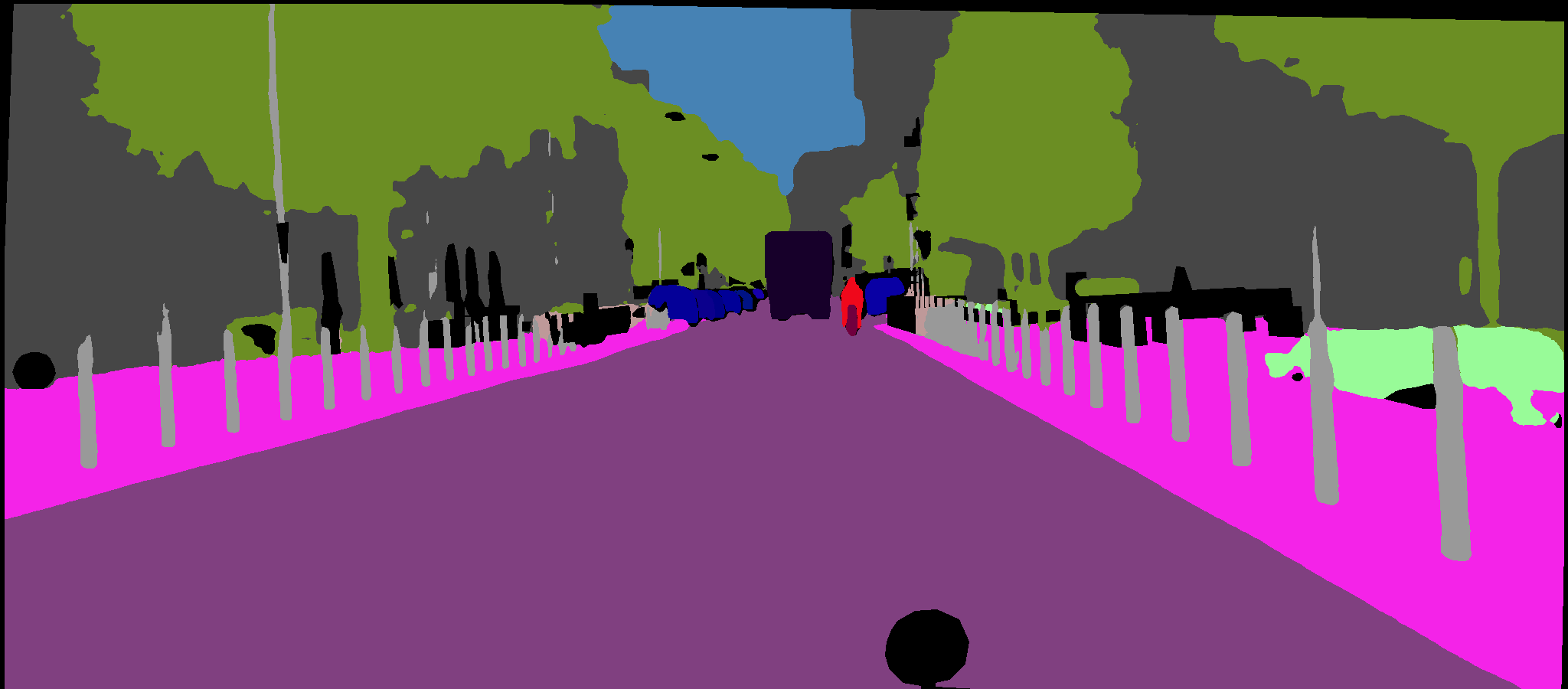} &
    \includegraphics[width=0.23\textwidth]{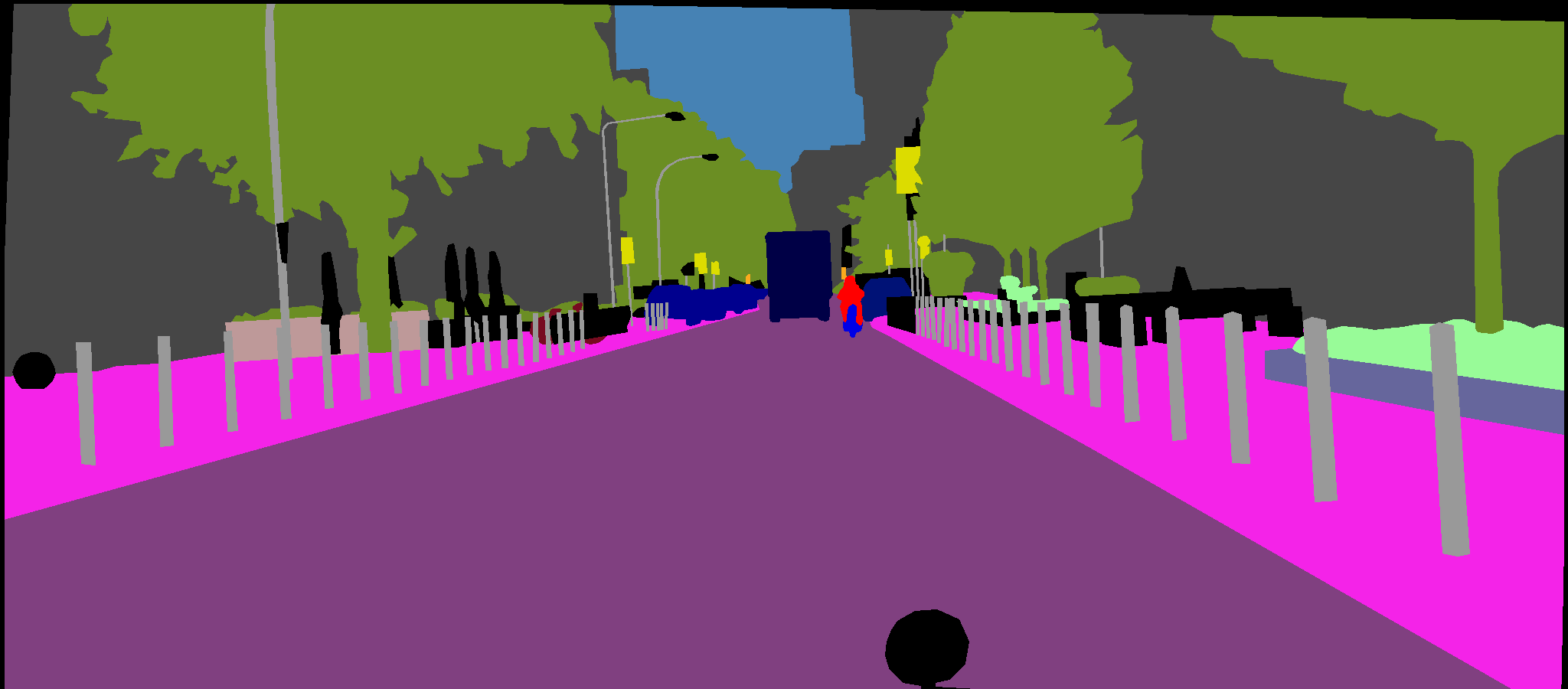} \\
    & \multicolumn{3}{c}{\small{Panoptic Segmentation}} \\
    & \includegraphics[width=0.23\textwidth]{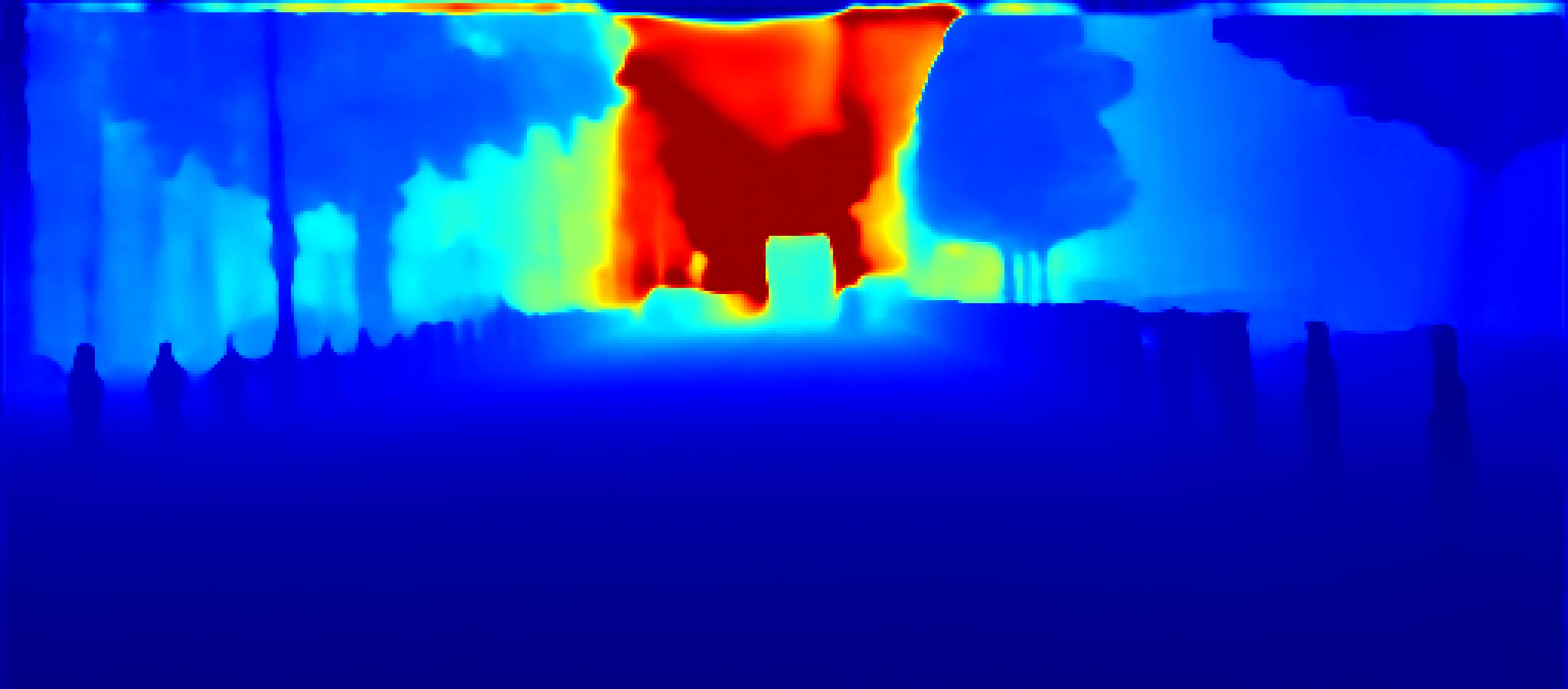} &
    \includegraphics[width=0.23\textwidth]{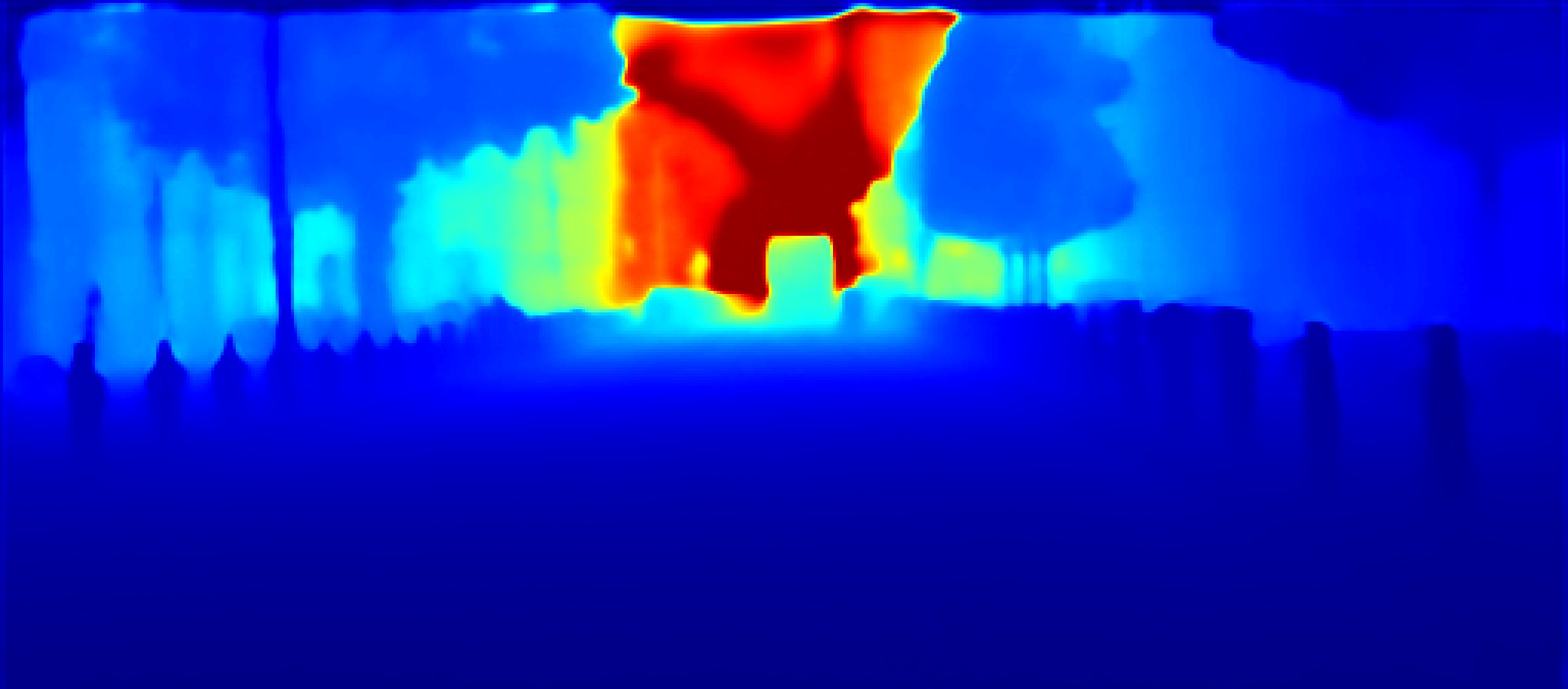} &
    \includegraphics[width=0.23\textwidth]{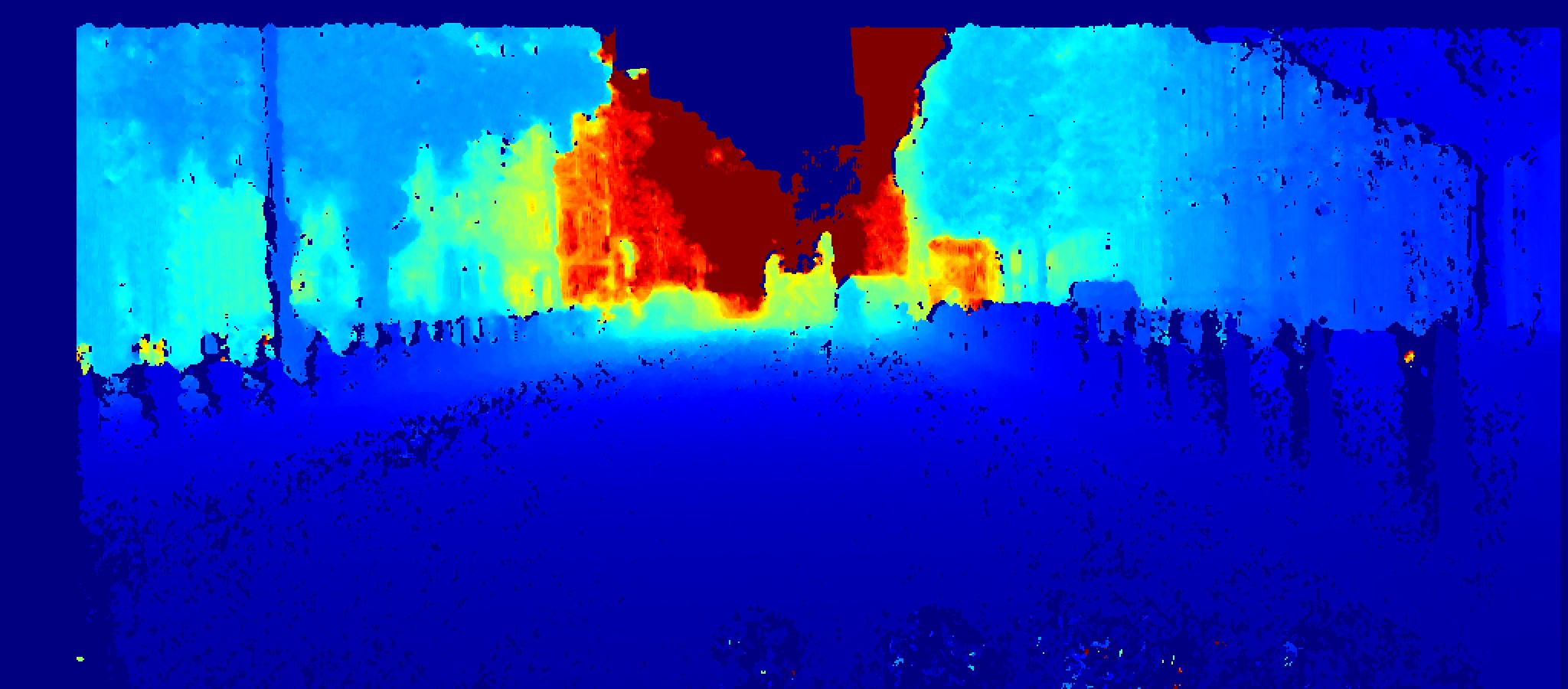} \\
    & \multicolumn{3}{c}{\small{Depth}} \\ \midrule
    \includegraphics[width=0.23\textwidth]{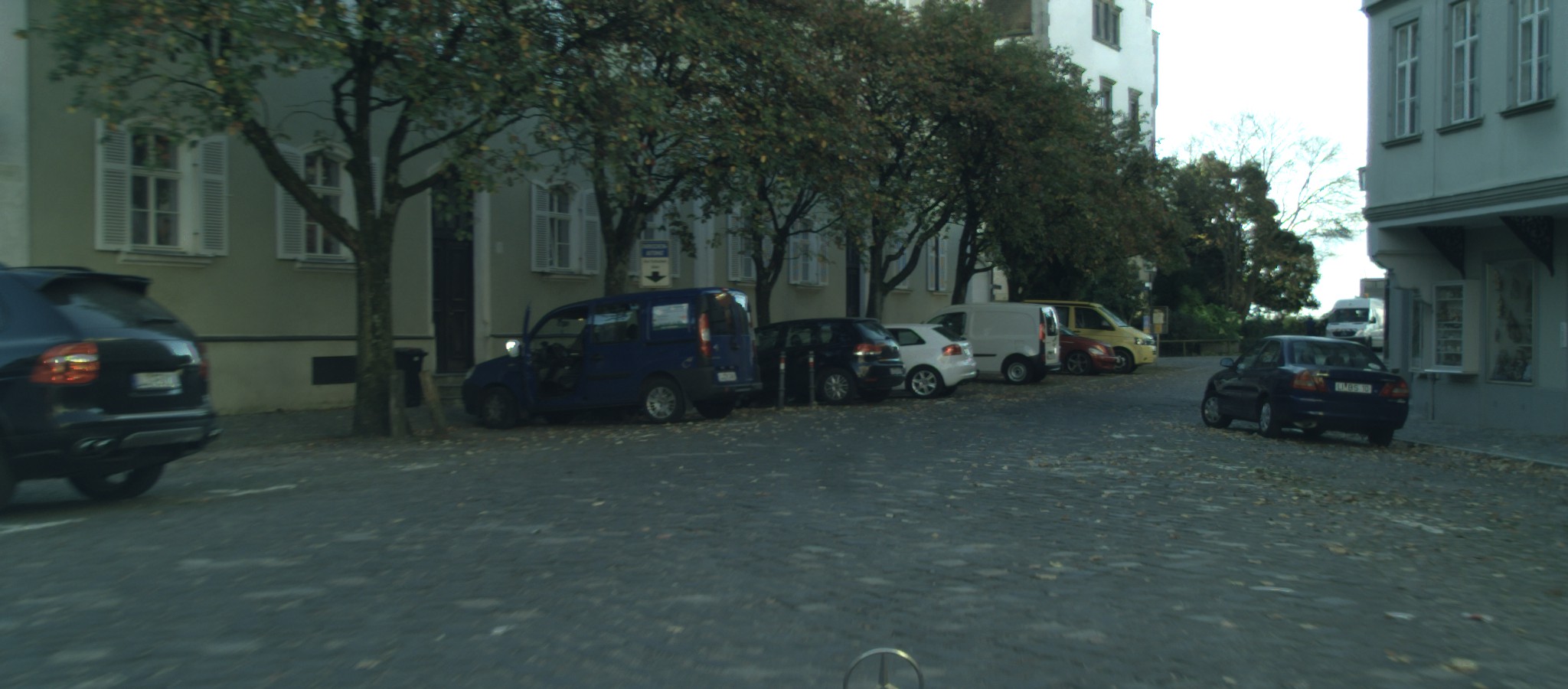} &
    \includegraphics[width=0.23\textwidth]{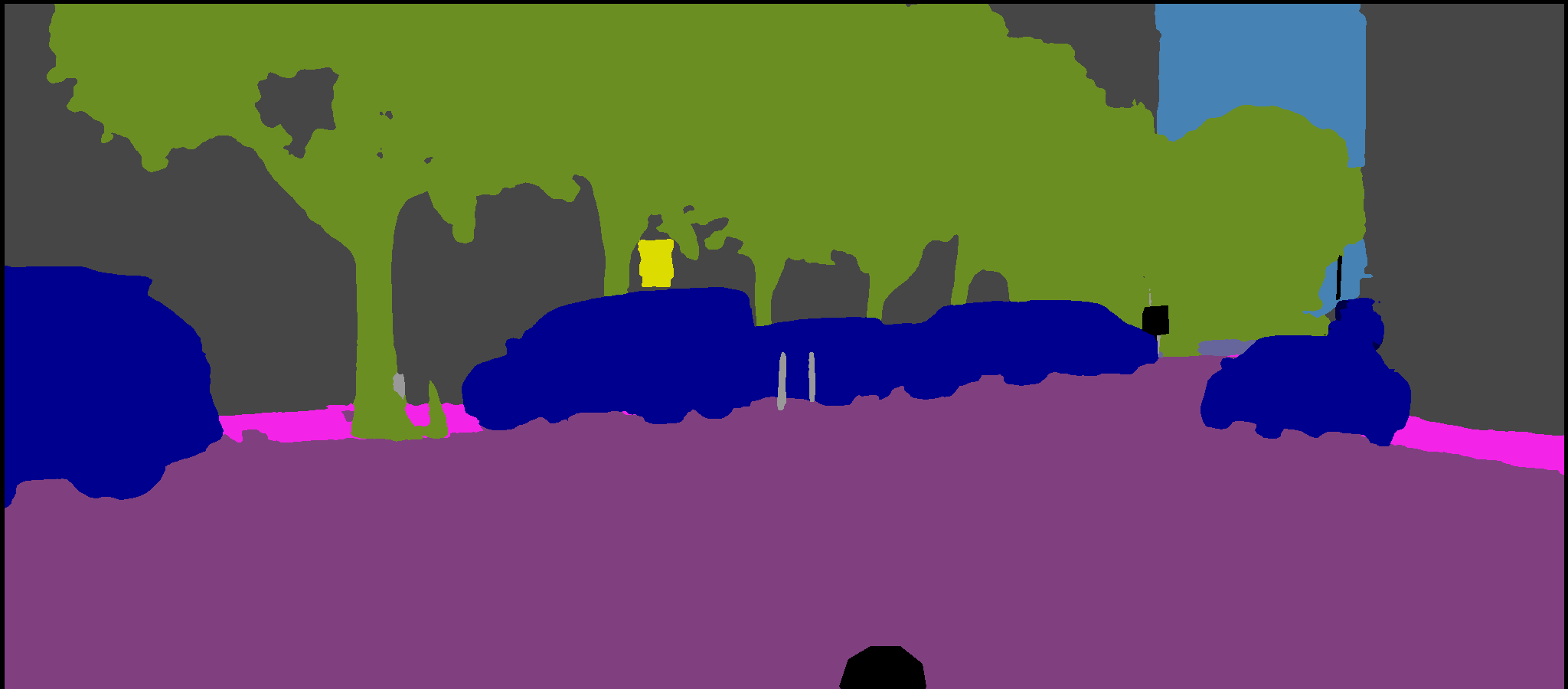} &
    \includegraphics[width=0.23\textwidth]{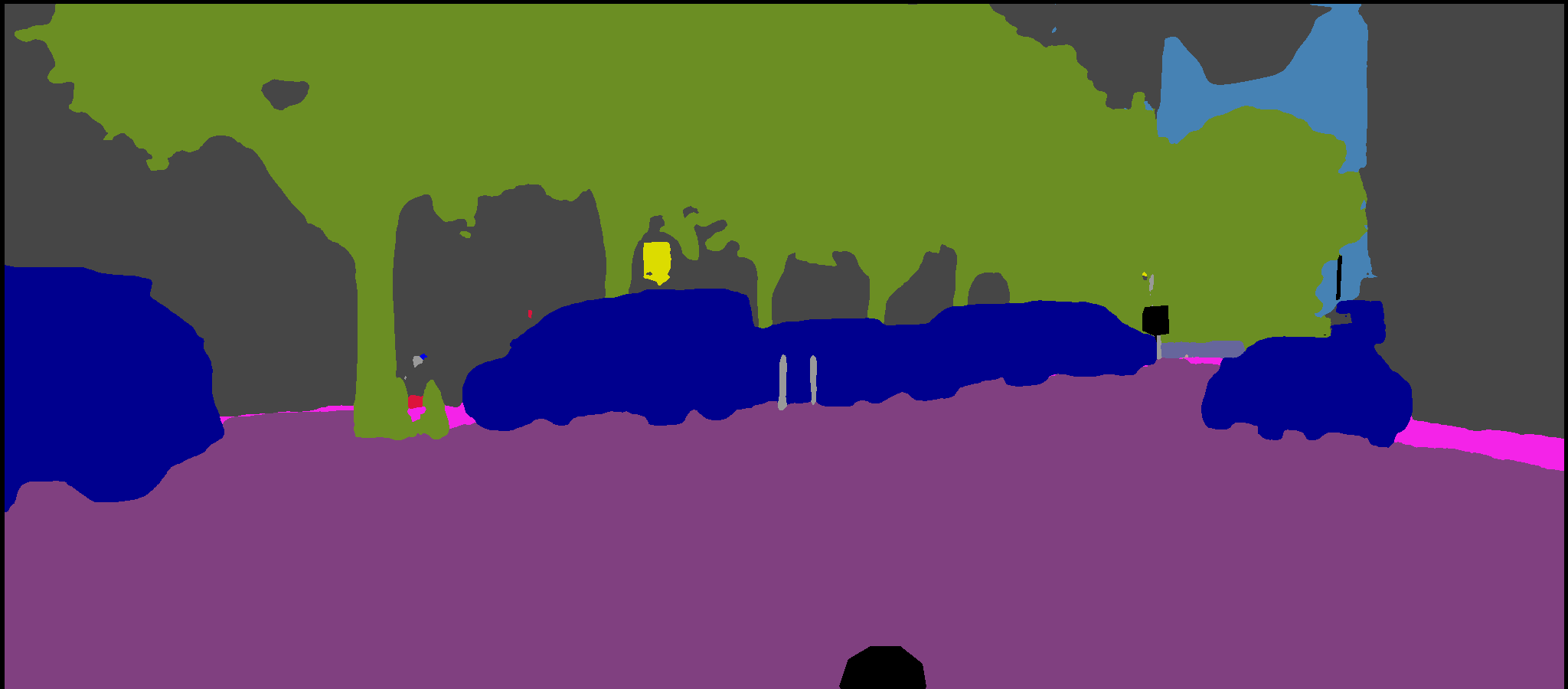} &
    \includegraphics[width=0.23\textwidth]{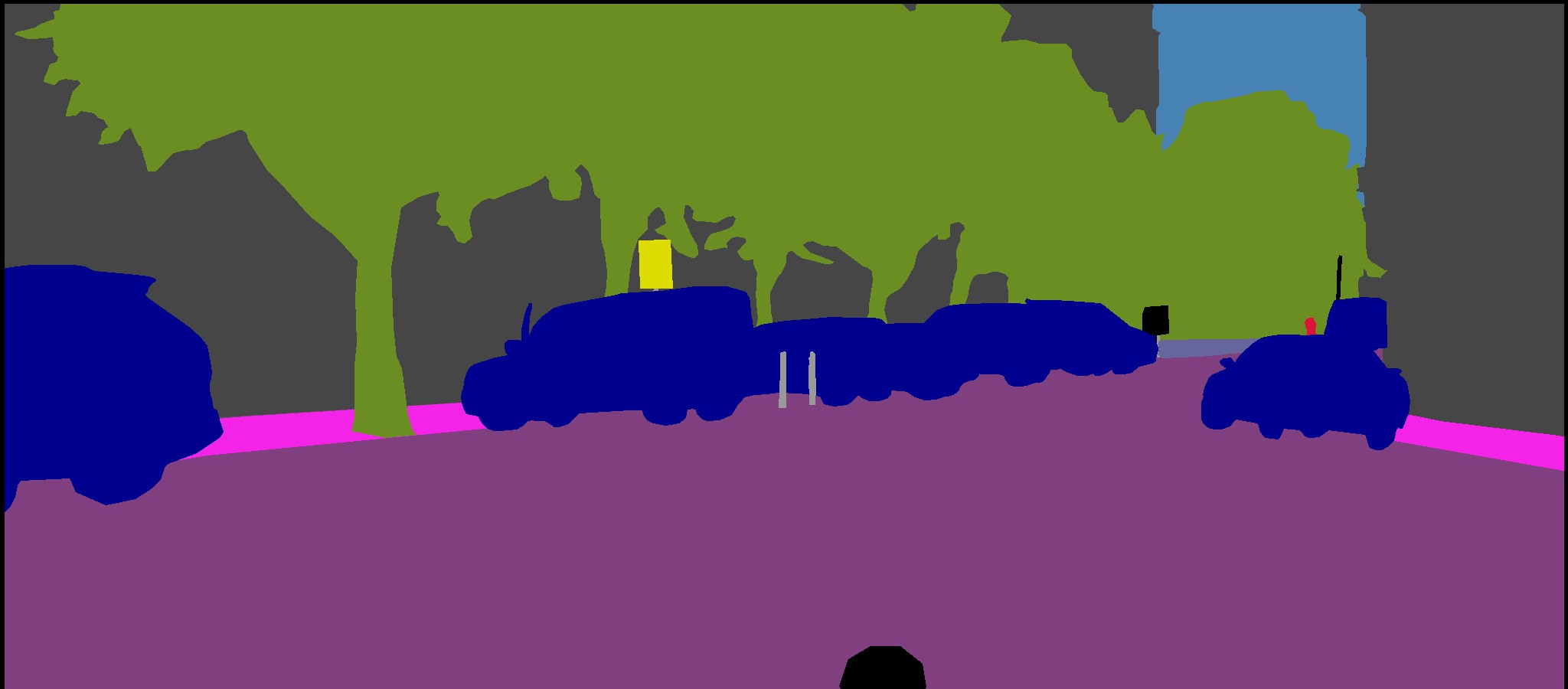} \\
    & \multicolumn{3}{c}{\small{Semantic Segmentation}} \\
    & \includegraphics[width=0.23\textwidth]{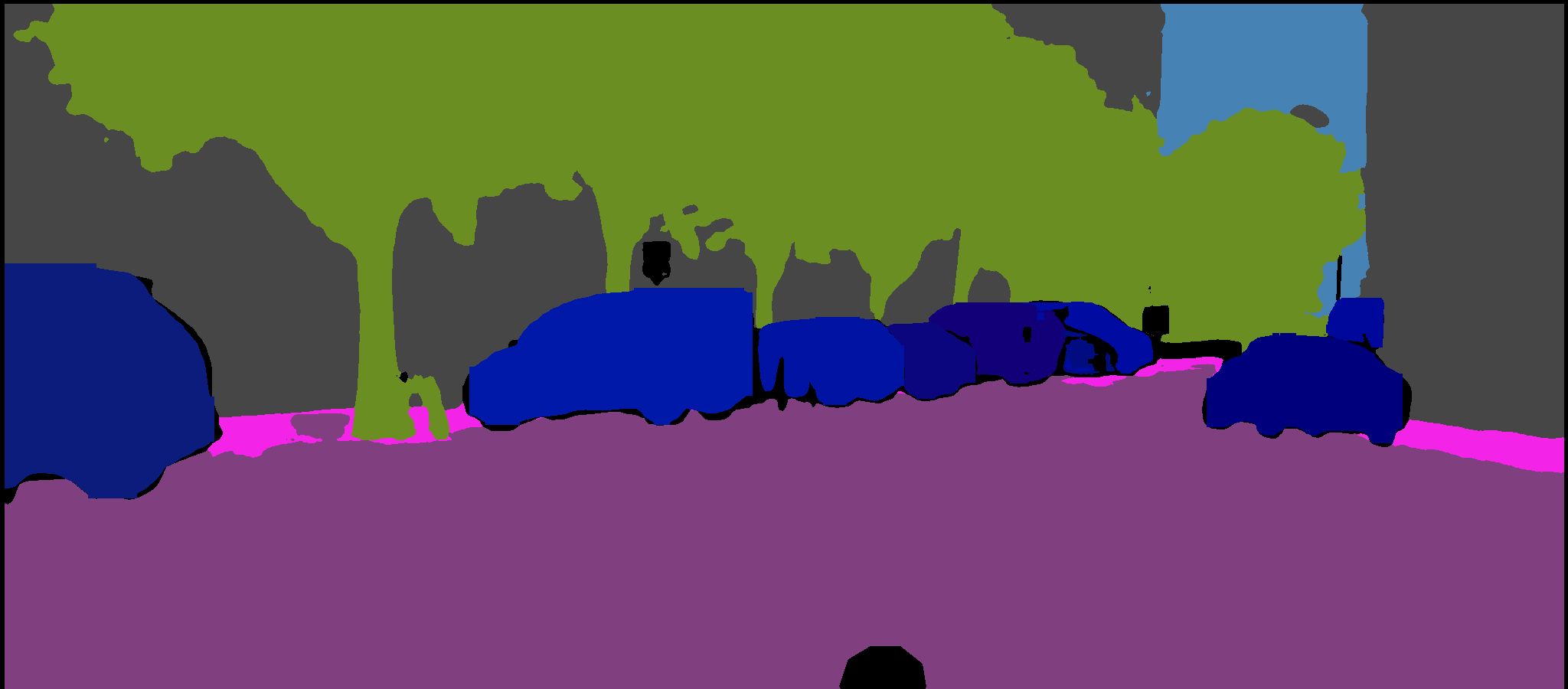} &
    \includegraphics[width=0.23\textwidth]{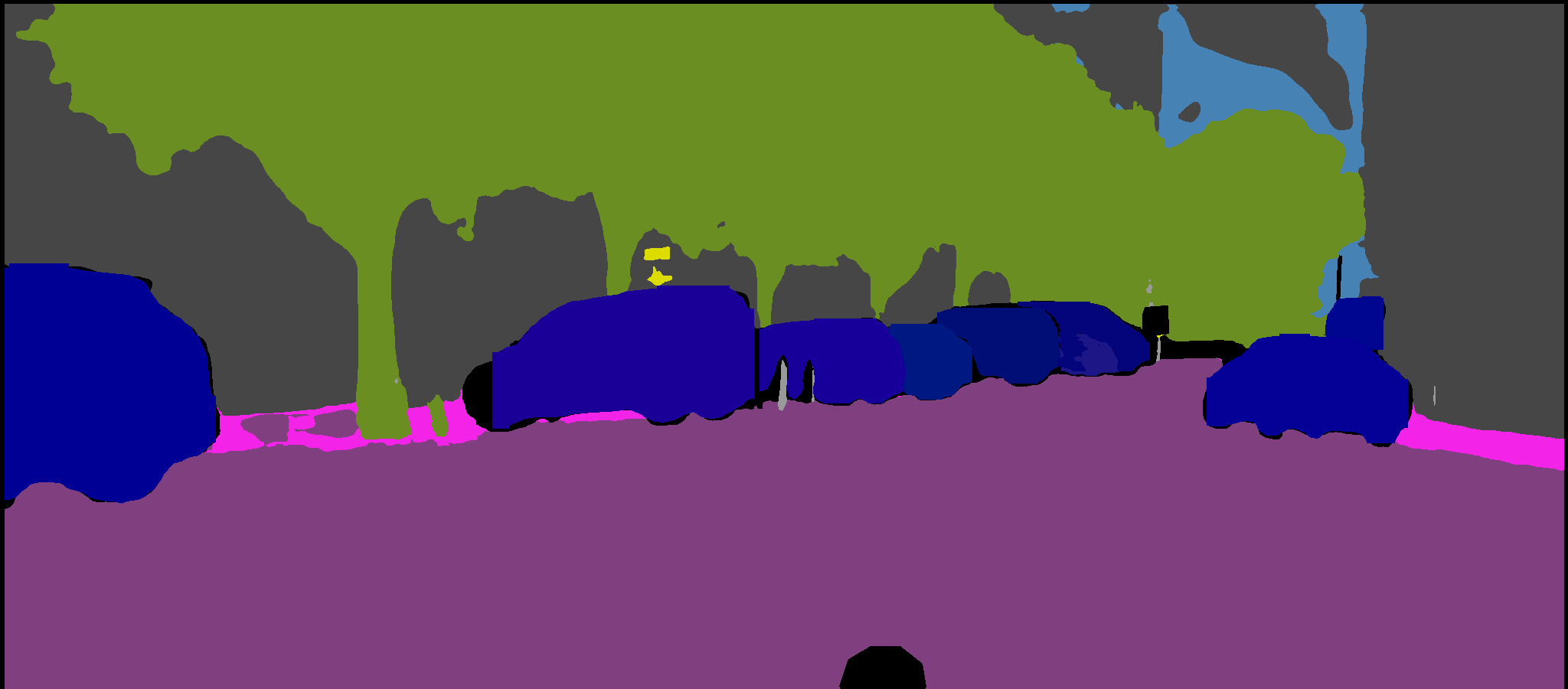} &
    \includegraphics[width=0.23\textwidth]{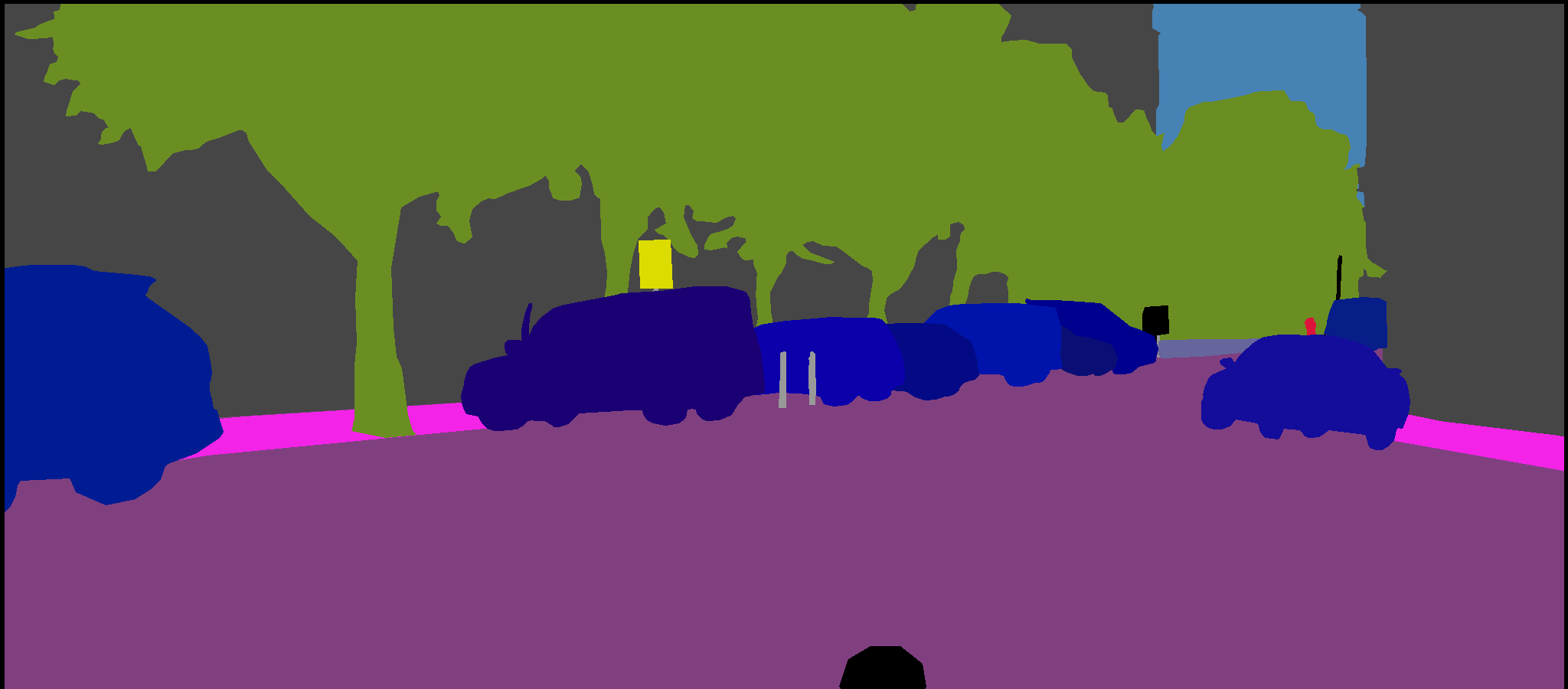} \\
    & \multicolumn{3}{c}{\small{Panoptic Segmentation}} \\
    & \includegraphics[width=0.23\textwidth]{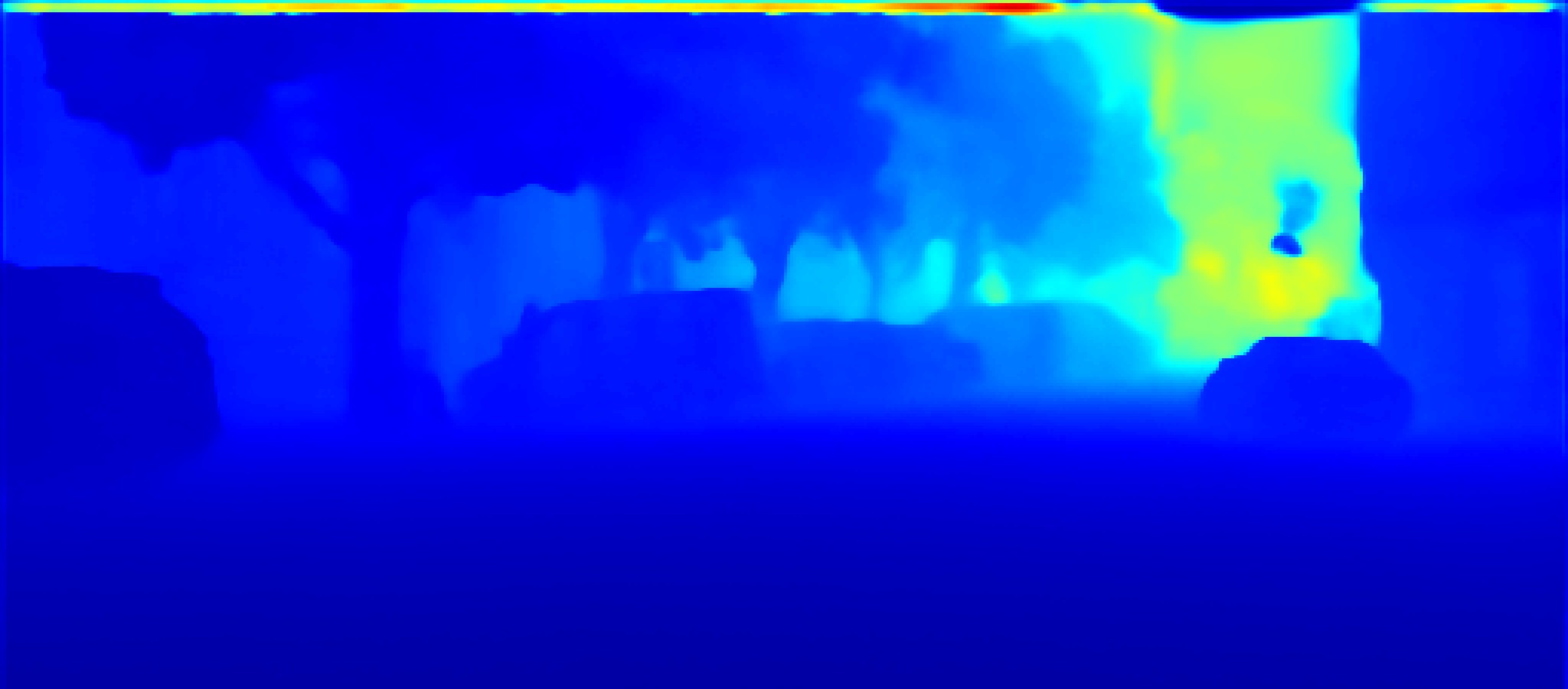} &
    \includegraphics[width=0.23\textwidth]{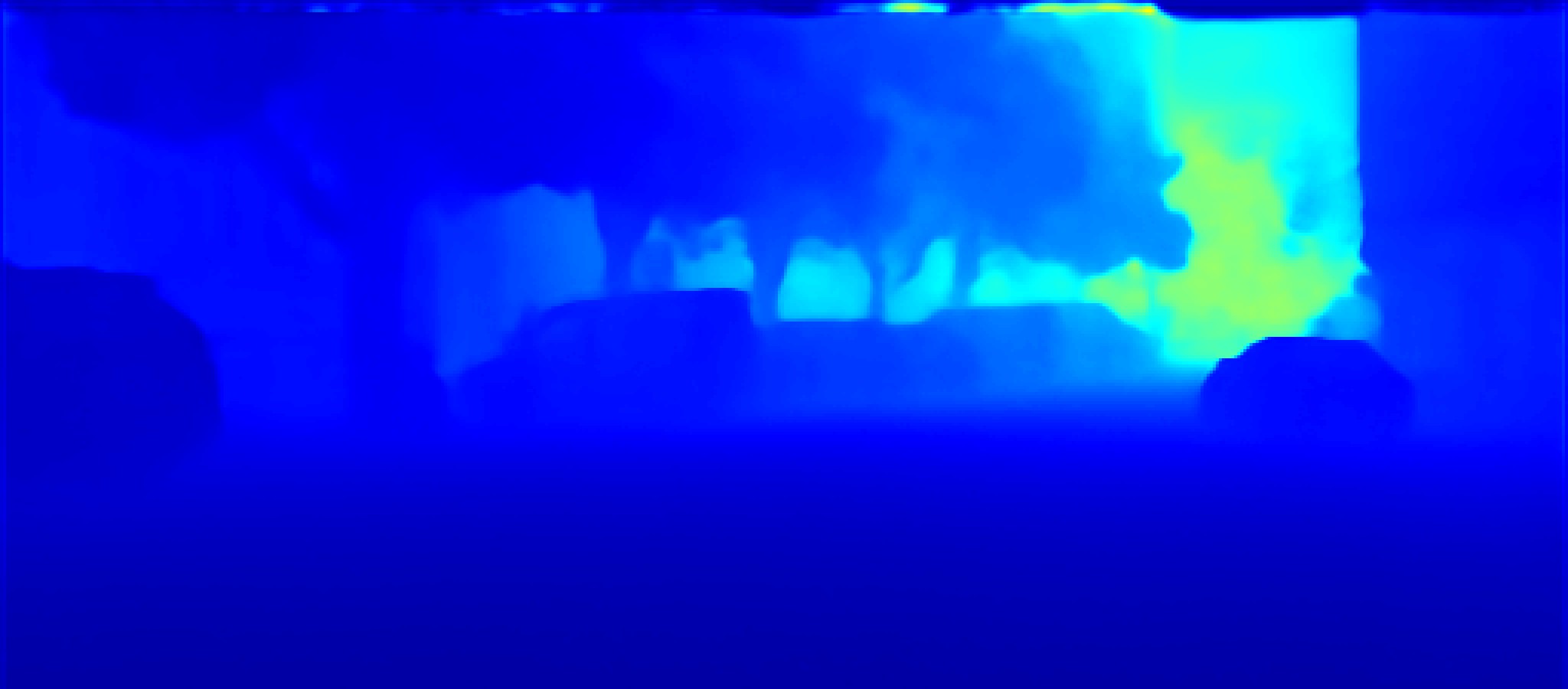} &
    \includegraphics[width=0.23\textwidth]{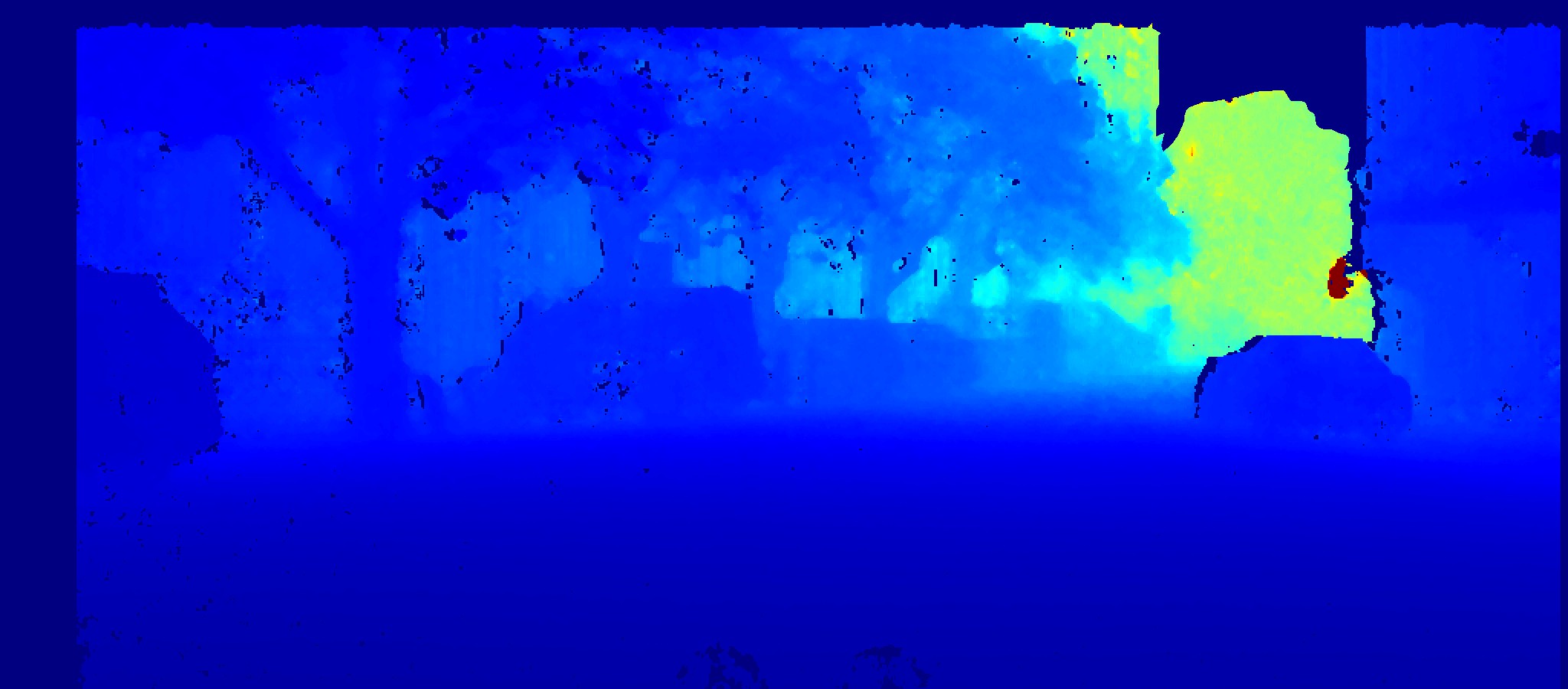} \\
    & \multicolumn{3}{c}{\small{Depth}} \\ \midrule
    \includegraphics[width=0.23\textwidth]{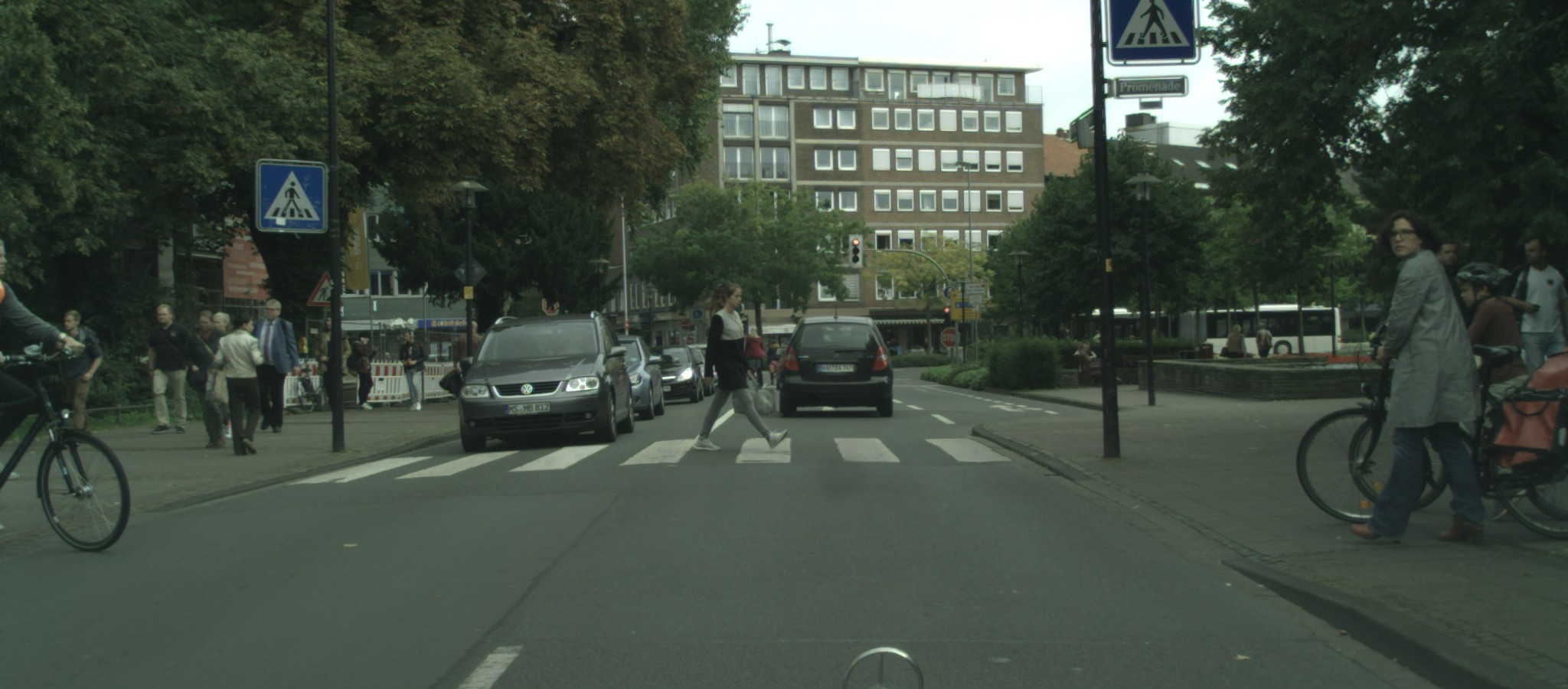} &
    \includegraphics[width=0.23\textwidth]{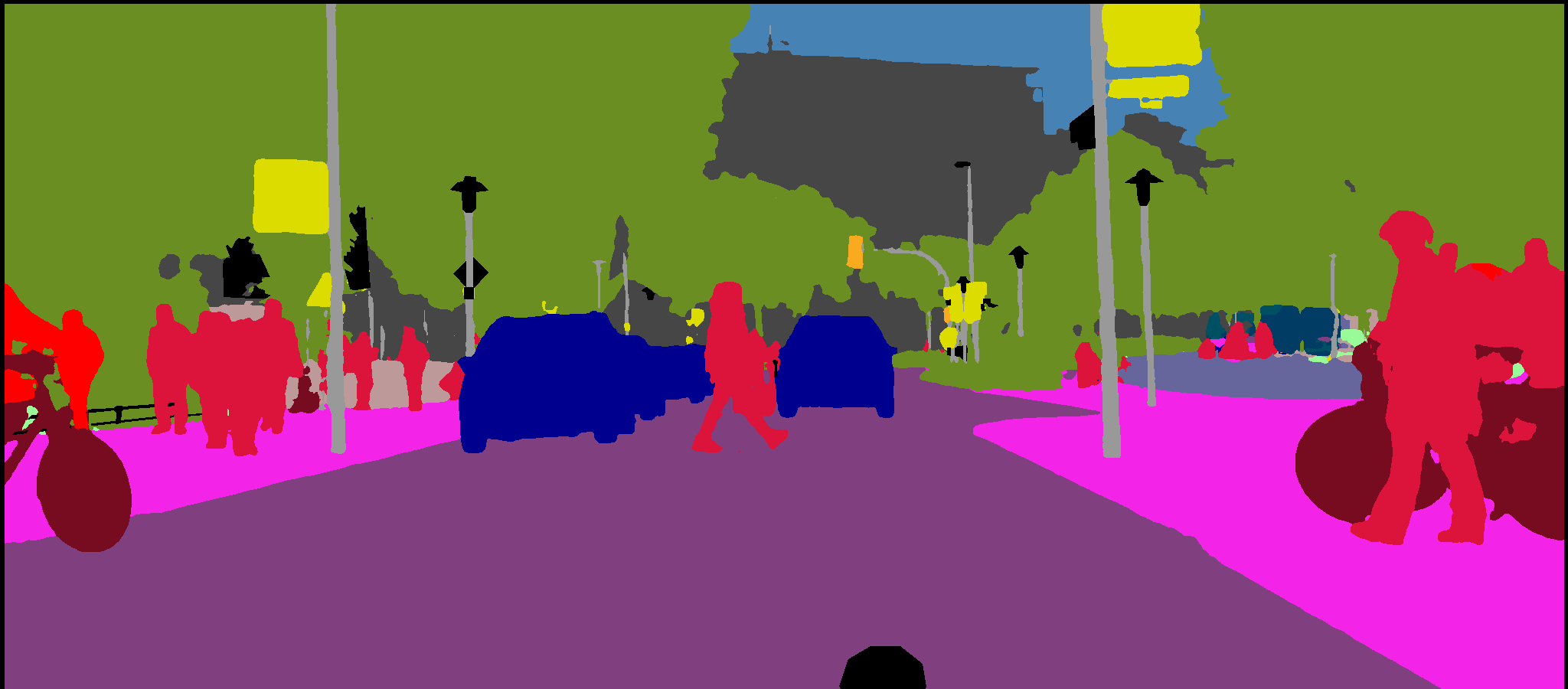} &
    \includegraphics[width=0.23\textwidth]{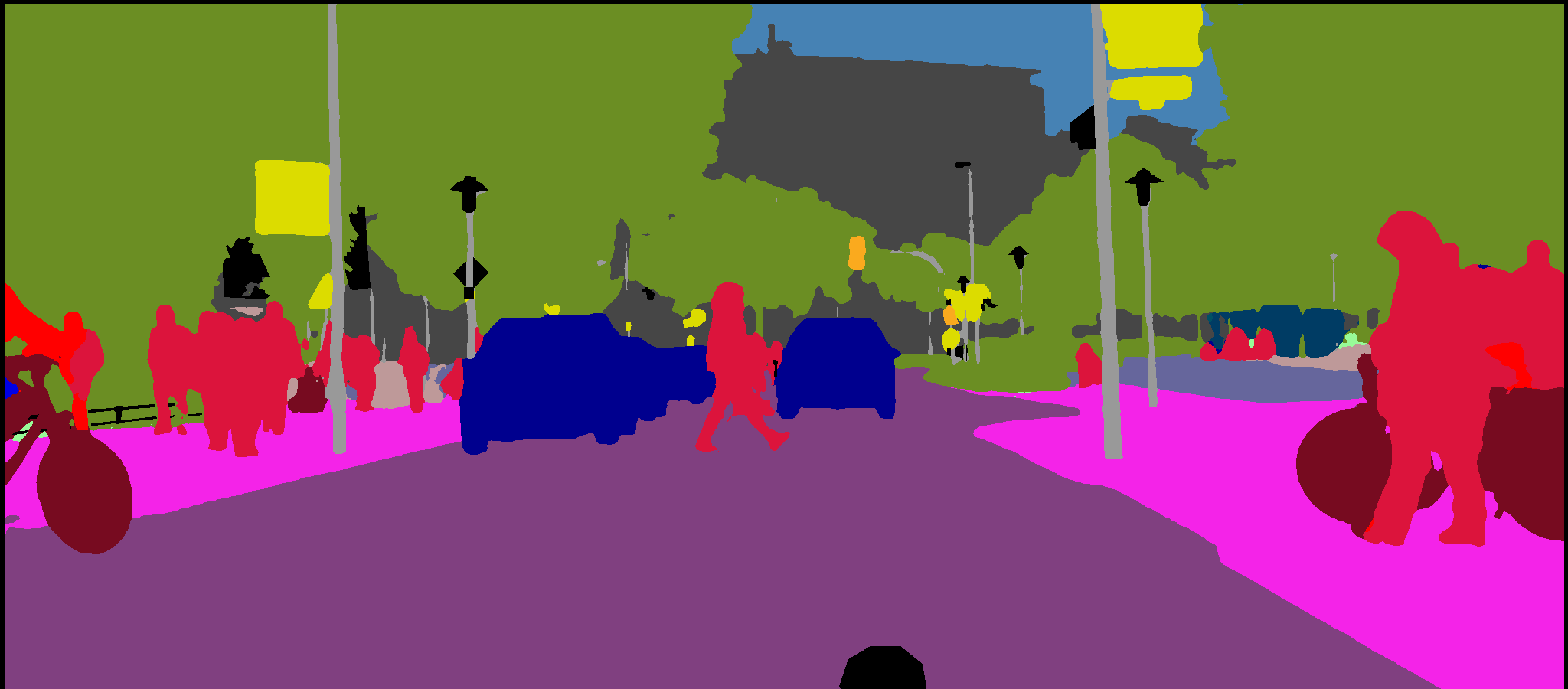} &
    \includegraphics[width=0.23\textwidth]{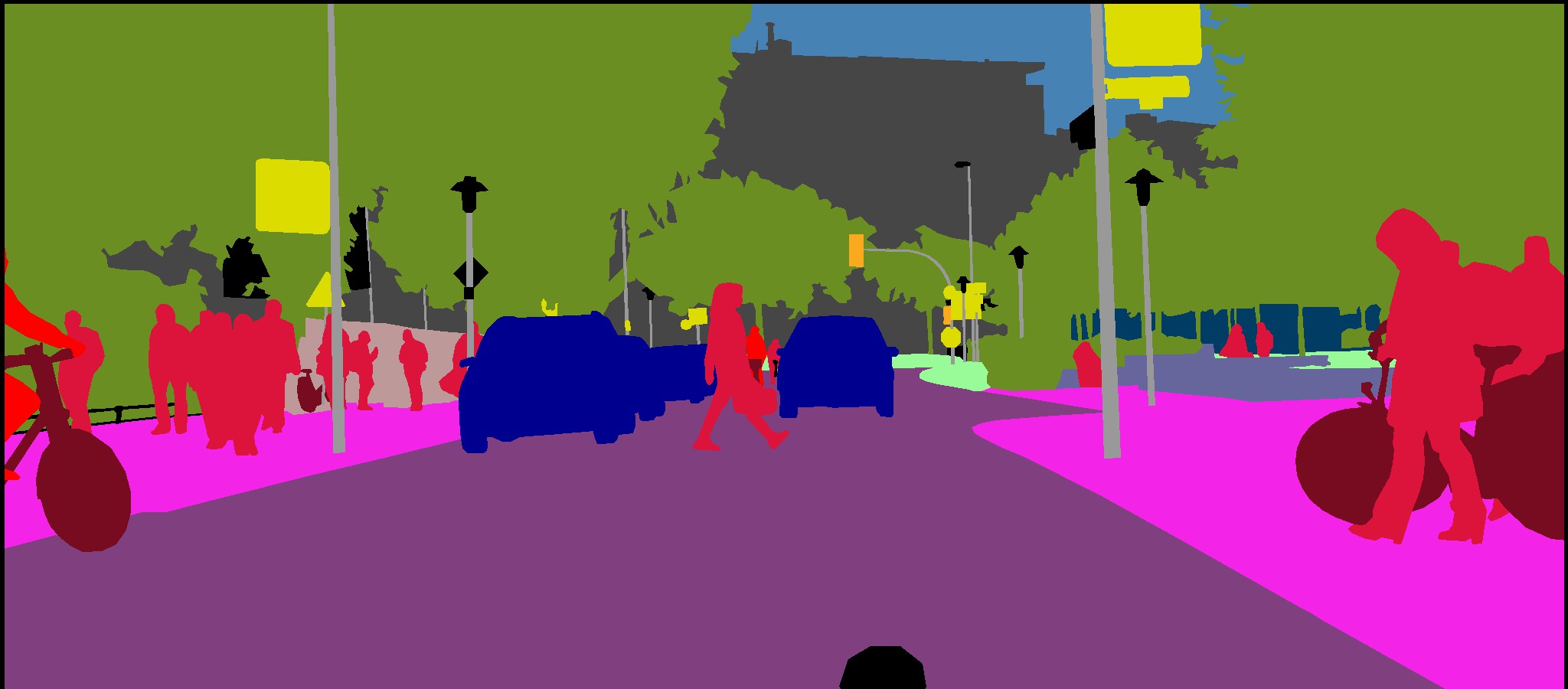} \\
    & \multicolumn{3}{c}{\small{Semantic Segmentation}} \\
    & \includegraphics[width=0.23\textwidth]{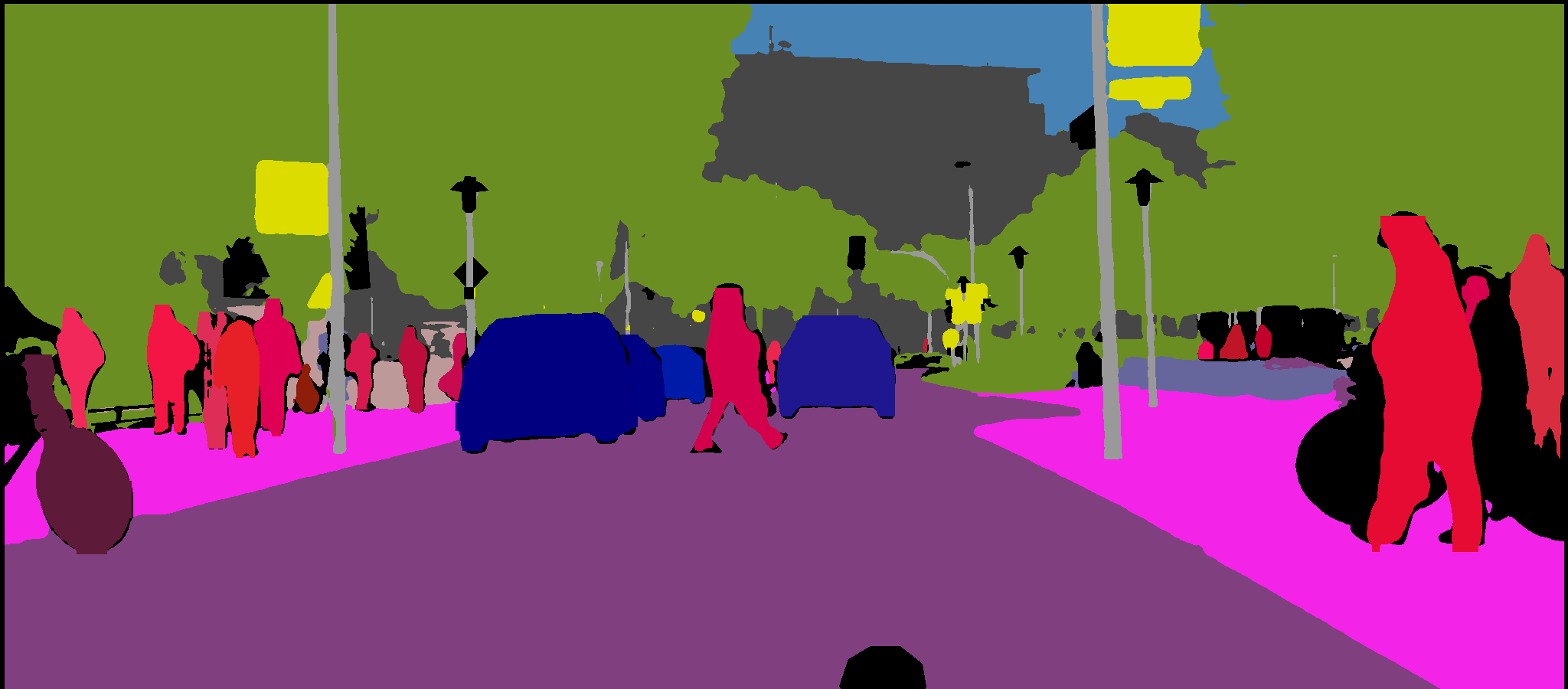} &
    \includegraphics[width=0.23\textwidth]{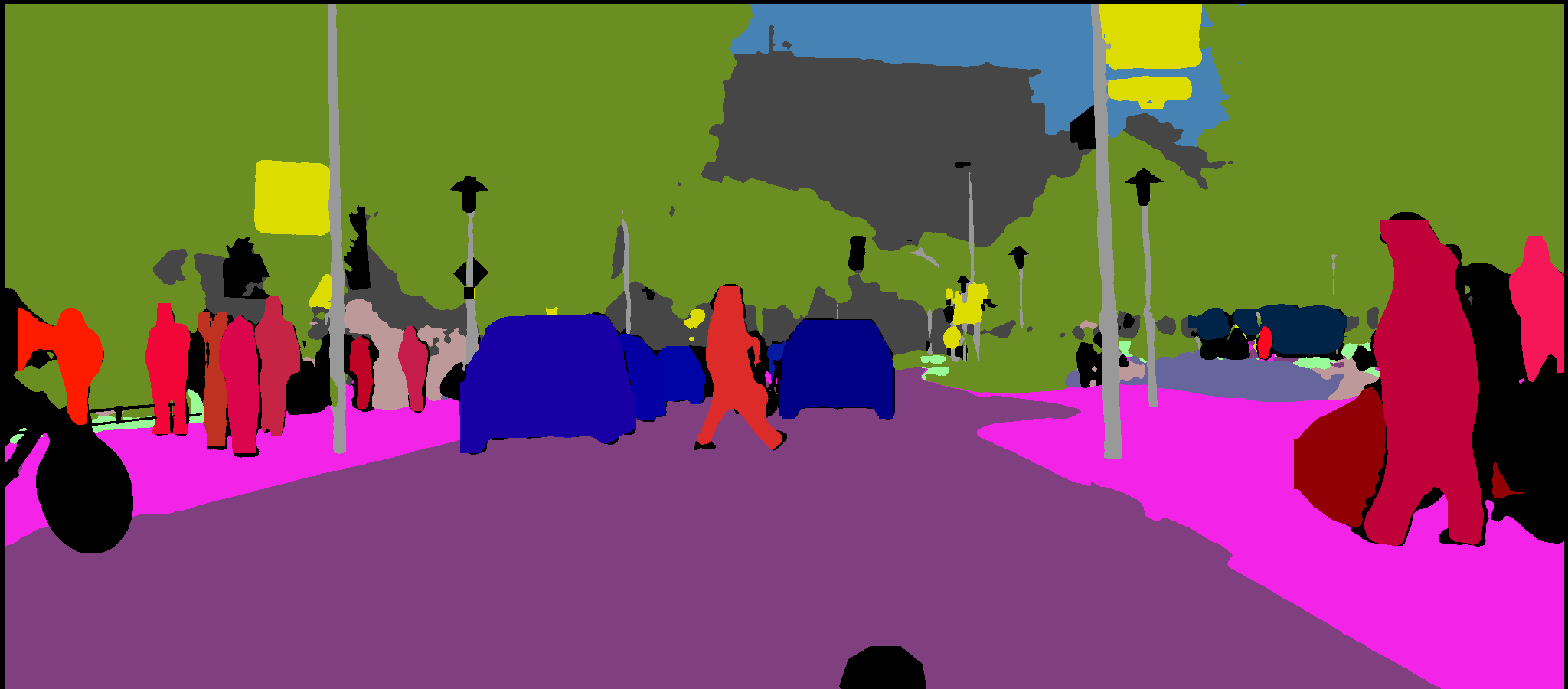} &
    \includegraphics[width=0.23\textwidth]{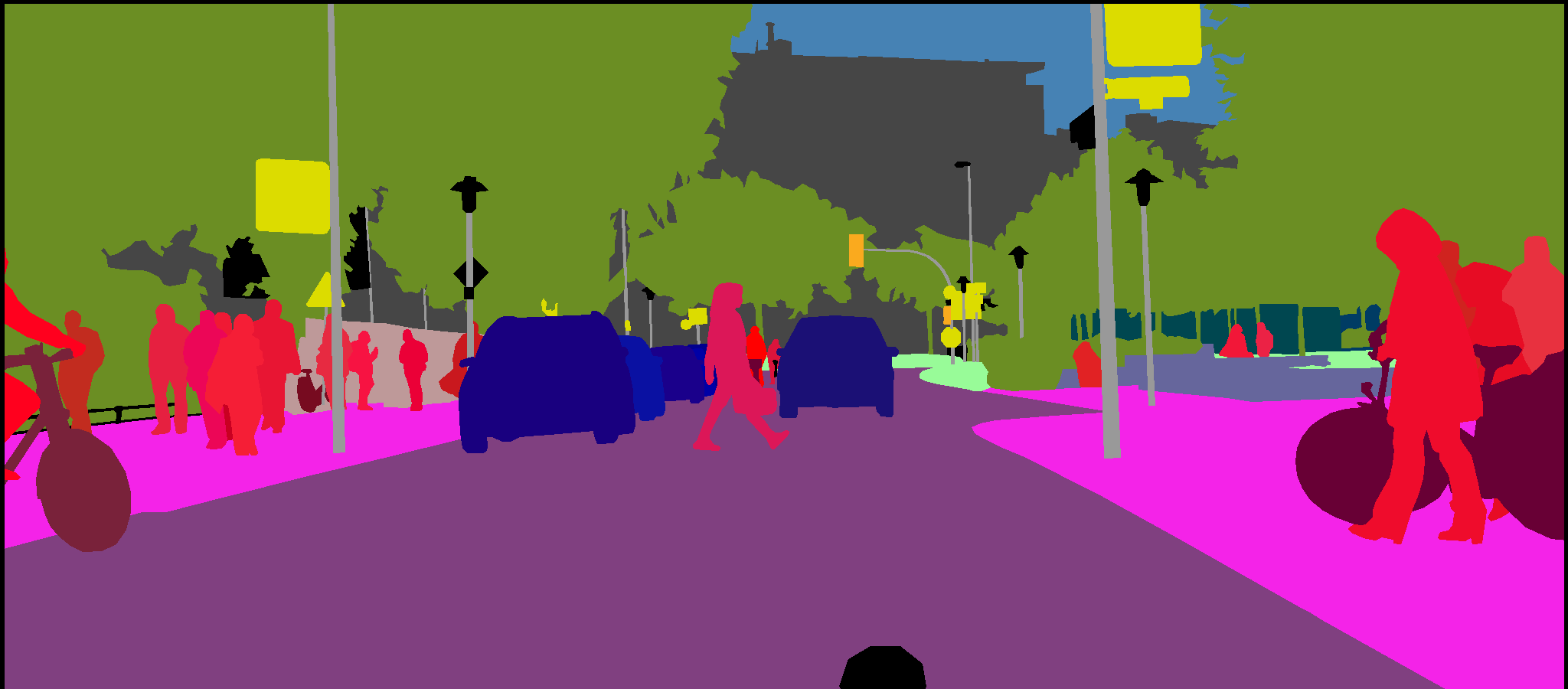} \\
    & \multicolumn{3}{c}{\small{Panoptic Segmentation}} \\
    & \includegraphics[width=0.23\textwidth]{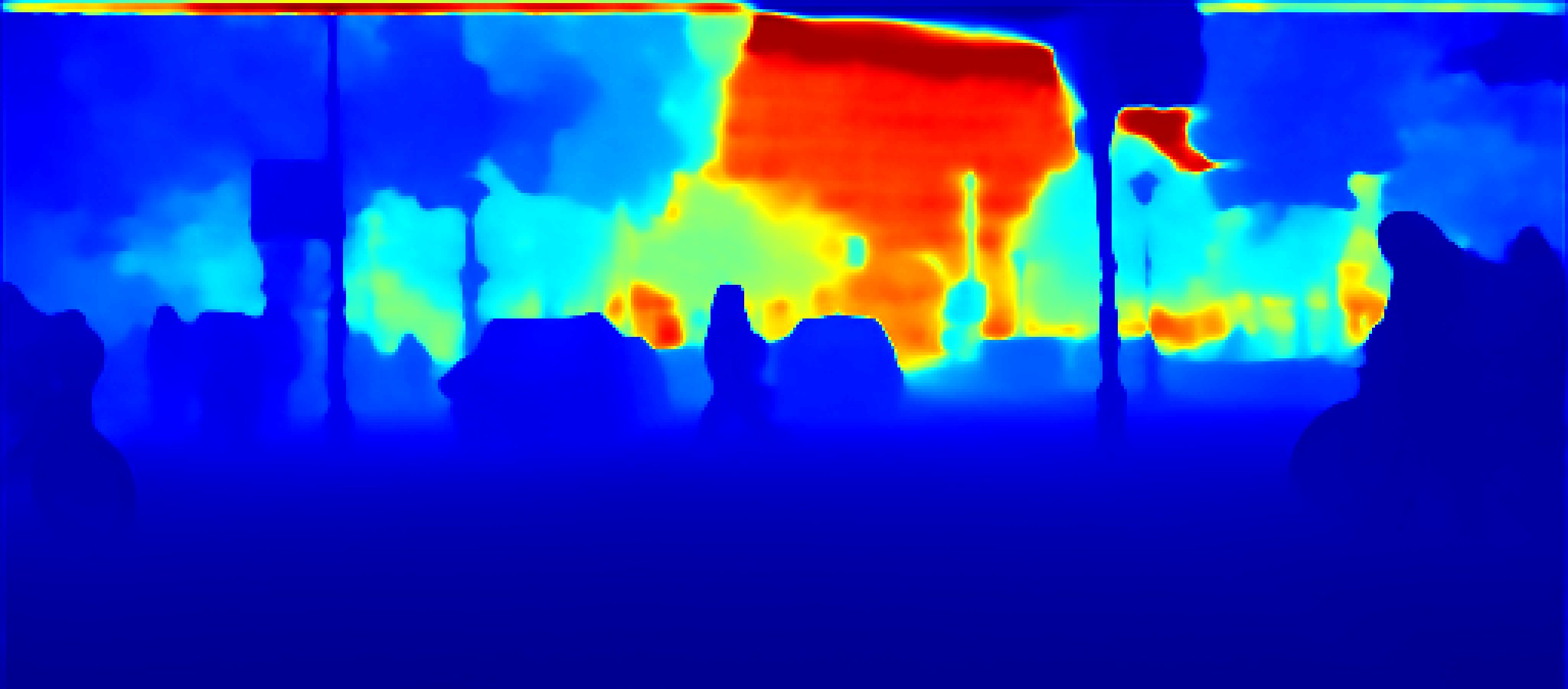} &
    \includegraphics[width=0.23\textwidth]{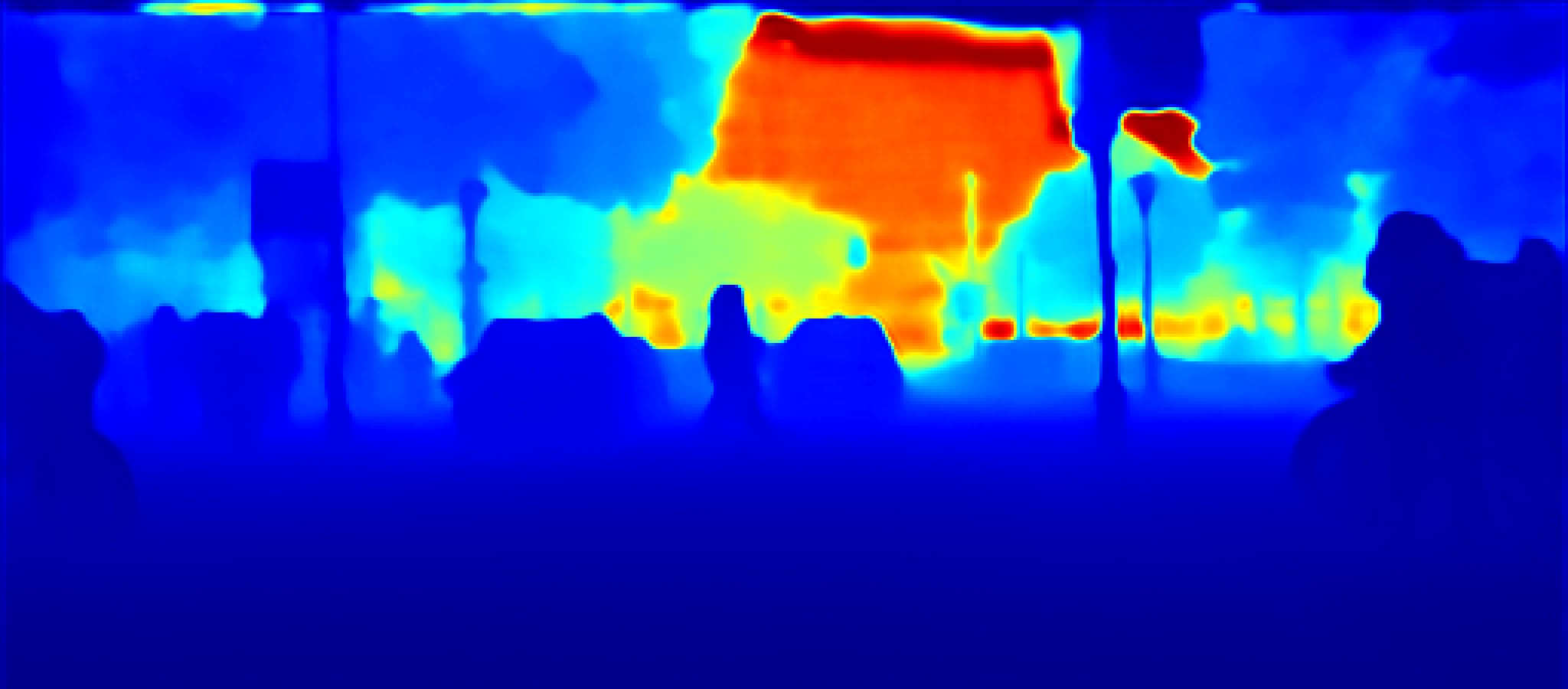} &
    \includegraphics[width=0.23\textwidth]{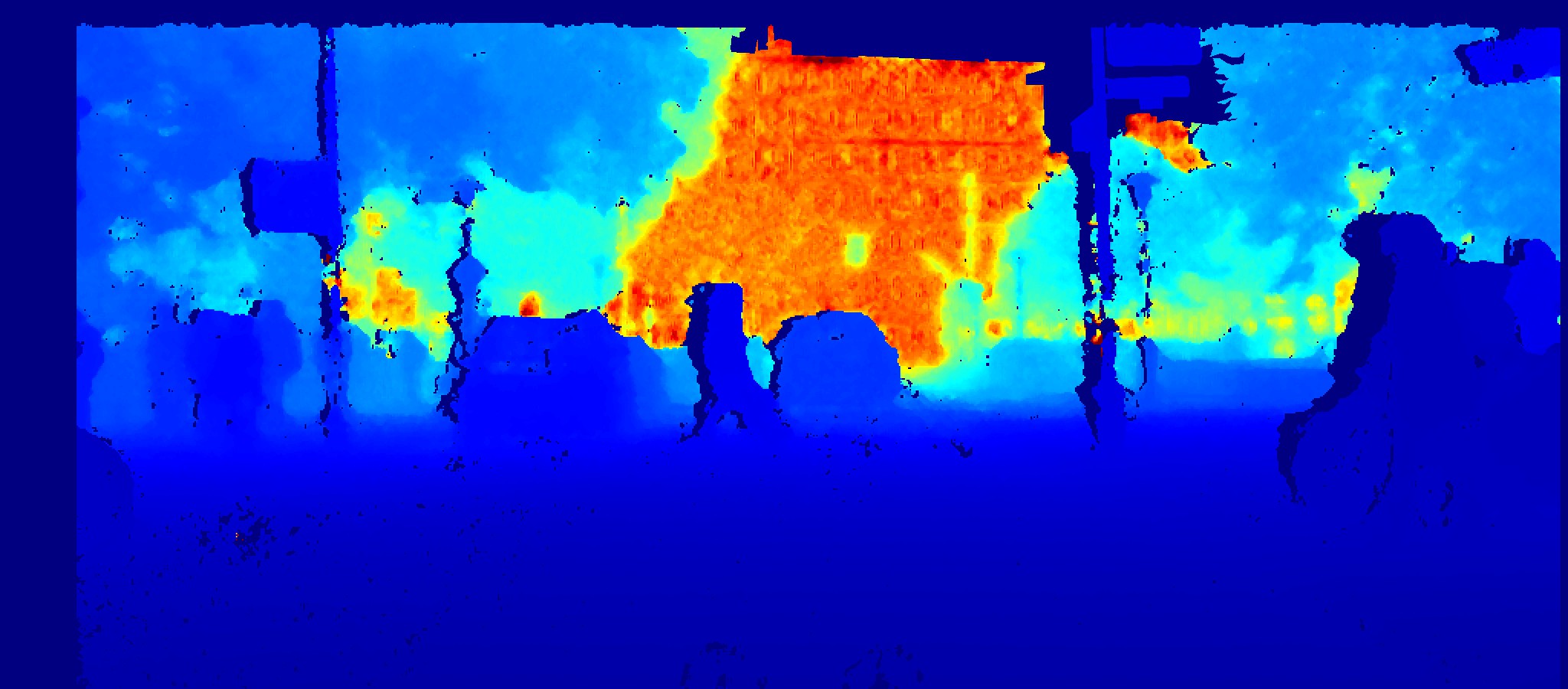} \\
    & \multicolumn{3}{c}{\small{Depth}} \\
    \bottomrule
    \end{tabular} 
  }
  \caption{Visual results on Cityscapes validation set.}\label{tab:cs_visual}
\end{figure}

\paragraph{Results.} Parameters and MAdds of comparison methods are presented in Table. \ref{tab:cs_com}.
Results and some visual examples on segmentation and depth prediction are shown in Table. \ref{tab:cs_res} and Fig. \ref{tab:cs_visual}, respectively. The priority of the proposed VarGNet v1 and v2 is proved by the above tables. VarGNet v1 and v2 are efficient and can perform equally well when compared with large networks.

\begin{table}
  \centering
  \caption{Details of comparison methods and ours on pixel level parsing tasks with input size 640$\times$360.}\label{tab:cs_com}
  \begin{tabular}{lccc}
  \toprule
  Method & Backbone & MAdds(G) & Params \\ \midrule
  SegNet\cite{badrinarayanan2017segnet} & VGG16 & 286.0 & 29.5M \\
  Enet\cite{paszke2016enet} & From scratch & 3.8 & 0.4M \\
  BiSeNet\cite{yu2018bisenet} & Xception39 & 2.9 & 5.8M \\
  BiSeNet\cite{yu2018bisenet} & Res18 & 10.8 & 49.0M \\ \midrule
  MobileNet v2 & - & 6.82 &	7.64M \\ \midrule
  VarGNet v1 & - & 6.16 & 13.23M \\
  VarGNet v2  & - & 2.76 & 7.41M \\
  % VarGNet v2 (w headpooling) & - & 2.64 & 7.40M   \\ 
  \bottomrule
  \end{tabular}
\end{table}

\begin{table}
  \centering
  \caption{Results on Cityscapes validation set.}\label{tab:cs_res}
  \begin{tabular}{c}
    (a) Semantic Segmentation (image size 2048$\times$1024)\\
    \begin{tabular}{lcc}
    \toprule
    Method & Backbone & Mean IoU(\%)  \\ \midrule
    BiSeNet\cite{yu2018bisenet} & Xception39 & 69.0 \\
    BiSeNet\cite{yu2018bisenet} & Res18 & 74.8 \\ \midrule
    MobileNet v2 & - & 64.8 \\ \midrule
    VarGNet v1 & - & 76.6 \\
    VarGNet v2 & - & 74.2 \\
    % VarGNet v2(w headpooling) & - & 73.3 \\ 
    \bottomrule
    \end{tabular} 
  \end{tabular}   
  % &
  \begin{tabular}{c}
    (b) Depth \\
    \begin{tabular}{lcccc}
      \toprule
      Method & AbsRel & SqlRel & RMSE & RMSE Log \\ \midrule
      MobileNet v2 & 0.167 & 3.22 & 15.46 & 0.553 \\ \midrule
      VarGNet v1 & 0.092 & 1.327 & 8.864 & 0.163 \\
      % VarGNet v2(wo headpooling) & 0.094 & 1.371 & 8.59 & 0.163 \\
      VarGNet v2 & 0.096 & 1.404 & 8.85 & 0.168 \\ 
      \bottomrule
    \end{tabular}   
  \end{tabular}   
  \begin{tabular}{c}
    (c) Panoptic Segmentation (MAdds calculated with 2048$\times$1024 input size.) \\
    \begin{tabular}{lcccccc}
      \toprule
      Method & Backbone & MAdds & PQ & PQ(Things) & PQ(Stuff) & Mean IoU(\%)  \\ \midrule
      PFPnet\cite{Panoptic} & resnet101 & 533G & 58.1 & 52 & 62.5 & 75.1 \\
      VarGNet v1 & - & 104G & 57.1 & 50 & 62.3 & 73.4 \\
      VarGNet v2 & - & 68G & 54.5 & 45.1 & 59.8 & 71.4 \\
      \bottomrule
    \end{tabular}   
  \end{tabular}  
  \begin{tabular}{c}
    (d) Panoptic Segmentation + Depth (MAdds calculated with 2048$\times$1024 input size.) \\
    \begin{tabular}{lccccccc}
      \toprule
      Method & MAdds & PQ & PQ(Things) & PQ(Stuff) & Mean IoU(\%) & AbsRel & RMSE \\\midrule
      VarGNet v1  & 109G & 56 & 48.8 & 61.3 & 71 & 0.1 & 9.2 \\
      VarGNet v2  & 70G & 53.9 & 46.2 & 59.5 & 70.5 & 0.116 & 10.06 \\
      \bottomrule
    \end{tabular}   
  \end{tabular}  
\end{table}

\subsection{KITTI}
\paragraph{Traning setup.} For single image depth prediction and stereo tasks on KITTI dataset \cite{geiger2013vision}, we present the performance of our VarGNet based models. A U-Net style architecture (\ref{fig:netskitti}) is employed in the experiments.
All the depth models are trained on KITTI RAW datasets, We test on 697 images from 29 scenes split by Eigen et al. \cite{eigen2014depth}, and train on about 23488 images from the remaining 32 scenes. All the experiment results are evaluated with the depth ranging from 0m to 80m and 0m to 50m. The evaluation metrics are the same as previous works. 
All the stereo models are trained on KITTI RAW datasets, We test on test set split by Eigen et al. \cite{eigen2014depth}, and train set of KITTI15. The evaluation metrics for stereo are EPE and D1.
During training, standard SGD Optimizer is used, and the momentum set to 0.9. The standard weight decay is set to 0.0001 for resnet18 and resnet50, and 0.00004 for others. The iteration number is set to 300 epochs. The initial learning rate is 0.001, and learning rate decay 0.1 at [120, 180, 240] epoch. We use 4 GPU to train models, and the batch size is set to 24.

\begin{figure}
  \centering
  \subfloat[Input Image]{\includegraphics[width=0.3\textwidth]{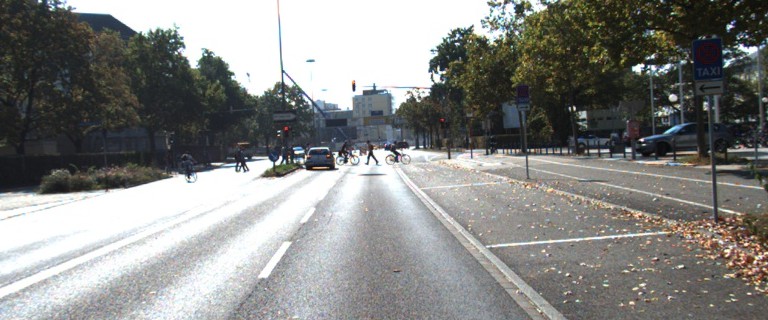}} \quad
  \subfloat[GT]{\includegraphics[width=0.3\textwidth]{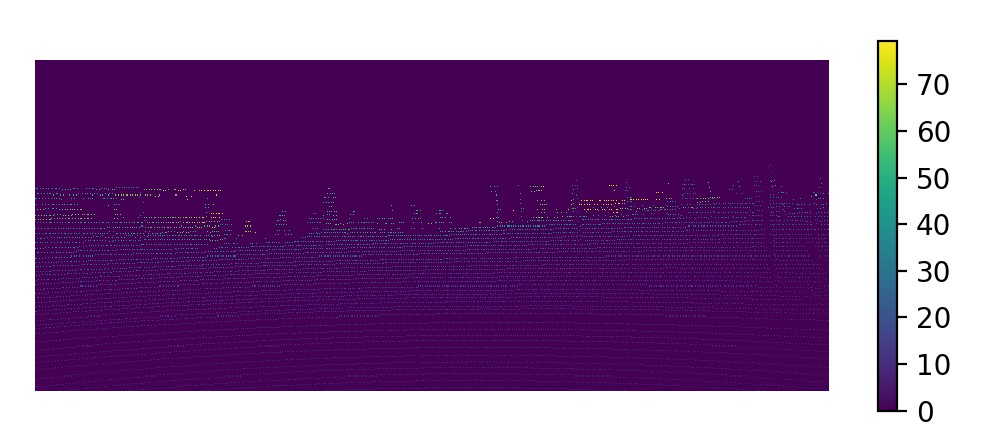}} \\
  \subfloat[MobileNet v2 1.0]{\includegraphics[width=0.3\textwidth]{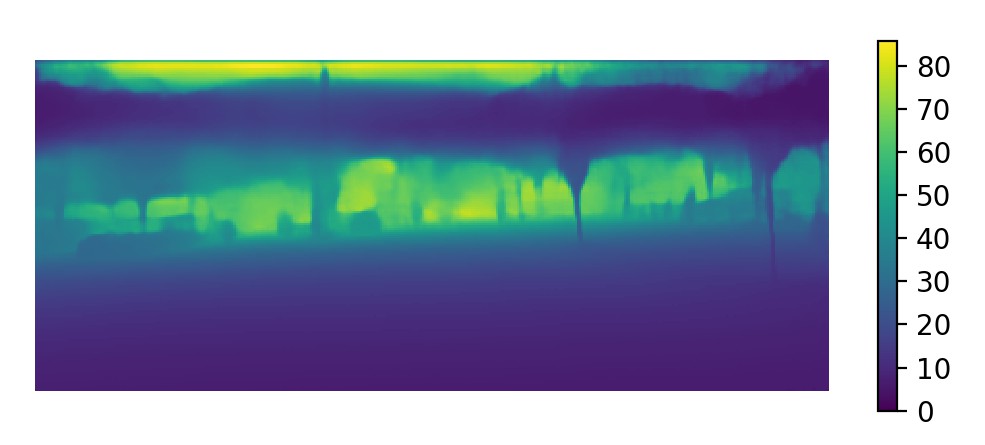}} \quad
  \subfloat[MobileNet v2 0.5]{\includegraphics[width=0.3\textwidth]{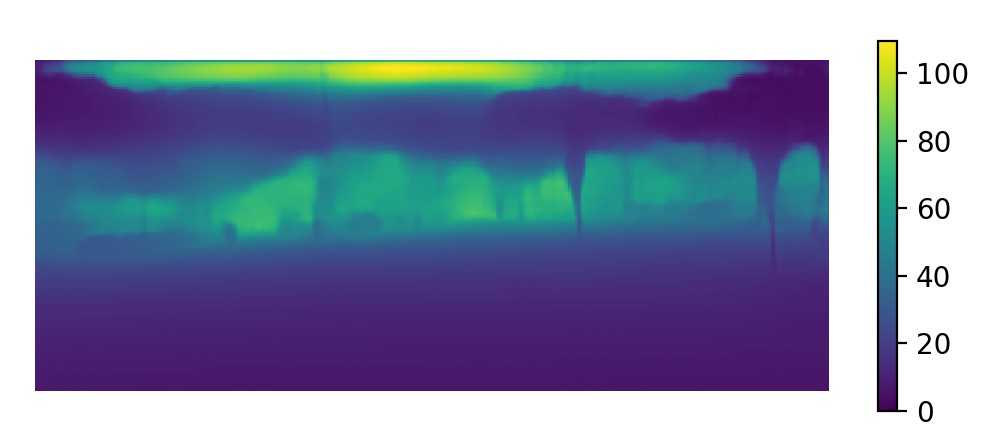}} \quad
  \subfloat[MobileNet v2 0.25]{\includegraphics[width=0.3\textwidth]{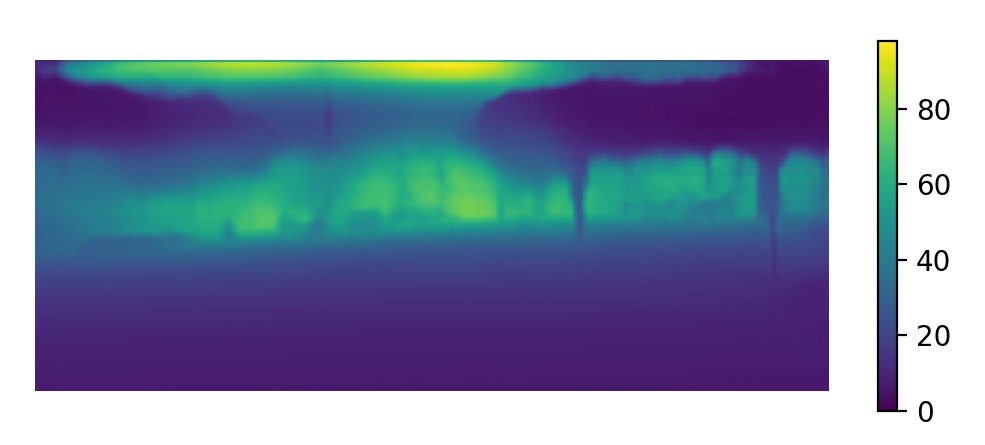}} \\
  \subfloat[ResNet 18]{\includegraphics[width=0.3\textwidth]{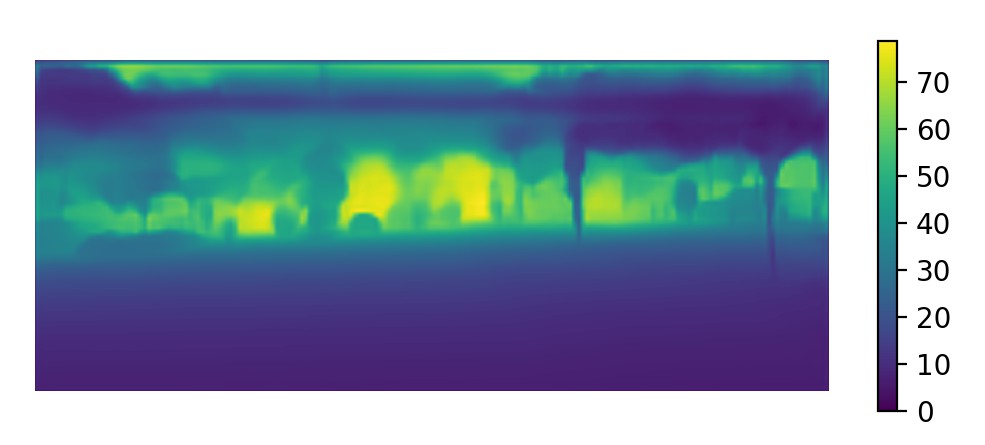}} \quad
  \subfloat[ResNet 50]{\includegraphics[width=0.3\textwidth]{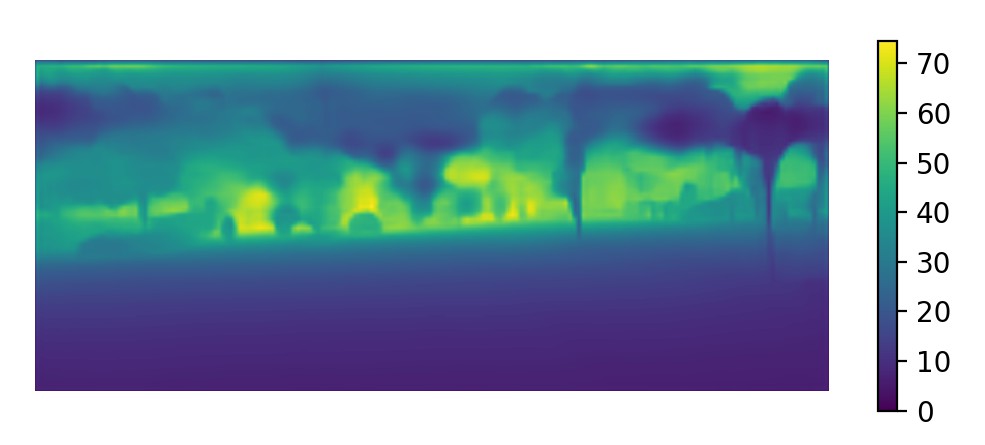}} \\
  \subfloat[VarGNet v1 1.0]{\includegraphics[width=0.3\textwidth]{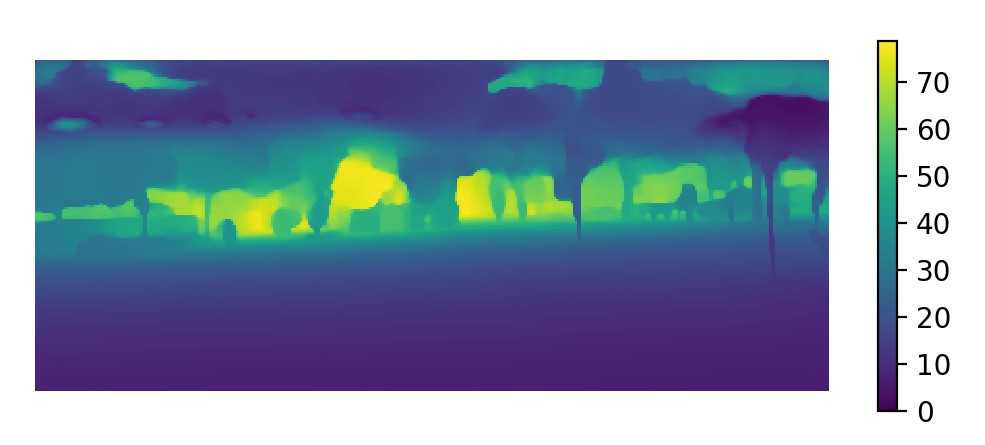}} \quad
  \subfloat[VarGNet v1 0.5]{\includegraphics[width=0.3\textwidth]{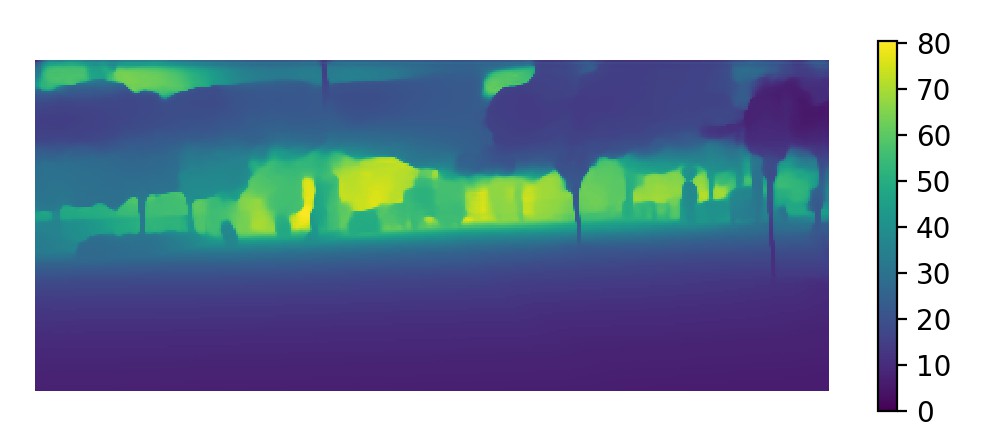}} \quad
  \subfloat[VarGNet v1 0.25]{\includegraphics[width=0.3\textwidth]{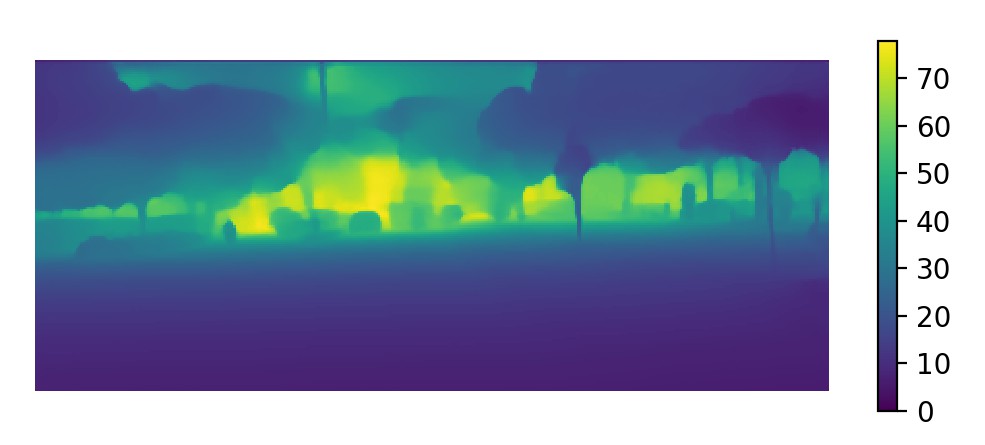}} \\
  \subfloat[VarGNet v2 1.0]{\includegraphics[width=0.3\textwidth]{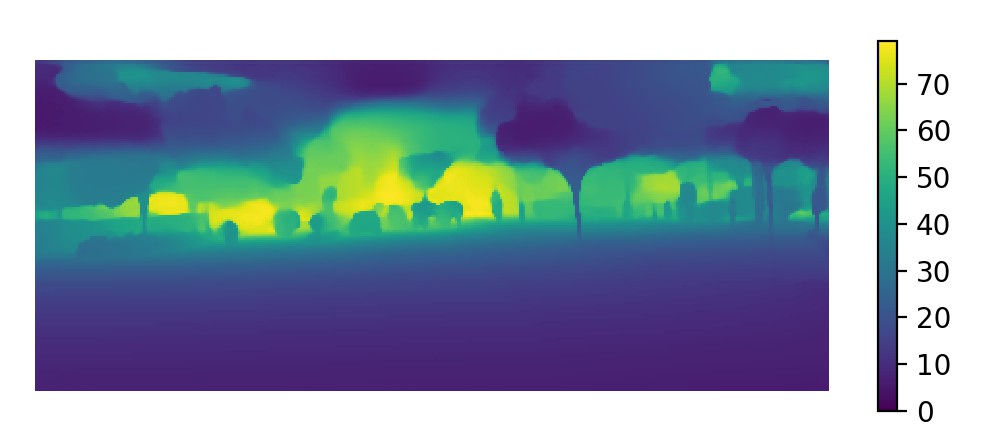}} \quad
  \subfloat[VarGNet v2 0.5]{\includegraphics[width=0.3\textwidth]{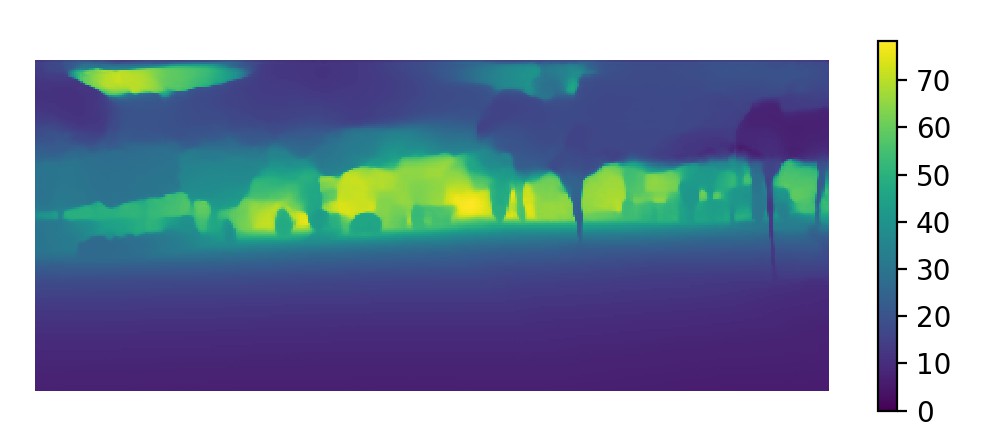}} \quad
  \subfloat[VarGNet v2 0.25]{\includegraphics[width=0.3\textwidth]{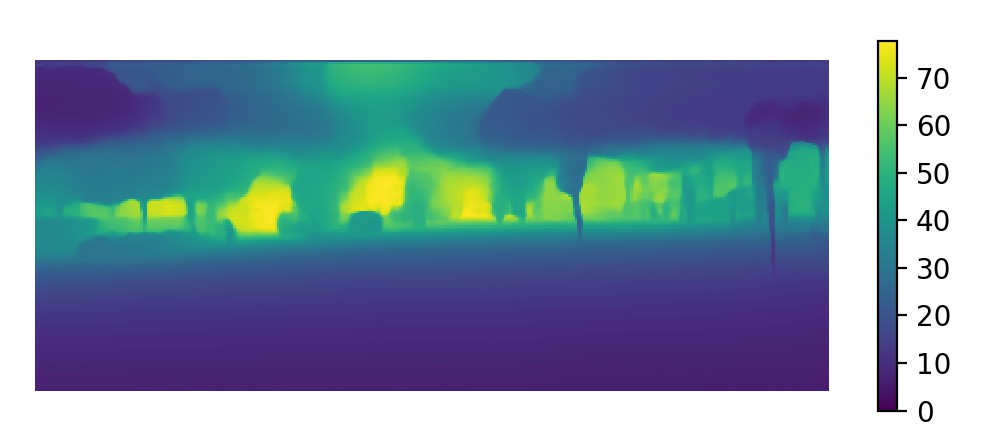}} 
  \caption{Visualization of depth results on KITTI RAW.}\label{fig:kittidepth}  
\end{figure}

\begin{figure}
  \centering
  \subfloat[Left Input Image]{\includegraphics[width=0.3\textwidth]{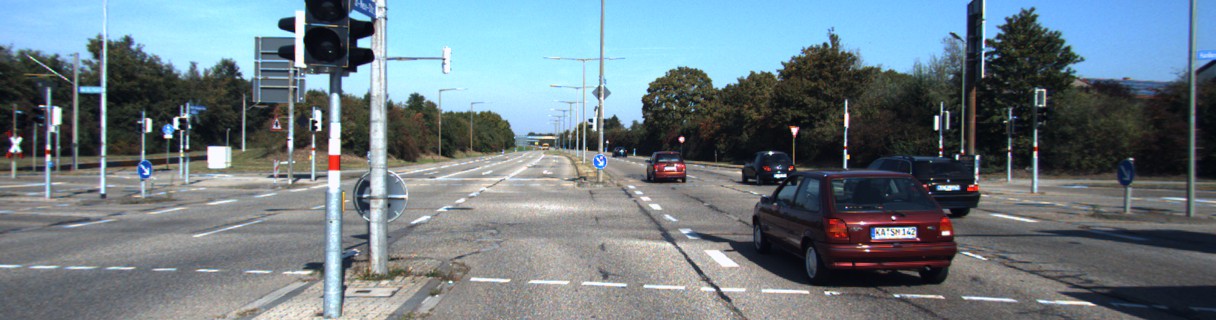}} \quad
  \subfloat[Right Input Image]{\includegraphics[width=0.3\textwidth]{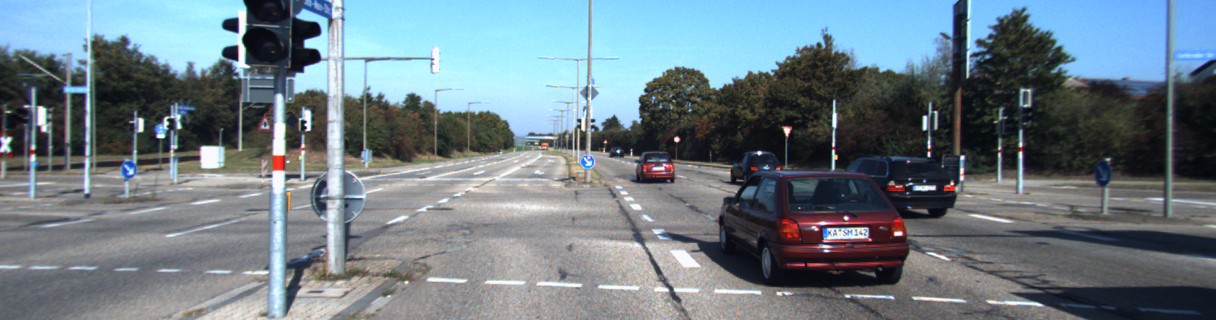}} \quad
  \subfloat[GT]{\includegraphics[width=0.3\textwidth]{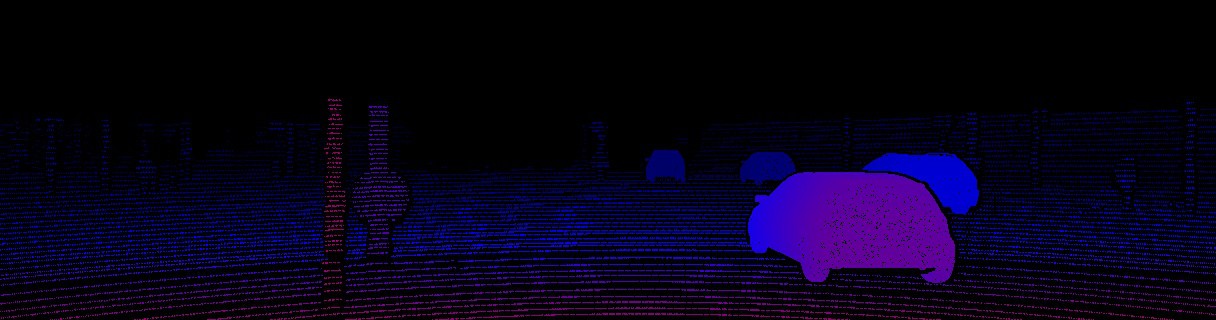}} \\
  \subfloat[MobileNet v2 1.0]{\includegraphics[width=0.3\textwidth]{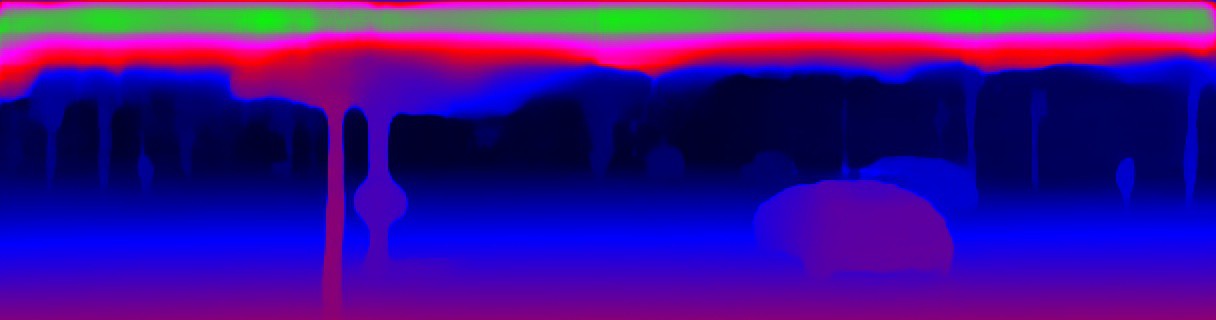}} \quad
  \subfloat[MobileNet v2 0.5]{\includegraphics[width=0.3\textwidth]{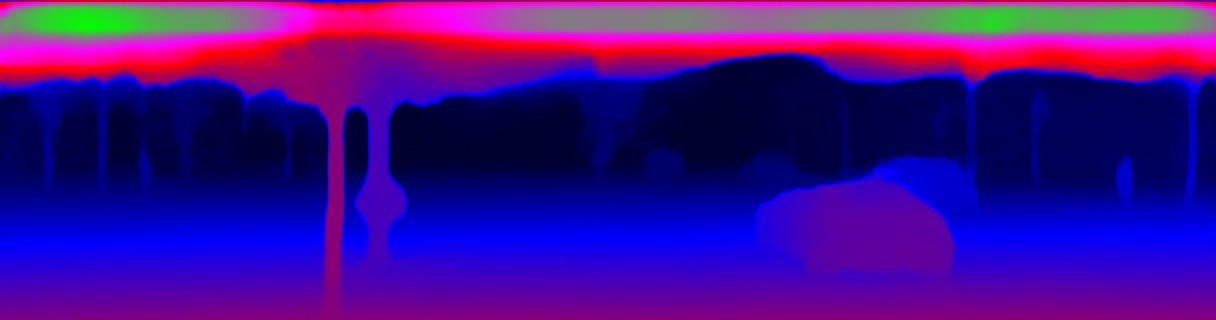}} \quad
  \subfloat[MobileNet v2 0.25]{\includegraphics[width=0.3\textwidth]{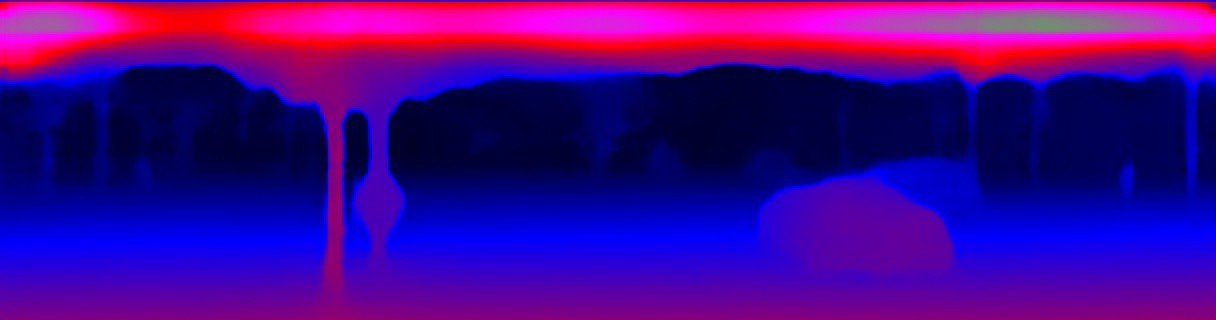}} \\
  \subfloat[ResNet 18]{\includegraphics[width=0.3\textwidth]{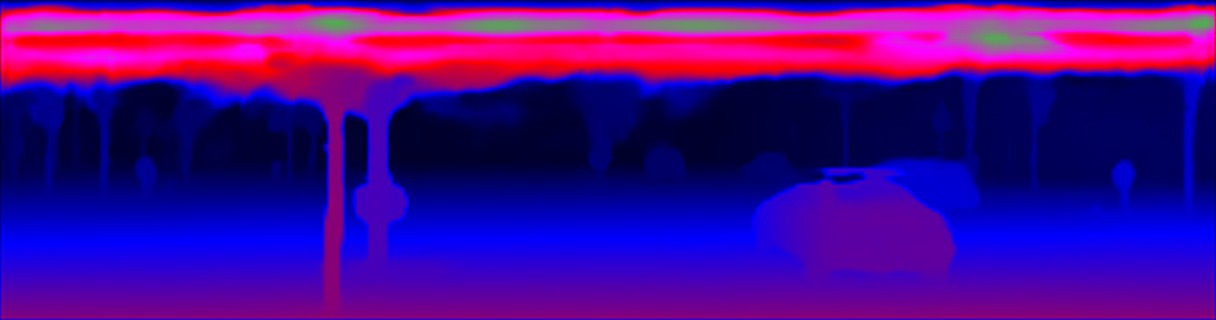}} \quad
  \subfloat[ResNet 50]{\includegraphics[width=0.3\textwidth]{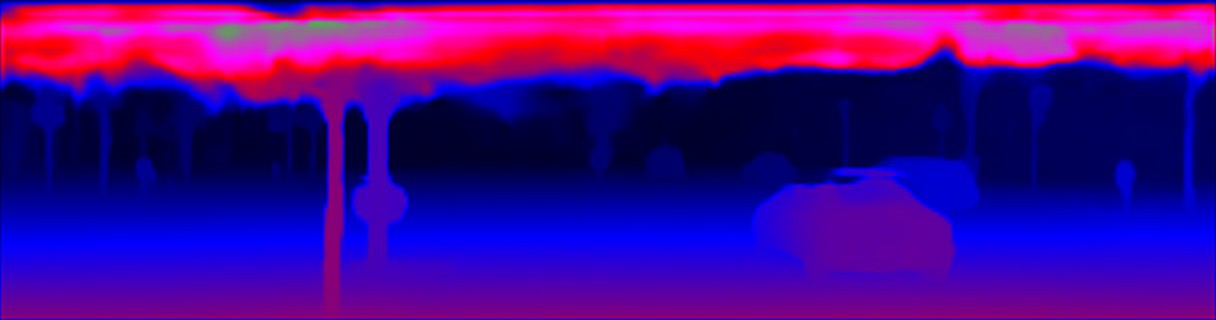}} \\
  \subfloat[VarGNet v1 1.0]{\includegraphics[width=0.3\textwidth]{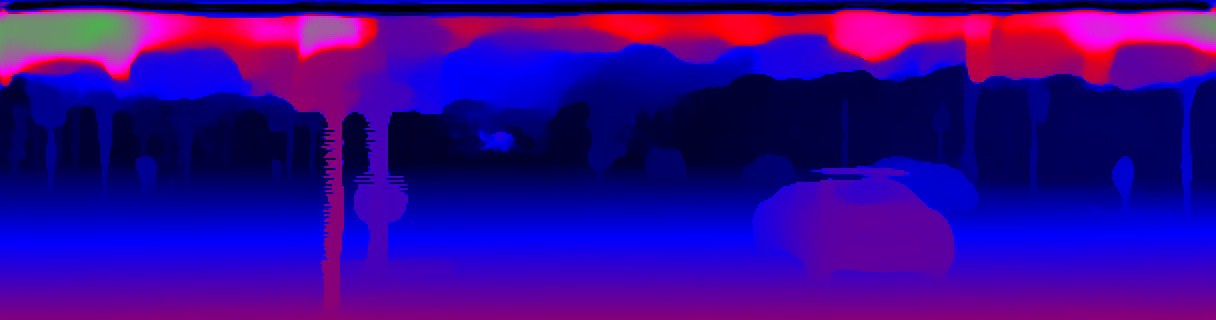}} \quad
  \subfloat[VarGNet v1 0.5]{\includegraphics[width=0.3\textwidth]{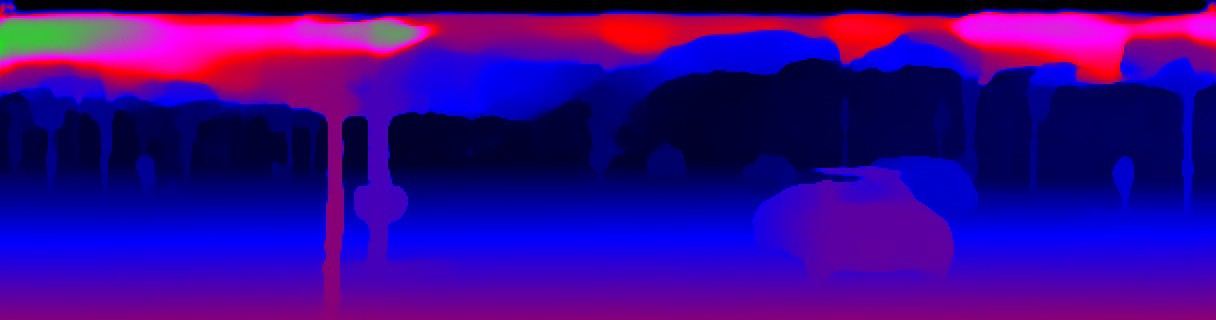}} \quad
  \subfloat[VarGNet v1 0.25]{\includegraphics[width=0.3\textwidth]{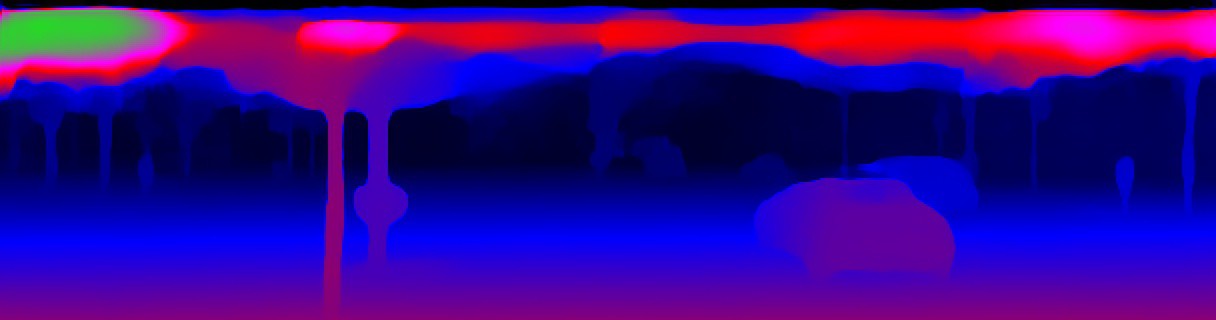}} \\
  \subfloat[VarGNet v2 1.0]{\includegraphics[width=0.3\textwidth]{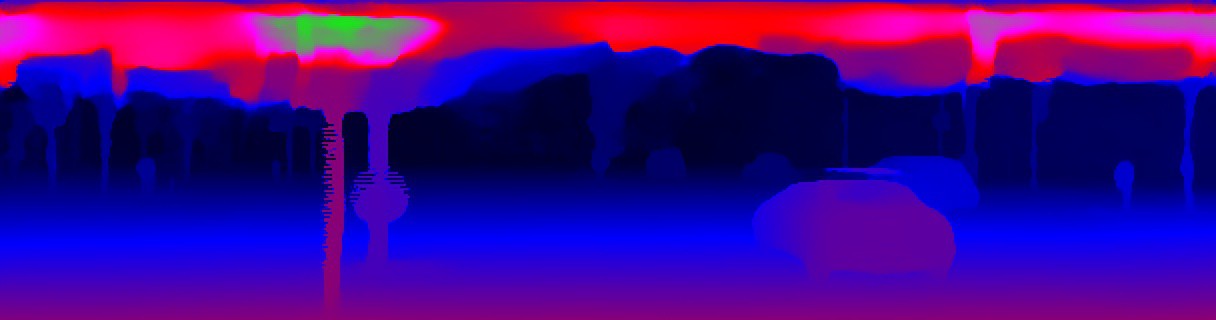}} \quad
  \subfloat[VarGNet v2 0.5]{\includegraphics[width=0.3\textwidth]{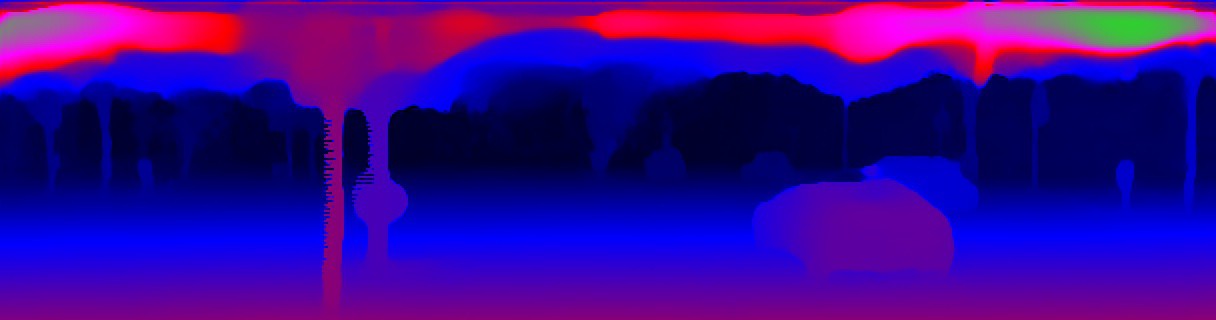}} \quad
  \subfloat[VarGNet v2 0.25]{\includegraphics[width=0.3\textwidth]{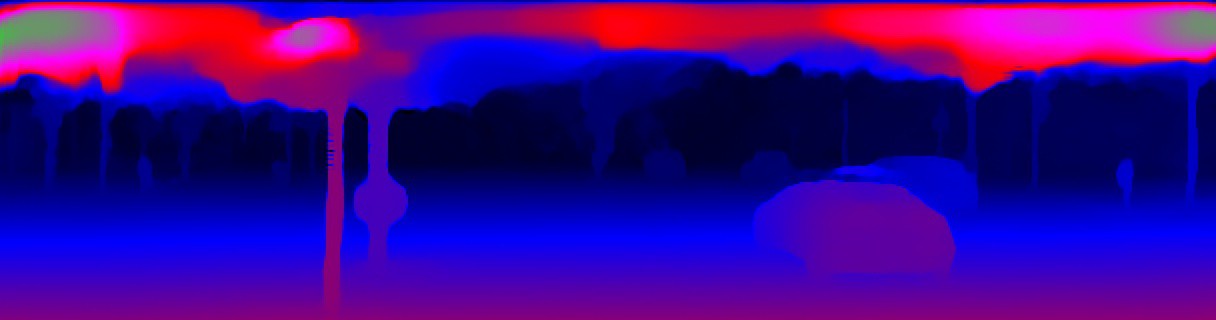}} 
  \caption{Visualization of stereo results on KITTI15.}\label{fig:kittistereo}  
\end{figure}

\paragraph{Results.} In Table. \ref{tab:kitti_depth} and Table. \ref{tab:kitti_ste}, we show our depth results and stereo results under various evaluation metrics. Also, we report our implemented MobileNet and ResNet as comparison. Further, visual effects are presented in Fig. \ref{fig:kittidepth} and Fig. \ref{fig:kittistereo}.

\begin{table}
  \centering
  \caption{Depth results on KITTI test set.}\label{tab:kitti_depth}
  % \caption{Depth results on KITTI test set. (*)MobileNet means our optimized version.}\label{tab:kitti_depth}
  \resizebox{\textwidth}{!}{
  \begin{tabular}{c}
    (a) 0-80m \\
    \begin{tabular}{lccccccccc}
      \toprule
      Method & AbsRel & SqlRel & RMSE & RMSE Log & $\delta$<1.25 & $\delta$<1.25$^2$ & $\delta$<1.25$^3$ & MAdds(G) & Params  \\ \midrule
      MobileNet v2 1.0 &0.103 & 0.744 & 4.686 & 0.17 & 0.888 & 0.966 & 0.987 & 36.8 & 7.6 M \\
      MobileNet v2 0.5 &0.112 & 0.865 & 5.01 & 0.183 & 0.869 & 0.959 & 0.983 & 10.0 & 1.9 M \\
      MobileNet v2 0.25 &0.113 & 0.831 & 4.988 & 0.183 & 0.866 & 0.96 & 0.985 & 2.9 & 539.2 K \\ 
      % (*)MobileNet v2 1.0 & 0.112 & 0.836 & 4.969 & 0.181 & 0.869 & 0.961 & 0.985 & 9.7 & 3.4 M \\
      % (*)MobileNet v2 0.5 & 0.116 & 0.886 & 5.123 & 0.188 & 0.861 & 0.958 & 0.984 & 2.7 & 904.1 K \\
      % (*)MobileNet v2 0.25 & 0.127 & 0.976 & 5.348 & 0.199 & 0.838 & 0.952 & 0.982 & 0.8 & 237.5 K \\ 
      ResNet 18 & 0.109 & 0.767 & 4.76 & 0.178 & 0.869 & 0.961 & 0.986 & 203.4 & 30.6 M \\
      ResNet 50 & 0.109 & 0.788 & 4.796 & 0.18 & 0.868 & 0.959 & 0.984 & 247.5 & 46.7 M \\ \midrule
      VarGNet v1 1.0 &  0.105 & 0.798 & 4.92 & 0.175 & 0.883 & 0.965 & 0.986  & 36.0   &13.2 M\\
      VarGNet v1 0.5 &  0.107 & 0.803 & 4.86 & 0.175 & 0.881 & 0.964 & 0.986  & 12.8   &3.8 M\\
      VarGNet v1 0.25 & 0.113 & 0.845 & 5.003 & 0.18 & 0.87 & 0.962 & 0.986   &  5.1 &1.2 M\\ \midrule
      VarGNet v2 1.0 &  0.108 & 0.823 & 4.898 & 0.176 & 0.881 & 0.965 & 0.986 & 20.0& 7.4 M\\
      VarGNet v2 0.5 &  0.111 & 0.851 & 4.98 & 0.179 & 0.874 & 0.961 & 0.985  & 7.7& 2.2 M\\
      VarGNet v2 0.25 & 0.118 & 0.9 & 5.11 & 0.186 & 0.863 & 0.959 & 0.985    & 3.3& 788.1 K\\ \bottomrule
    \end{tabular}
     \\
    (b) 0-50m \\
    % \resizebox{\textwidth}{!}{
      \begin{tabular}{lccccccccc}
      \toprule
      Method & AbsRel & SqlRel & RMSE & RMSE Log & $\delta$<1.25 & $\delta$<1.25$^2$ & $\delta$<1.25$^3$ & MAdds(G) & Params  \\ \midrule
      MobileNet v2 1.0 & 0.097 & 0.557 & 3.424 & 0.155 & 0.903 & 0.972 & 0.989 & 36.8 & 7.6 M \\
      MobileNet v2 0.5 & 0.106 & 0.649 & 3.665 & 0.167 & 0.886 & 0.966 & 0.986 & 10.0 & 1.9 M \\
      MobileNet v2 0.25 & 0.106 & 0.63 & 3.693 & 0.168 & 0.883 & 0.966 & 0.988 & 2.9 & 539.2 K \\
      % (*)MobileNet v2 1.0 & 0.106 & 0.634 & 3.669 & 0.166 & 0.886 & 0.968 & 0.988 & 9.7    & 3.4 M \\
      % (*)MobileNet v2 0.5 & 0.11 & 0.675 & 3.775 & 0.172 & 0.878 & 0.965 & 0.987 & 2.7    & 904.1 K \\
      % (*)MobileNet v2 0.25 & 0.121 & 0.746 & 3.974 & 0.183 & 0.856 & 0.96 & 0.985 & 0.8    & 237.5 K \\ 
      ResNet 18 & 0.104 & 0.584 & 3.525 & 0.164 & 0.883 & 0.967 & 0.988 & 203.4    & 30.6 M \\
      ResNet 50 & 0.104 & 0.592 & 3.521 & 0.165 & 0.883 & 0.965 & 0.987 & 247.5    & 46.7 M \\ \midrule
      VarGNet v1 1.0 &  0.098 & 0.578 & 3.534 & 0.158 & 0.899 & 0.973 & 0.99 &  36.0   &13.2 M\\
      VarGNet v1 0.5 &  0.1 & 0.603 & 3.535 & 0.159 & 0.897 & 0.97 & 0.989 &  12.8   &3.8 M\\
      VarGNet v1 0.25 & 0.106 & 0.637 & 3.648 & 0.165 & 0.887 & 0.969 & 0.989 &  5.1 &1.2 M\\\midrule
      VarGNet v2 1.0 & 0.101 & 0.612 & 3.556 & 0.16 & 0.896 & 0.971 & 0.989 & 20.0& 7.4 M\\
      VarGNet v2 0.5 & 0.104 & 0.635 & 3.639 & 0.163 & 0.89 & 0.968 & 0.988 & 7.7& 2.2 M\\
      VarGNet v2 0.25 & 0.112 & 0.681 & 3.768 & 0.171 & 0.88 & 0.966 & 0.988 & 3.3& 788.1 K\\ \bottomrule
    \end{tabular}
  \end{tabular}
  }
\end{table}

\begin{table}
  \centering
  \caption{Stereo results on KITTI.}\label{tab:kitti_ste}
  % \caption{Stereo results on KITTI. (*)MobileNet means our optimized version.}\label{tab:kitti_ste}
  % \resizebox{\textwidth}{!}{
    \begin{tabular}{c}
      (a) On KITTI RAW \\
      \begin{tabular}{lcccc}
        \toprule
        Method & EPE & D1 & MAdds(G) & Params  \\ \midrule
      MobileNet v2 1.0 & 1.424 & 0.0777 & 37.0 & 7.6 M \\
      MobileNet v2 0.5 & 1.4904 & 0.0832 & 10.1 & 1.9 M \\
      MobileNet v2 0.25 & 1.5897 & 0.0902 & 2.9 & 539.5 K \\
      ResNet 18 & 1.5269 & 0.0886 & 205.4    & 30.6 M \\
      ResNet 50 & 1.531 & 0.0887 & 249.5    & 46.7 M \\ \midrule
      VarGNet v1 1.0 & 1.3296 & 0.0703 & 36.1    & 13.2 M \\
      VarGNet v1 0.5 & 1.4045 & 0.0757 & 12.9    & 3.8 M \\
      VarGNet v1 0.25 & 1.5111 & 0.0835 & 5.1    & 1.2 M \\ \midrule
      VarGNet v2 1.0 & 1.3582 & 0.0728 & 20.7    & 7.4 M \\
      VarGNet v2 0.5 & 1.44 & 0.079 & 8.0    & 2.2 M \\
      VarGNet v2 0.25 & 1.5346 & 0.0862 & 3.4    & 790.2 K \\ \bottomrule
    \end{tabular} \\
    (b) On KITTI 15 \\
    \begin{tabular}{lcccc}
      \toprule
      Method & EPE & D1 & MAdds & Params  \\ \midrule
      MobileNet v2 1.0 &  1.7387 & 0.0753 & 37.0 & 7.6 M \\
      MobileNet v2 0.5 &  1.6861 & 0.0772 & 10.1 & 1.9 M \\
      MobileNet v2 0.25 & 1.6754 & 0.0819 & 2.9 & 539.5 K \\
      ResNet 18 & 1.7318 & 0.0873 & 205.4    & 30.6 M \\
      ResNet 50 & 1.7305 & 0.0868 & 249.5    & 46.7 M \\ \midrule
      VarGNet v1 1.0 & 1.5767 & 0.07 & 36.1    & 13.2 M \\
      VarGNet v1 0.5 & 1.5868 & 0.0708 & 12.9    & 3.8 M \\
      VarGNet v1 0.25 & 1.6685 & 0.0747 & 5.1    & 1.2 M \\ \midrule
      VarGNet v2 1.0 & 1.5856 & 0.0697 & 20.7    & 7.4 M \\
      VarGNet v2 0.5 & 1.5994 & 0.0735 & 8.0    & 2.2 M \\
      VarGNet v2 0.25 & 1.6302 & 0.0777 & 3.4    & 790.2 K \\ \bottomrule
    \end{tabular}
  \end{tabular}
  % }
\end{table}

\subsection{Face Recognition}
% \paragraph{Training setup}
All the networks are trained on the DeepGlint MS-Celeb-1M-v1c dataset \cite{dg} cleaned from MS-Celeb-1M \cite{guo2016ms}. There are 3,923,399 aligned face images from 86,876 ids. The LFW \cite{huang2008labeled} , CFP-FP \cite{sengupta2016frontal} and AgeDB-30 \cite{moschoglou2017agedb} are used as the validation datasets. Finally, all network models are evaluated on MegaFace Challenge 1 \cite{nech2017level}. Table. \ref{tab:face} lists the best face recognition accuracies on validation datasets, as well as face verification true accepted rates under 1e-6 false accepted rate on the refined version of MegaFace dataset \cite{deng2018arcface}. 
We use MobileNet v1 and MobileNet v2 as baseline models. To adapt the input image size of 112x112, the stride of the first convolutional layer is set to 1 for each baseline and vagnet model. To achieve better performance, we further replace the pooling layer by a ``BN-Dropout-FC-BN'' structure as InsightFace \cite{deng2018arcface}, followed by the ArcFace loss \cite{deng2018arcface}. 
The standard SGD optimizer is used with momentum 0.9 and the batch-size is set to 512 with 8 GPUs. The learning rate begins with 0.1 and is divided by 10 at the 100K, 140K and 160K iterations. We set the weight decay to be 5e-4. The embedding feature dimension is 256 with 0.4 dropout rate. The normalization scale is 64 and the ArcFace margin is set to 0.5. All training are based on the InsightFace toolbox \cite{deng2018arcface}. 

\begin{table}
  \centering
  \caption{Face recognition results.}\label{tab:face}
  % \resizebox{\textwidth}{!}{
  \begin{tabular}{lccccc}
      \toprule
      Networks & MAdds & LFW \cite{huang2008labeled} & CFP-FP \cite{sengupta2016frontal} & AgeDB-30 \cite{moschoglou2017agedb} & MegaFace \cite{deng2018arcface} \\ \midrule
MobileNet v1 & 554M & 0.99617 & 0.89714 & 0.96600 & 0.935848 \\ 
MobileNet v2 & 313M & 0.99500 & 0.86386 & 0.95583 & 0.898219 \\ 
VarGNet v1 & 603M & 0.99733 & 0.88929 & 0.97583 & 0.961499 \\ 
VarGNet v2 & 355M & 0.99733 & 0.89829 & 0.97333 & 0.954261 \\ \bottomrule
\end{tabular}
% }
\end{table}

\bibliographystyle{plain}
\bibliography{varg}

\begin{thebibliography}{10}

\bibitem{dg}
\url{http://trillionpairs.deepglint.com/overview}.

\bibitem{abdelfattah2018dla}
Mohamed~S Abdelfattah, David Han, Andrew Bitar, Roberto DiCecco, Shane
  O'Connell, Nitika Shanker, Joseph Chu, Ian Prins, Joshua Fender, Andrew~C
  Ling, et~al.
\newblock Dla: Compiler and fpga overlay for neural network inference
  acceleration.
\newblock In {\em International Conference on Field Programmable Logic and
  Applications}, pages 411--4117. IEEE, 2018.

\bibitem{badrinarayanan2017segnet}
Vijay Badrinarayanan, Alex Kendall, and Roberto Cipolla.
\newblock Segnet: A deep convolutional encoder-decoder architecture for image
  segmentation.
\newblock {\em IEEE Transactions on Pattern Analysis and Machine Intelligence},
  39(12):2481--2495, 2017.

\bibitem{cai2018proxylessnas}
Han Cai, Ligeng Zhu, and Song Han.
\newblock Proxyless{NAS}: Direct neural architecture search on target task and
  hardware.
\newblock In {\em International Conference on Learning Representations (ICLR)},
  2019.

\bibitem{chang2017compiling}
Andre Xian~Ming Chang, Aliasger Zaidy, Vinayak Gokhale, and Eugenio
  Culurciello.
\newblock Compiling deep learning models for custom hardware accelerators.
\newblock {\em arXiv preprint arXiv:1708.00117}, 2017.

\bibitem{chen2018tvm}
Tianqi Chen, Thierry Moreau, Ziheng Jiang, Haichen Shen, Eddie~Q Yan, Leyuan
  Wang, Yuwei Hu, Luis Ceze, Carlos Guestrin, and Arvind Krishnamurthy.
\newblock Tvm: end-to-end optimization stack for deep learning.
\newblock {\em arXiv preprint arXiv:1802.04799}, pages 1--15, 2018.

\bibitem{chen2014diannao}
Tianshi Chen, Zidong Du, Ninghui Sun, Jia Wang, Chengyong Wu, Yunji Chen, and
  Olivier Temam.
\newblock Diannao: A small-footprint high-throughput accelerator for ubiquitous
  machine-learning.
\newblock In {\em ACM Sigplan Notices}, volume~49, pages 269--284. ACM, 2014.

\bibitem{chollet2017xception}
Fran{\c{c}}ois Chollet.
\newblock Xception: Deep learning with depthwise separable convolutions.
\newblock In {\em IEEE Conference on Computer Vision and Pattern Recognition
  (CVPR)}, pages 1251--1258, 2017.

\bibitem{cordts2016cityscapes}
Marius Cordts, Mohamed Omran, Sebastian Ramos, Timo Rehfeld, Markus Enzweiler,
  Rodrigo Benenson, Uwe Franke, Stefan Roth, and Bernt Schiele.
\newblock The cityscapes dataset for semantic urban scene understanding.
\newblock In {\em IEEE Conference on Computer Vision and Pattern Recognition
  (CVPR)}, pages 3213--3223, 2016.

\bibitem{dai2018chamnet}
Xiaoliang Dai, Peizhao Zhang, Bichen Wu, Hongxu Yin, Fei Sun, Yanghan Wang,
  Marat Dukhan, Yunqing Hu, Yiming Wu, Yangqing Jia, et~al.
\newblock Chamnet: Towards efficient network design through platform-aware
  model adaptation.
\newblock {\em arXiv preprint arXiv:1812.08934}, 2018.

\bibitem{deng2018arcface}
Jiankang Deng, Jia Guo, Niannan Xue, and Stefanos Zafeiriou.
\newblock Arcface: Additive angular margin loss for deep face recognition.
\newblock {\em arXiv preprint arXiv:1801.07698}, 2018.

\bibitem{eigen2014depth}
David Eigen, Christian Puhrsch, and Rob Fergus.
\newblock Depth map prediction from a single image using a multi-scale deep
  network.
\newblock In {\em Advances in neural information processing systems}, pages
  2366--2374, 2014.

\bibitem{FarabetPHL09}
Cl{\'{e}}ment Farabet, Cyril Poulet, Jefferson~Y. Han, and Yann LeCun.
\newblock {CNP:} an fpga-based processor for convolutional networks.
\newblock In {\em International Conference on Field Programmable Logic and
  Applications}, pages 32--37, 2009.

\bibitem{geiger2013vision}
Andreas Geiger, Philip Lenz, Christoph Stiller, and Raquel Urtasun.
\newblock Vision meets robotics: The kitti dataset.
\newblock {\em The International Journal of Robotics Research},
  32(11):1231--1237, 2013.

\bibitem{guo2016angel}
Kaiyuan Guo, Lingzhi Sui, Jiantao Qiu, Song Yao, Song Han, Yu~Wang, and
  Huazhong Yang.
\newblock Angel-eye: A complete design flow for mapping cnn onto customized
  hardware.
\newblock In {\em IEEE Computer Society Annual Symposium on VLSI (ISVLSI)},
  pages 24--29. IEEE, 2016.

\bibitem{guo2016ms}
Yandong Guo, Lei Zhang, Yuxiao Hu, Xiaodong He, and Jianfeng Gao.
\newblock Ms-celeb-1m: A dataset and benchmark for large-scale face
  recognition.
\newblock In {\em European Conference on Computer Vision (ECCV)}, pages
  87--102. Springer, 2016.

\bibitem{gupta2015deep}
Suyog Gupta, Ankur Agrawal, Kailash Gopalakrishnan, and Pritish Narayanan.
\newblock Deep learning with limited numerical precision.
\newblock In {\em International Conference on Machine Learning (ICML)}, pages
  1737--1746, 2015.

\bibitem{he2016deep}
Kaiming He, Xiangyu Zhang, Shaoqing Ren, and Jian Sun.
\newblock Deep residual learning for image recognition.
\newblock In {\em IEEE conference on computer vision and pattern recognition
  (CVPR)}, pages 770--778, 2016.

\bibitem{hegde2018ucnn}
Kartik Hegde, Jiyong Yu, Rohit Agrawal, Mengjia Yan, Michael Pellauer, and
  Christopher~W Fletcher.
\newblock Ucnn: Exploiting computational reuse in deep neural networks via
  weight repetition.
\newblock In {\em ACM/IEEE Annual International Symposium on Computer
  Architecture (ISCA)}, pages 674--687. IEEE Press, 2018.

\bibitem{howard2017mobilenets}
Andrew~G Howard, Menglong Zhu, Bo~Chen, Dmitry Kalenichenko, Weijun Wang,
  Tobias Weyand, Marco Andreetto, and Hartwig Adam.
\newblock Mobilenets: Efficient convolutional neural networks for mobile vision
  applications.
\newblock {\em arXiv preprint arXiv:1704.04861}, 2017.

\bibitem{huang2008labeled}
Gary~B Huang, Marwan Mattar, Tamara Berg, and Eric Learned-Miller.
\newblock Labeled faces in the wild: A database forstudying face recognition in
  unconstrained environments.
\newblock In {\em Workshop on faces in'Real-Life'Images: detection, alignment,
  and recognition}, 2008.

\bibitem{2016_SqueezeNet}
Forrest~N. Iandola, Song Han, Matthew~W. Moskewicz, Khalid Ashraf, William~J.
  Dally, and Kurt Keutzer.
\newblock {SqueezeNet}: Alexnet-level accuracy with 50x fewer parameters and
  $<$0.5mb model size.
\newblock {\em arXiv:1602.07360}, 2016.

\bibitem{jouppi2017datacenter}
Norman~P Jouppi, Cliff Young, Nishant Patil, David Patterson, Gaurav Agrawal,
  Raminder Bajwa, Sarah Bates, Suresh Bhatia, Nan Boden, Al~Borchers, et~al.
\newblock In-datacenter performance analysis of a tensor processing unit.
\newblock In {\em ACM/IEEE Annual International Symposium on Computer
  Architecture (ISCA)}, pages 1--12. IEEE, 2017.

\bibitem{Panoptic}
Alexander Kirillov, Ross~B. Girshick, Kaiming He, and Piotr Doll{\'{a}}r.
\newblock Panoptic feature pyramid networks.
\newblock {\em arXiv preprint arXiv:1901.02446}, 2019.

\bibitem{krizhevsky2012imagenet}
Alex Krizhevsky, Ilya Sutskever, and Geoffrey~E Hinton.
\newblock Imagenet classification with deep convolutional neural networks.
\newblock In {\em Advances in neural information processing systems}, pages
  1097--1105, 2012.

\bibitem{Lin2017FeaturePN}
Tsung-Yi Lin, Piotr Doll{\'a}r, Ross~B. Girshick, Kaiming He, Bharath
  Hariharan, and Serge~J. Belongie.
\newblock Feature pyramid networks for object detection.
\newblock {\em IEEE Conference on Computer Vision and Pattern Recognition
  (CVPR)}, pages 936--944, 2017.

\bibitem{Lin2014MicrosoftCC}
Tsung-Yi Lin, Michael Maire, Serge~J. Belongie, Lubomir~D. Bourdev, Ross~B.
  Girshick, James Hays, Pietro Perona, Deva Ramanan, Piotr Doll{\'a}r, and
  C.~Lawrence Zitnick.
\newblock Microsoft coco: Common objects in context.
\newblock In {\em European Conference on Computer Vision (ECCV)}, 2014.

\bibitem{liu2018darts}
Hanxiao Liu, Karen Simonyan, and Yiming Yang.
\newblock {DARTS}: Differentiable architecture search.
\newblock In {\em International Conference on Learning Representations (ICLR)},
  2019.

\bibitem{luo2017dadiannao}
Tao Luo, Shaoli Liu, Ling Li, Yuqing Wang, Shijin Zhang, Tianshi Chen, Zhiwei
  Xu, Olivier Temam, and Yunji Chen.
\newblock Dadiannao: A neural network supercomputer.
\newblock {\em IEEE Transactions on Computers}, 66(1):73--88, 2017.

\bibitem{ma2018shufflenet}
Ningning Ma, Xiangyu Zhang, Hai-Tao Zheng, and Jian Sun.
\newblock Shufflenet v2: Practical guidelines for efficient cnn architecture
  design.
\newblock In {\em European Conference on Computer Vision (ECCV)}, pages
  116--131, 2018.

\bibitem{ma2017optimizing}
Yufei Ma, Yu~Cao, Sarma Vrudhula, and Jae-sun Seo.
\newblock Optimizing loop operation and dataflow in fpga acceleration of deep
  convolutional neural networks.
\newblock In {\em ACM/SIGDA International Symposium on Field-Programmable Gate
  Arrays}, pages 45--54. ACM, 2017.

\bibitem{moschoglou2017agedb}
Stylianos Moschoglou, Athanasios Papaioannou, Christos Sagonas, Jiankang Deng,
  Irene Kotsia, and Stefanos Zafeiriou.
\newblock Agedb: the first manually collected, in-the-wild age database.
\newblock In {\em Proceedings of the IEEE Conference on Computer Vision and
  Pattern Recognition Workshops}, pages 51--59, 2017.

\bibitem{nech2017level}
Aaron Nech and Ira Kemelmacher-Shlizerman.
\newblock Level playing field for million scale face recognition.
\newblock In {\em IEEE Conference on Computer Vision and Pattern Recognition
  (CVPR)}, pages 7044--7053, 2017.

\bibitem{paszke2016enet}
Adam Paszke, Abhishek Chaurasia, Sangpil Kim, and Eugenio Culurciello.
\newblock Enet: A deep neural network architecture for real-time semantic
  segmentation.
\newblock {\em arXiv preprint arXiv:1606.02147}, 2016.

\bibitem{pham2018efficient}
Hieu Pham, Melody~Y. Guan, Barret Zoph, Quoc~V. Le, and Jeff Dean.
\newblock Efficient neural architecture search via parameter sharing.
\newblock In {\em International Conference on Machine Learning (ICML)}, pages
  4092--4101, 2018.

\bibitem{reagen2016minerva}
Brandon Reagen, Paul Whatmough, Robert Adolf, Saketh Rama, Hyunkwang Lee,
  Sae~Kyu Lee, Jos{\'e}~Miguel Hern{\'a}ndez-Lobato, Gu-Yeon Wei, and David
  Brooks.
\newblock Minerva: Enabling low-power, highly-accurate deep neural network
  accelerators.
\newblock In {\em ACM/IEEE Annual International Symposium on Computer
  Architecture (ISCA)}, pages 267--278. IEEE, 2016.

\bibitem{Real2018Regularized}
Esteban Real, Alok Aggarwal, Yanping Huang, and Quoc~V. Le.
\newblock Regularized evolution for image classifier architecture search.
\newblock {\em CoRR}, abs/1802.01548, 2018.

\bibitem{sandler2018mobilenetv2}
Mark Sandler, Andrew Howard, Menglong Zhu, Andrey Zhmoginov, and Liang-Chieh
  Chen.
\newblock Mobilenetv2: Inverted residuals and linear bottlenecks.
\newblock In {\em IEEE Conference on Computer Vision and Pattern Recognition
  (CVPR)}, pages 4510--4520, 2018.

\bibitem{sengupta2016frontal}
Soumyadip Sengupta, Jun-Cheng Chen, Carlos Castillo, Vishal~M Patel, Rama
  Chellappa, and David~W Jacobs.
\newblock Frontal to profile face verification in the wild.
\newblock In {\em IEEE Winter Conference on Applications of Computer Vision
  (WACV)}, pages 1--9. IEEE, 2016.

\bibitem{stamoulis2019single}
Dimitrios Stamoulis, Ruizhou Ding, Di~Wang, Dimitrios Lymberopoulos, Bodhi
  Priyantha, Jie Liu, and Diana Marculescu.
\newblock Single-path nas: Designing hardware-efficient convnets in less than 4
  hours.
\newblock {\em arXiv preprint arXiv:1904.02877}, 2019.

\bibitem{sun2018igcv3}
Ke~Sun, Mingjie Li, Dong Liu, and Jingdong Wang.
\newblock Igcv3: Interleaved low-rank group convolutions for efficient deep
  neural networks.
\newblock {\em arXiv preprint arXiv:1806.00178}, 2018.

\bibitem{venieris2017fpgaconvnet}
Stylianos~I Venieris and Christos-Savvas Bouganis.
\newblock fpgaconvnet: Automated mapping of convolutional neural networks on
  fpgas.
\newblock In {\em ACM/SIGDA International Symposium on Field-Programmable Gate
  Arrays}, pages 291--292. ACM, 2017.

\bibitem{venieris2018toolflows}
Stylianos~I Venieris, Alexandros Kouris, and Christos-Savvas Bouganis.
\newblock Toolflows for mapping convolutional neural networks on fpgas: A
  survey and future directions.
\newblock {\em ACM Computing Surveys (CSUR)}, 51(3):56, 2018.

\bibitem{fbnet}
Bichen Wu, Xiaoliang Dai, Peizhao Zhang, Yanghan Wang, Fei Sun, Yiming Wu,
  Yuandong Tian, Peter Vajda, Yangqing Jia, and Kurt Keutzer.
\newblock Fbnet: Hardware-aware efficient convnet design via differentiable
  neural architecture search.
\newblock {\em CoRR}, abs/1812.03443, 2018.

\bibitem{xiao2017exploring}
Qingcheng Xiao, Yun Liang, Liqiang Lu, Shengen Yan, and Yu-Wing Tai.
\newblock Exploring heterogeneous algorithms for accelerating deep
  convolutional neural networks on fpgas.
\newblock In {\em ACM/EDAC/IEEE Design Automation Conference (DAC)}, pages
  1--6. IEEE, 2017.

\bibitem{xie2018interleaved}
Guotian Xie, Jingdong Wang, Ting Zhang, Jianhuang Lai, Richang Hong, and
  Guo-Jun Qi.
\newblock Interleaved structured sparse convolutional neural networks.
\newblock In {\em IEEE Conference on Computer Vision and Pattern Recognition
  (CVPR)}, pages 8847--8856, 2018.

\bibitem{xie2017aggregated}
Saining Xie, Ross Girshick, Piotr Doll{\'a}r, Zhuowen Tu, and Kaiming He.
\newblock Aggregated residual transformations for deep neural networks.
\newblock In {\em IEEE Conference on Computer Vision and Pattern Recognition
  (CVPR)}, pages 1492--1500, 2017.

\bibitem{xing2019dnnvm}
Yu~Xing, Shuang Liang, Lingzhi Sui, Xijie Jia, Jiantao Qiu, Xin Liu, Yushun
  Wang, Yu~Wang, and Yi~Shan.
\newblock Dnnvm: End-to-end compiler leveraging heterogeneous optimizations on
  fpga-based cnn accelerators.
\newblock {\em arXiv preprint arXiv:1902.07463}, 2019.

\bibitem{yu2018bisenet}
Changqian Yu, Jingbo Wang, Chao Peng, Changxin Gao, Gang Yu, and Nong Sang.
\newblock Bisenet: Bilateral segmentation network for real-time semantic
  segmentation.
\newblock In {\em European Conference on Computer Vision (ECCV)}, pages
  325--341, 2018.

\bibitem{ZhangLSGXC15}
Chen Zhang, Peng Li, Guangyu Sun, Yijin Guan, Bingjun Xiao, and Jason Cong.
\newblock Optimizing fpga-based accelerator design for deep convolutional
  neural networks.
\newblock In {\em ACM/SIGDA International Symposium on Field-Programmable Gate
  Arrays}, pages 161--170, 2015.

\bibitem{zhang2017interleaved}
Ting Zhang, Guo-Jun Qi, Bin Xiao, and Jingdong Wang.
\newblock Interleaved group convolutions.
\newblock In {\em IEEE International Conference on Computer Vision (ICCV)},
  pages 4373--4382, 2017.

\bibitem{zhang2018shufflenet}
Xiangyu Zhang, Xinyu Zhou, Mengxiao Lin, and Jian Sun.
\newblock Shufflenet: An extremely efficient convolutional neural network for
  mobile devices.
\newblock In {\em IEEE Conference on Computer Vision and Pattern Recognition
  (CVPR)}, pages 6848--6856, 2018.

\bibitem{zoph2016neural}
Barret Zoph and Quoc~V. Le.
\newblock Neural architecture search with reinforcement learning.
\newblock {\em CoRR}, abs/1611.01578, 2016.

\bibitem{zoph2017learning}
Barret Zoph, Vijay Vasudevan, Jonathon Shlens, and Quoc~V. Le.
\newblock Learning transferable architectures for scalable image recognition.
\newblock {\em CoRR}, abs/1707.07012, 2017.

\end{thebibliography}

\end{document}